\documentclass[lettersize,journal, compsoc]{IEEEtran}
\usepackage{amsmath,amsfonts}
\usepackage{algorithmic}
\usepackage{algorithm}
\usepackage{array}
\usepackage[caption=false,font=normalsize,labelfont=sf,textfont=sf]{subfig}

\usepackage{textcomp}
\usepackage{stfloats}
\usepackage{url}
\usepackage{verbatim}
\usepackage{graphicx}
\usepackage{cite}
\usepackage{lipsum}
\usepackage{tikz}
\usepackage{hyperref}
\usepackage{xspace}
\usepackage{tabularx}
\usepackage{booktabs}
\usepackage{amssymb}
\usetikzlibrary{shapes.multipart, calc}
\usepackage{tikz}
\usepackage[edges]{forest}
\tikzstyle{mybox}=[
    rectangle,
    draw=hiddendraw,
    rounded corners,
    text opacity=1,
    minimum height=1.5em,
    minimum width=5em,
    inner sep=2pt,
    align=center,
    fill opacity=.5,
    ]

\begin{document}

\title{A Survey on Diffusion Language Models}

\author{Tianyi Li, 
        Mingda Chen,
        Bowei Guo,
        and Zhiqiang Shen
    
    \IEEEcompsocitemizethanks{
        \IEEEcompsocthanksitem Tianyi Li, Mingda Chen, Bowei Guo, and Zhiqiang Shen are with VILA Lab, 
        Mohamed bin Zayed University of Artificial Intelligence. Mingda Chen is also with Department of Automation, Tsinghua University.
        % \protect\\
        E-mail: \{Tianyi.Li, Bowei.Guo, Zhiqiang.Shen\}@mbzuai.ac.ae, cmd22@mails.tsinghua.edu.cn
    } 
}

% The paper headers
% \markboth{Journal of \LaTeX\ Class Files,~Vol.~14, No.~8, August~2021}%
\markboth{}
{Shell \MakeLowercase{\textit{et al.}}: A Sample Article Using IEEEtran.cls for IEEE Journals}

% \IEEEpubid{0000--0000/00\$00.00~\copyright~2021 IEEE}
% Remember, if you use this you must call \IEEEpubidadjcol in the second
% column for its text to clear the IEEEpubid mark.

\IEEEtitleabstractindextext{
\begin{abstract}
% Tianyi
Diffusion Language Models (DLMs) are rapidly emerging as a powerful and promising alternative to the dominant autoregressive (AR) paradigm. 
By generating tokens in parallel through an iterative denoising process, DLMs possess inherent advantages in reducing inference latency and capturing bidirectional context, thereby enabling fine-grained control over the generation process. While achieving several-fold speedups, recent advancements have allowed DLMs to show performance comparable to their autoregressive counterparts, making them a compelling choice for various natural language processing tasks.
Despite their growing prevalence, DLMs present challenges and opportunities that warrant further exploration, requiring a detailed and systematic understanding of their principles, techniques, and limitations. 
In this survey, we provide a holistic overview of the current DLM landscape. 
We trace its evolution and relationship with other paradigms, such as autoregressive and masked language models, and cover both foundational principles and state-of-the-art models. 
Our work offers an up-to-date, comprehensive taxonomy and an in-depth analysis of current techniques, from pre-training strategies to advanced post-training methods.
Another contribution of this survey is a thorough review of DLM inference strategies and optimizations, including improvements in decoding parallelism, caching mechanisms, and generation quality. 
We also highlight the latest approaches to multimodal extensions of DLMs and delineate their applications across various practical scenarios. 
Furthermore, our discussion addresses the limitations and challenges of DLMs, including efficiency, long-sequence handling, and infrastructure requirements, 
while outlining future research directions to sustain progress in this rapidly evolving field.
Project GitHub is available at \url{https://github.com/VILA-Lab/Awesome-DLMs}.

\end{abstract}


}

\maketitle

\section{Introduction}
% Tianyi
\label{sec:intro}
\IEEEPARstart{R}{ecent} advancements toward artificial general intelligence (AGI) have been largely driven by the emergence of autoregressive large language models (LLMs)~\cite{brown2020language,achiam2023gpt, chowdhery2023palm,touvron2023llama,bai2023qwen,zhao2023survey, guo2025deepseek} and diffusion models for image and video generation~\cite{rombach2022high,saharia2022photorealistic,podellsdxl, esser2024scaling,brooks2024video}.
These models exhibit remarkable capabilities in both understanding and generation across diverse modalities, achieving levels of performance that were previously unimaginable.
The unprecedented scale of these models, reflected in massive parameter counts, vast datasets, substantial efforts in training, and significant computational demands during inference, has pushed AI to new heights, equipping these models with broad general knowledge and a deep understanding of language and the real world.

\begin{figure*}[!t]
    \centering
    \includegraphics[width=\textwidth]{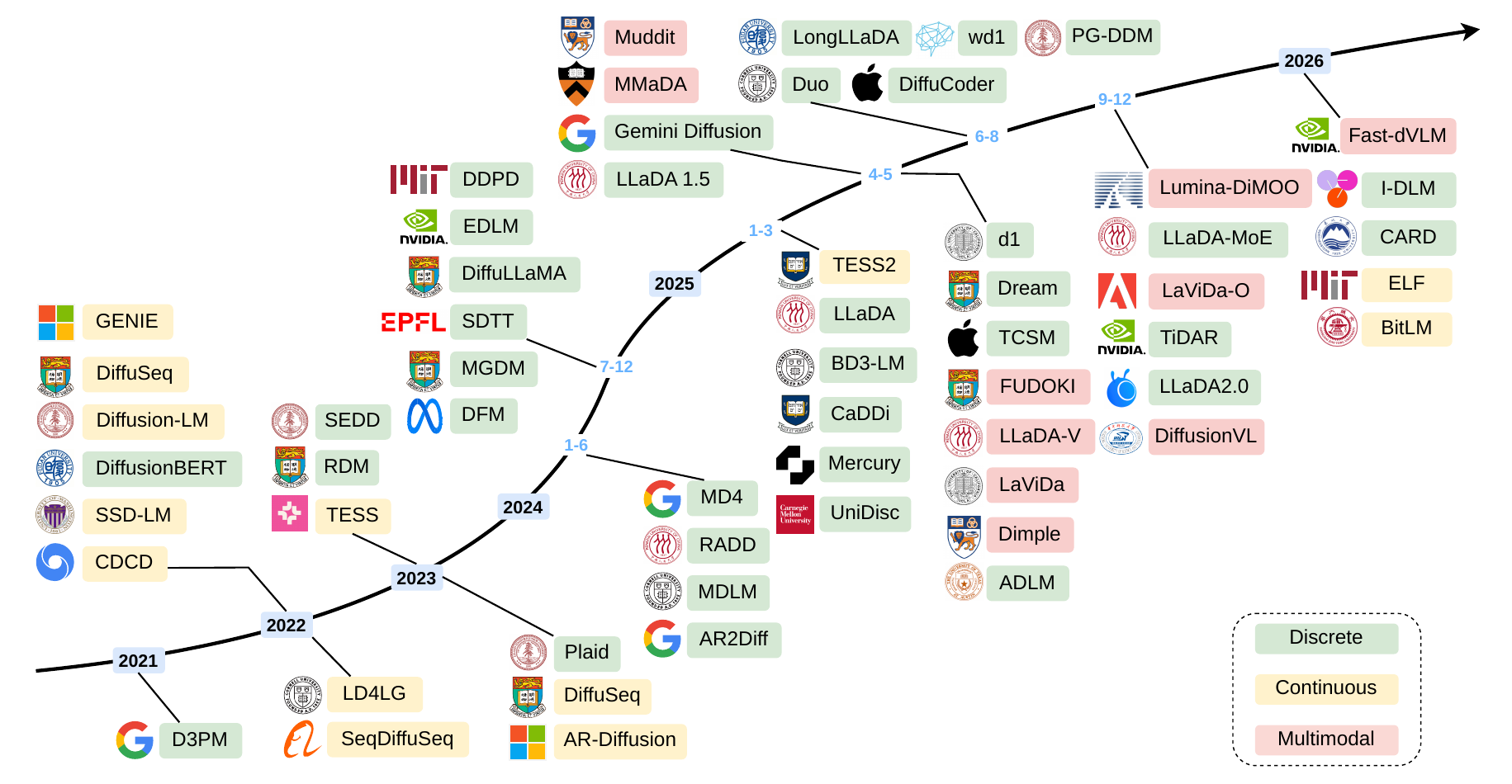}
    \caption{Timeline of Diffusion Language Models.  
    This figure highlights key milestones in the development of DLMs, categorized into three groups: continuous DLMs, discrete DLMs, and recent multimodal DLMs.  
    We observe that while early research predominantly focused on continuous DLMs, discrete DLMs have gained increasing popularity in more recent years. 
    }
    \label{fig:timeline}
\end{figure*}

The rise of the GPT series~\cite{radford2018improving, radford2019language, brown2020language}, particularly with the public release of ChatGPT~\cite{achiam2023gpt}, has propelled autoregressive (AR) language models to a dominant position in natural language processing.
By training to predict the next token using causal attention and teacher forcing, AR models~\cite{touvron2023llama, team2023gemini, liu2024deepseek} can effectively scale to large datasets and model sizes.
Generating text in a sequential, token-by-token fashion, AR models excel at supporting a wide range of tasks, from simple question answering to complex reasoning and creative writing.
However, this sequential nature imposes a major bottleneck on inference speed.
The autoregressive generation process, which produces one token at a time, inherently limits parallelism and significantly constrains computational efficiency and throughput.

Diffusion models are another highly promising generative paradigm.
They are trained to recover data from progressively noised versions through a denoising process, and generate new samples by reversing this stochastic corruption step by step.
Excelling at modeling complex data distributions, diffusion models have achieved state-of-the-art results in image and video synthesis~\cite{dhariwal2021diffusion}.
Academic breakthroughs in diffusion modeling~\cite{ho2020denoising, song2020denoising, songscore, liuflow} have established a solid theoretical foundation for training and inference.
Concurrently, large-scale practical models like Stable Diffusion~\cite{rombach2022high, podellsdxl, esser2024scaling}, Imagen~\cite{saharia2022photorealistic}, and Sora~\cite{brooks2024video} demonstrate the remarkable scalability and generalization of diffusion paradigm, enabling generation of high-fidelity, art-level images and videos from simple text prompts—often with just a few words.  
Beyond their strong capacity for modeling complex data distributions, diffusion models offer an inherent advantage in parallelism. 
Through an iterative denoising process, they can generate multiple tokens or an entire sequence simultaneously, potentially leading to superior inference throughput and better utilization of modern parallel computing hardware. 
While challenges remain, particularly in modeling discrete data and handling dynamic sequence lengths, 
Diffusion Language Models (DLMs) have emerged as a compelling alternative to address the trade-off between generation quality and speed. 

\begin{figure}[!t]
    \centering
    \includegraphics[page=1, width=0.49\textwidth, clip]{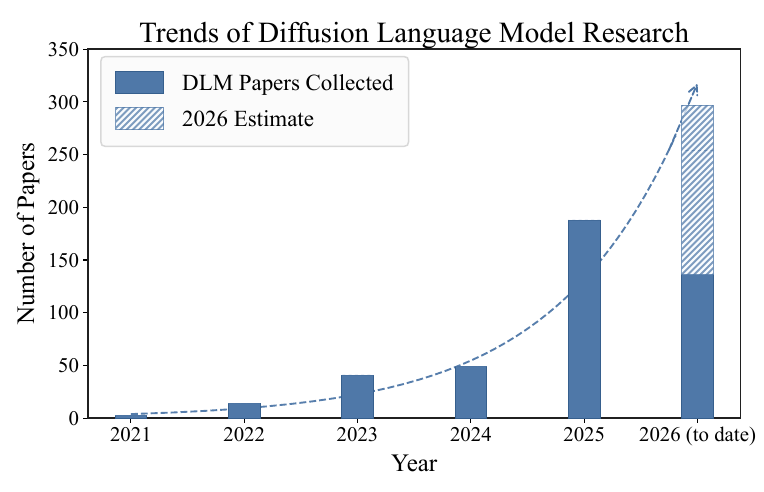}
    \caption{Trend of diffusion language model papers. Counts are based on unique dated arXiv entries collected in the associated Awesome-DLMs repository. For 2026, the lighter part shows a linear estimate based on papers collected by mid-May. The statistics are for reference only.
}
    \label{fig:trending}
\end{figure}

To adapt diffusion for discrete language data, several key approaches have been proposed.
In the early stages, the development of DLMs was primarily driven by diffusion models' success in continuous domains like image synthesis.
Continuous DLMs map tokens into embeddings and perform denoising in continuous space, as in pioneering works Diffusion-LM~\cite{li2022diffusion} and SED~\cite{strudel2022self}. 
Discrete DLMs, on the other hand, define the diffusion process directly in token space. 
Early efforts such as D3PM~\cite{austin2021structured} introduced structured transition matrices with absorbing states, allowing token-level corruption and iterative denoising. 
Subsequent work like DiffusionBERT~\cite{he2023diffusionbert} integrated pre-trained masked language models (e.g., BERT) to enhance denoising quality, and proposed tailored noise schedules (e.g., the spindle schedule) to better align token corruption with token frequency.
These early models demonstrated the feasibility of applying iterative denoising to non-autoregressive text generation, offering controllability and parallelism, though their performance still lagged behind strong autoregressive baselines.
As core challenges in DLMs are gradually addressed and the paradigm matures, larger-scale DLMs have been developed.
By initializing from autoregressive models, 7B-level models like Dream~\cite{dream2025} and DiffuLLaMA~\cite{gongscaling} have shown that DLMs can be effectively adapted from existing models while achieving competitive performance.  
LLaDA-8B~\cite{nie2025largelanguagediffusionmodels} further demonstrates the potential of training DLMs from scratch, achieving performance comparable to similarly sized LLaMA3-8B models. 
Multimodal DLMs, also known as diffusion multimodal large language models (dMLLMs), have also shown promise in modeling hybrid data such as text and images.  
Built upon open-source DLMs, models like LLaDA-V~\cite{you2025llada}, Dimple~\cite{yu2025dimple}, and MMaDA~\cite{yang2025mmada} integrate cross-modal reasoning and generation into the diffusion framework.  
% Mecury, Gemini Diffusion
Meanwhile, industry efforts have also shown growing interest in DLMs.  
The Mercury series~\cite{labs2025mercury}, Gemini Diffusion~\cite{deepmind2024geminidiffusion}, and Seed Diffusion~\cite{song2025seed} report strong performance while achieving inference speeds of thousands of tokens per second.  
These developments highlight the growing practicality and commercial potential of DLMs. We provide a timeline of DLMs' development in Fig.~\ref{fig:timeline}, ranging from representative models to recent advancements\cite{xu2024energy,deschenauxbeyond,han2024transfer,sahoo2025diffusion,zhang2025non,dang2025inference,rout2025anchored}, followed by a visualization of DLM trends in Fig.~\ref{fig:trending}.

Diffusion language models also present unique challenges and opportunities in both training and inference.  
Pretraining typically follows strategies similar to those used in autoregressive language models or image diffusion models~\cite{dream2025, yang2025mmada,yu2025dimple}.
To accelerate training and reuse previous training efforts, many DLMs are initialized from pretrained autoregressive model weights~\cite{gongscaling,dream2025}.
Supervised fine-tuning (SFT) in DLMs also mirrors that of autoregressive models: clean prompt data is provided, and the model learns to generate the target completion.
Reinforcement learning (RL) is also adopted for DLMs post-training to improve performance on complex tasks.  
Variants of GRPO~\cite{shao2024deepseekmath} algorithm such as diffu-GRPO~\cite{zhao2025d1} and UniGRPO~\cite{yang2025mmada} have been proposed to enhance the reasoning capabilities and alignment of DLMs at scale.
During inference, various strategies and optimizations have been developed to fully utilize the capabilities of DLMs.  
Continuous DLMs can leverage ODE/SDE solvers or other few-step generation techniques to accelerate the iterative denoising process~\cite{chen2025dlm}.
As discrete DLMs face more challenges in parallel generation, specialized parallel decoding strategies~\cite{wu2025fast,yu2025dimple,israel2025accelerating} have been proposed to enable acceptance of multiple tokens at a single step and overcome the parallelism curse.
Unmasking and remasking strategies~\cite{nie2025largelanguagediffusionmodels,wang2025remasking} further improve generation quality by selectively revealing low-confidence tokens, while caching techniques~\cite{liu2025dllm,ma2025dkv} can significantly reduce computation and enhance inference speed for both paradigms.

Compared to autoregressive models, diffusion language models are widely believed to offer several distinct advantages as follows:
\begin{itemize}
    \item \textbf{Parallel Generation:} DLMs can generate multiple tokens in parallel through an iterative denoising process, significantly improving inference speed and throughput over autoregressive models.
    \item \textbf{Bidirectional Context:} DLMs naturally incorporate bidirectional context, enabling more nuanced language understanding and generation. They also produce richer contextual embeddings, which are beneficial for cross-modal generation tasks. This enables fine-grained control over the generation process as well.
    \item \textbf{Iterative Refinement:} The iterative denoising process allows DLMs to update their perceptions over multiple steps. By accepting high-confidence tokens early and retaining low-confidence regions as masked, Masked DLMs can progressively improve uncertain areas, often resulting in more coherent and higher-quality text generation.
    \item \textbf{Controllability:} DLMs can be conditioned on specific token positions or structures, making them well-suited for tasks like infilling and structured generation. Additionally, guidance techniques (e.g., classifier-free guidance) enable better control over style and semantic relevance.
    \item \textbf{Unified Modeling Across Modalities:} By applying a shared denoising-based modeling framework, DLMs naturally support unified text and vision generation tasks. 
    This makes them particularly promising for multimodal applications that require both generation and understanding within a single model.
\end{itemize}

\begin{figure*}[!t]
    \centering
    \definecolor{c1}{RGB}{102,178,255} % Light Blue
    \definecolor{c2}{RGB}{255,179,102} % Orange
    \definecolor{c3}{RGB}{153,221,153} % Mint Green
    \definecolor{c4}{RGB}{204,179,255} % Lavender Purple

    \resizebox{\textwidth}{!}{
        \begin{forest}
            for tree={
                grow=east, reversed,          
                forked edges,       
                anchor=base west,   
                parent anchor=east,
                child anchor=west,
                base=center,
                font=\large,
                rectangle,
                draw=gray,
                rounded corners,
                align=center,       
                minimum width=6em,
                edge+={darkgray, line width=0.7mm},
                s sep=4pt,          
                inner xsep=5pt,     
                inner ysep=5pt,     
                line width=0.8pt,
                ver/.style={rotate=90, child anchor=north, parent anchor=south, anchor=center},
            },
            where level=1{text width=12em, font=\normalsize, minimum height=3.5em}{},
            where level=2{text width=13em, font=\normalsize, minimum height=3.5em}{},
            where level=3{
                text width=40em,
                font=\small,
                align=left,
                fill=gray!10,
                text=black,
                inner xsep=6pt,
                inner ysep=6pt,
            }{},
            [
                {\Large \textbf{Taxonomy of Diffusion Language Models}}, ver, line width=0.7mm
                [
                    {\large \shortstack{\textbf{Paradigms} \\ (\S~\ref{sec:paradigms})}}, fill=c1!60, draw=c1, line width=0.5mm
                    [
                        {\textbf{Continuous Space Models}}, fill=c1!60, draw=c1
                        [{
                            Diffusion-LM~\cite{li2022diffusion},
                            SED~\cite{strudel2022self},
                            LATENTOPS~\cite{liu2022composable},
                            Diffuseq~\cite{gongdiffuseq},
                            CDCD~\cite{dieleman2022continuous},\\
                            Difformer~\cite{gao2022empowering},
                            LD4LG~\cite{lovelace2023latent},
                            GENIE~\cite{lin2023text},
                            InfoDiffusion~\cite{wang2023infodiffusion},
                            EDDPMs~\cite{liu2024unified},\\
                            SMOOTHIE~\cite{shabalin2025smoothie},
                            TESS~\cite{mahabadi2024tess},
                            TESS 2~\cite{tae2025tess},
                            LDEBM~\cite{yu2022latent},\\
                            LangFlow~\cite{chen2026langflow},
                            ELF~\cite{hu2026elf},
                            BitLM~\cite{zhuang2026bitlm},
                            FMLM~\cite{lee2026flow},
                            Cola-DLM~\cite{guo2026continuous}
                            }], 
                    ]
                    [
                        {\textbf{Discrete Space Models}}, fill=c1!60, draw=c1, align=center
                        [{
                            D3PM~\cite{austin2021structured}, 
                            DiffusionBERT~\cite{he2023diffusionbert},
                            LLaDA~\cite{nie2025largelanguagediffusionmodels},
                            RDMs~\cite{zhengreparameterized},\\
                            MD4~\cite{shi2024simplified},
                            MDLM~\cite{sahoo2024simple},
                            Diffusion-LLM~\cite{ye2023diffusion},
                            Diffusion-NAT~\cite{zhou2024diffusion},\\
                            Plaid~\cite{gulrajani2023likelihood},
                            SEDD~\cite{lou2024discrete},
                            RADD~\cite{ou2024your},
                            DFM~\cite{gat2024discrete},\\
                            DDPD~\cite{liu2024think},
                            MGDM~\cite{ye2024beyond},
                            Diffu-LLaMA~\cite{gongscaling},
                            Dream-7B~\cite{dream2025},\\
                            GIDD~\cite{von2025generalized},
                            LongLLaDA~\cite{liu2025longllada},
                            Seed Diffusion~\cite{song2025seed},
                            LLaDA-MoE~\cite{zhu2025lladamoe},
                            LLaDA2.0~\cite{bie2025llada2}
                            }]
                    ]
                    [
                        {\textbf{Hybrid DLMs}}, fill=c1!60, draw=c1, align=center
                        [{
                            SSD-LM~\cite{han2023ssd}, AR-DIFFUSION~\cite{wu2023ar}, BD3-LM~\cite{arriolablock}, CtrlDiff~\cite{huang2025ctrldiff}, SpecDiff~\cite{christopher2025speculative},
                            SDAR~\cite{cheng2025sdar}, \\
                            TiDAR~\cite{liu2025tidar},
                            SDLM~\cite{liu2025sequential},
                            NBDiff~\cite{tian2025next},\\
                            Efficient-DLM~\cite{fu2025efficient},
                            I-DLM~\cite{yu2026introspective},
                            ReFusion~\cite{li2025refusion},
                            CARD~\cite{ruan2026causal}}]
                    ]
                ]
                [
                    {\large \shortstack{\textbf{Training Strategies} \\ (\S~\ref{sec:training})}}, fill=c2!60, draw=c2, line width=0.5mm
                    [
                        {\textbf{Pre-training}}, fill=c2!60, draw=c2
                        [{
                            \textbf{From scratch:} LLaDA-8B~\cite{nie2025largelanguagediffusionmodels} \\
                            \textbf{Adapting from AR models:} Dream~\cite{dream2025}, DiffuLLaMA~\cite{gongscaling}\\
                            \textbf{Adapting from image diffusion models:} D-DiT~\cite{li2025dual}, Muddit~\cite{shi2025muddit}
                            }]
                    ]
                    [
                        {\textbf{Post-training}}, fill=c2!60, draw=c2
                        [{
                            DoT~\cite{ye2024diffusion}, DCoLT~\cite{huang2025reinforcing}\\
                            \textbf{Policy Gradient:}
                            diffu-GRPO~\cite{zhao2025d1}, 
                            UniGRPO~\cite{yang2025mmada}, 
                            SEPO~\cite{zekri2025fine},
                            Coupled-GRPO~\cite{gong2025diffucoder},\\
                            wd1~\cite{tang2025wd1},
                            IGPO~\cite{zhao2025inpainting},
                            SPG~\cite{wang2025spg},
                            SAPO~\cite{xie2025step},
                            BGPO~\cite{lin2025boundary},
                            JustGRPO~\cite{ni2026flexibility}\\
                            \textbf{Preference Optimization:} VRPO~\cite{zhu2025llada}
                            }]
                    ]
                ]
                [
                    {\large \shortstack{\textbf{Inference \&} \\ \textbf{Optimization} \\ (\S~\ref{sec:inference})}}, fill=c3!60, draw=c3, line width=0.5mm
                    [
                        {\textbf{Parallel Decoding}}, fill=c3!60, draw=c3
                        [{
                            % \textbf{E.g.,}
                            Fast-dLLM~\cite{wu2025fast}, APD~\cite{israel2025accelerating}, SlowFast Sampling~\cite{wei2025accelerating}, SpecDiff~\cite{christopher2025speculative}, Dimple~\cite{yu2025dimple}\\ 
                            Learn2PD~\cite{bao2025learning}, dParallel~\cite{chen2025dparallel},
                            DFlash~\cite{chen2026dflash}, DMax~\cite{chen2026dmax},
                            NAP~\cite{li2026diffusion}}]
                    ]
                    [
                        {\textbf{Unmasking/Remasking}}, fill=c3!60, draw=c3
                        [{
                            LLaDA~\cite{nie2025largelanguagediffusionmodels}, Dream~\cite{dream2025}, Masked DLM~\cite{sahoo2024simple}, Fast-dLLM~\cite{wu2025fast}, ReMDM~\cite{wang2025remasking}}]
                    ]
                    [
                        {\textbf{Guidance}}, fill=c3!60, draw=c3
                        [{
                            A-CFG~\cite{li2025adaptive}, Freecache~\cite{hu2025accelerating}, DINGO~\cite{suresh2025dingo}}] 
                    ]
                     [
                        {\textbf{Efficiency Techniques}}, fill=c3!60, draw=c3
                        [{
                           Key-Value Cache~\cite{wu2025fast, liu2025dllm, ma2025dkv, arriolablock, nguyen2025attention,jiang2025d};\\
                           Feature Cache~\cite{liu2025dllm, hu2025accelerating, ma2024deepcache, chen2024delta, ma2024learning, lvfastercache};\\
                           Distillation~\cite{hayakawadistillation, chen2025dlm, salimansprogressive,qian2026d3llm};
                           Pruning~\cite{myrzakhan2026sink};
                           }]
                    ]
                ]
                [
                    {\large \shortstack{\textbf{Multimodal \&} \\ \textbf{Applications} \\ (\S~\ref{sec:multimodal}, \S~\ref{sec:application})}}, fill=c4!60, draw=c4, line width=0.5mm
                    [
                        {\textbf{Multimodal DLMs}}, fill=c4!60, draw=c4
                        [{
                            LLaDA-V~\cite{you2025llada}, 
                            Dimple~\cite{yu2025dimple}, 
                            MMaDA~\cite{yang2025mmada}, 
                            D-DiT~\cite{li2025dual},
                            LaViDa~\cite{li2025lavida},
                            Fudoki~\cite{wang2025fudoki}\\
                            Muddit~\cite{shi2025muddit},  
                            UniDisc~\cite{swerdlow2025unified},
                            Lumina-DiMOO~\cite{xin2025lumina},
                            LaViDa-O~\cite{li2025lavidao},\\
                            MMaDA-Parallel~\cite{tian2025mmada},
                            DiffusionVL~\cite{zeng2025diffusionvl},
                            Fast-dVLM~\cite{wu2026fast},
                            VidLaDA~\cite{he2026vidlada}
                            }]
                    ]
                    [
                        {\textbf{Conventional NLP Tasks}}, fill=c4!60, draw=c4
                        [{
                           ROIC-DM~\cite{yuan2024roic}, DiffusionNER~\cite{shen2023diffusionner},
                           IPAD~\cite{yang2025ipad},
                           DiffusionABSA~\cite{liu2024let},
                           DiffuSum~\cite{zhang2023diffusum}\\
                           TermDiffuSum~\cite{dong2025termdiffusum},  
                           Diff-KPE~\cite{luo2024enhancing},
                           IPED~\cite{zhao2024iped}, 
                           EdiText~\cite{lee2025editext},
                           DIFFUSEMP~\cite{bi2023diffusemp}\\
                           DiffuDetox~\cite{floto2023diffudetox}, 
                           ParaGuide~\cite{horvitz2024paraguide}, 
                           PLANNER~\cite{zhang2023planner}, 
                           DiffuCom~\cite{liu2023diffucom}, 
                           DiffusionDialog~\cite{xiang2024diffusiondialog}\\ 
                           LDP~\cite{zou2024improved}, 
                           PoetryDiffusion~\cite{hu2024poetrydiffusion},  
                           XDLM~\cite{chen2023xdlm}, 
                           DiffusionRet~\cite{qiao2023diffusionret},  
                           DIFND~\cite{yan2025debunk}}]
                    ]
                    [
                        {\textbf{Code Generation}}, fill=c4!60, draw=c4
                        [{
                            DUS~\cite{luxembourg2025plan}, DiffuCoder~\cite{gong2025diffucoder}, DCoLT~\cite{huang2025reinforcing},
                            Mercury Coder~\cite{labs2025mercury},\\
                            Stable-DiffCoder~\cite{fan2026stable},
                            DICE~\cite{bai2026dice}}]
                    ]
                    [
                        {\textbf{Computational Biology}}, fill=c4!60, draw=c4
                        [{
                           \textbf{Molecular :} 
                           TransDLM~\cite{xiong2024text},
                           TGM-DLM~\cite{gong2024text}\\
                           \textbf{Protein Design:} 
                           MeMDLM~\cite{goel2024memdlm}, 
                           DPLM~\cite{wangdiffusion},
                           CFP-GEN~\cite{yincfp}\\
                           DRAKES~\cite{wangfine},
                           ForceGen~\cite{ni2024forcegen},
                           DSM~\cite{hallee2025diffusion}, 
                           DPLM2~\cite{wang2024dplm}
                           }]
                    ]
                    [
                        {\textbf{Robotics}}, fill=c4!60, draw=c4
                        [{
                            LLaDA-VLA~\cite{wen2025llada},
                            dVLA~\cite{wen2025dvla},
                            UD-VLA~\cite{chen2025unified}
                           }]
                    ]
                ]
            ]
        \end{forest}
    }
    \caption{A taxonomy of Diffusion Language Models, covering foundations, training and inference strategies, and key applications. The section numbers (\S) correspond to the sections in this survey.}
    \label{fig:taxonomy}
\end{figure*}
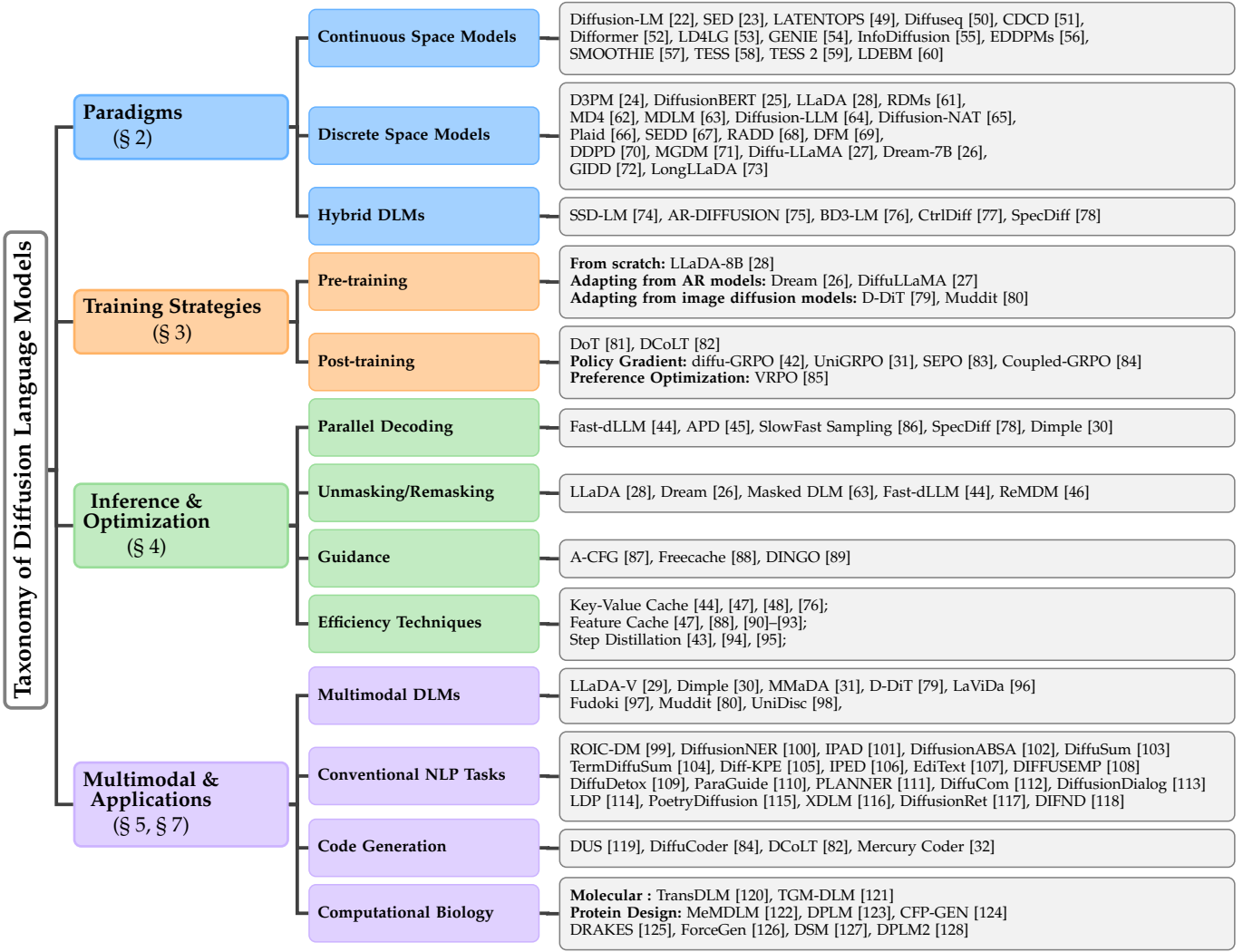

Despite the recent rise in popularity of DLMs, there remains a lack of a comprehensive survey that systematically covers the entire DLM ecosystem.
We structured our survey as follows: Section~\ref{sec:paradigms} provides a comprehensive overview of modern language modeling paradigms, including autoregressive, masked, and diffusion-based approaches. 
Section~\ref{sec:training} delves into the training methodologies for diffusion language models, covering both pre-training and subsequent fine-tuning techniques such as SFT and RL alignment.
Section~\ref{sec:inference} details various inference strategies and optimizations, focusing on techniques tailored for continuous and discrete space models.
Section~\ref{sec:multimodal} explores the extension of diffusion models to multimodal contexts, surveying state-of-the-art models and architectures like LLaDA-V~\cite{you2025llada}, MMaDA~\cite{yang2025mmada}, and Dimple~\cite{yu2025dimple}. 
Section~\ref{sec:performance} presents and visualizes performance comparisons of DLMs.
Section~\ref{sec:application} showcases the diverse applications of DLMs in tasks ranging from text and code generation to computational biology.
Section~\ref{sec:challenges} highlights the challenges and limitations of diffusion language models, including issues of efficiency, reasoning, agent capability, and infrastructure, and also outlines promising directions for future research. 
To provide a consolidated overview, a taxonomy of DLMs is presented in Fig.~\ref{fig:taxonomy}.

\section{Paradigms of Diffusion Language Models}
\begin{figure*}[!t] 
    \centering
    \includegraphics[width=\textwidth]{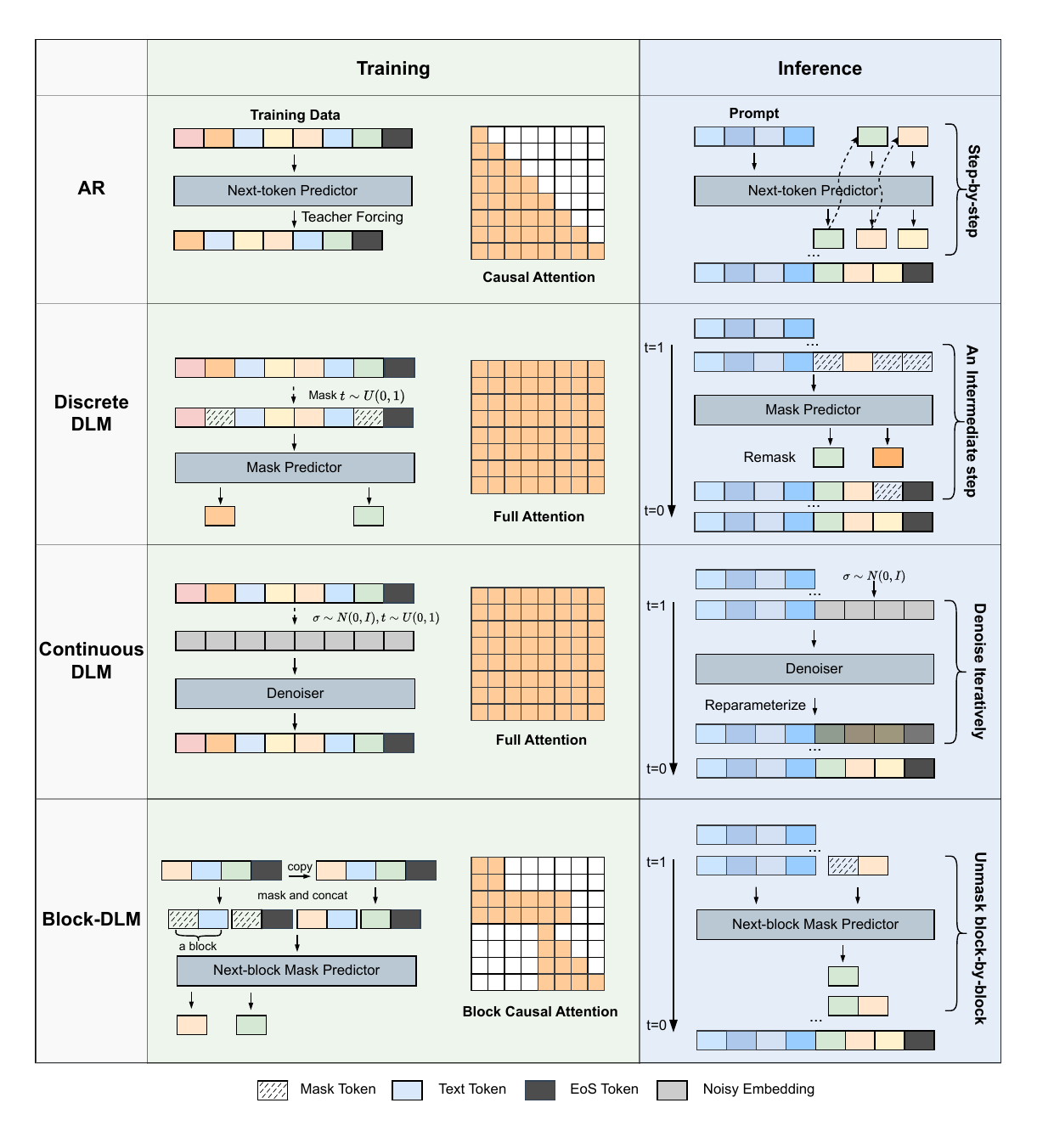}
    \caption{An overview of training and inference procedures across different paradigms of Diffusion Language Models, with autoregressive (AR) models included for comparison. AR models are trained using teacher forcing and causal attention, whereas both discrete and continuous DLMs employ fully bidirectional attention mechanisms. Block-wise diffusion models, exemplified by BD3-LM~\cite{arriolablock}, integrate autoregressive and diffusion strategies, and are trained using a specially designed block-causal attention mask.
    }

    \label{fig:paradigm}
\end{figure*}

% Tianyi
\label{sec:paradigms}
Diffusion Language Models have emerged as a powerful non-autoregressive paradigm that balances generative quality with inference parallelism. 
Inspired by principles from non-equilibrium thermodynamics~\cite{sohl2015deep}, DLMs learn to reverse a gradual noising process. 
This iterative refinement approach allows for parallel generation of the entire sequence, offering a potential solution to the inference bottleneck of AR models. 
DLMs can be broadly categorized based on the space in which the diffusion process operates: either continuous or discrete. Additionally, there are hybrid AR-Diffusion models that combine autoregressive and diffusion in various forms, aiming to leverage the complementary strengths of both paradigms. We present model information from several works in Table~\ref{tab:diffusion_models_summary} and provide a comparison of different paradigms in Fig.~\ref{fig:paradigm}.

\subsection{Preliminaries of Modern Language Modeling}
The field of language modeling has evolved through several distinct paradigms, each characterized by unique architectural choices, training objectives, and associated trade-offs.  
In this subsection, we provide a brief overview of recent transformer-based paradigms at scale, highlighting their core principles, mathematical formulations, and representative models. 
Earlier approaches are not included, as we focus on modern, large-scale designs here.
This review serves to establish the conceptual foundation for understanding the emergence of diffusion language models as a novel and promising alternative that addresses key limitations of prior methods.

\subsubsection{Masked Language Models}

Masked Language Models (MLMs), popularized by BERT~\cite{devlin2019bert}, represent a foundational paradigm that scales pretrained language models using transformer-based encoder-only architectures.  
Conceptually simple yet empirically powerful, MLMs learn bidirectional contextual representations by predicting randomly masked tokens within an input sequence, leveraging both preceding and succeeding context.  
This approach follows a denoising autoencoding framework, where a subset of input tokens is masked, and the model is trained to reconstruct them:
\begin{equation}
\mathcal{L}_{\text{MLM}} = \mathbb{E}_{x \sim \mathcal{D}} \, \mathbb{E}_{\mathcal{M} \sim Mask(x)} \left[ - \sum_{i \in \mathcal{M}} \log P_\theta(x_i \mid x_{\setminus \mathcal{M}}) \right]
\end{equation}
Here, $x$ denotes the input sequence, $\mathcal{M}$ is the set of masked positions, and $x_{\setminus \mathcal{M}}$ represents the visible (unmasked) context.  
BERT also introduces a next sentence prediction (NSP) objective to model inter-sentence relationships:  
\begin{equation}
\mathcal{L}_{\text{NSP}} = \mathbb{E}_{(A,B,y) \sim \mathcal{D}} \left[ - \log P_\theta(y \mid A, B) \right]
\end{equation}
where $(A, B)$ is a pair of text segments, and $y \in \{0,1\}$ indicates whether $B$ follows $A$ in the original text.  

BERT's effectiveness in language understanding tasks such as sentiment analysis, named entity recognition, and question answering has inspired numerous improved variants. 
For instance, RoBERTa~\cite{liu2019roberta} removes the NSP objective and adopts more aggressive training strategies, while ALBERT~\cite{lanalbert} introduces parameter sharing and matrix factorization for efficiency. 
DeBERTa~\cite{hedeberta} further enhances contextual encoding with disentangled attention and improved decoding mechanisms for masked token prediction.

Despite their strengths in understanding tasks, MLMs are not inherently designed for generative tasks, generating text requires specialized fine-tuning strategies or decoding schemes, making them unsuitable for open-ended generation without significant architectural modifications.

\subsubsection{Autoregressive Language Models}
Illustrated by GPT series~\cite{radford2018improving, radford2019language, brown2020language, achiam2023gpt} and Transformer-XL~\cite{dai2019transformer}, further advanced by subsequent LLMs~\cite{chowdhery2023palm,touvron2023llama,bai2023qwen,zhang2022opt}, autoregressive language models have become the backbone of modern generative AI, characterized by their unidirectional, left-to-right token generation process. 
Unlike bidirectional models, Autoregressive LMs factorize the joint probability of a text sequence into a product of conditional probabilities:
\begin{equation}
P(x) = \prod_{i=1}^{n} P_\theta(x_i \mid x_1, x_2, \dots, x_{i-1})
\end{equation}

Given a token sequence $X = (x_1, x_2, \dots, x_n)$, the training objective is to maximize the log-likelihood of the sequence under this factorization:
\begin{equation}
\mathcal{L}_{\text{AR}} = \mathbb{E}_{X \sim \mathcal{D}} \left[ - \sum_{i=1}^{n} \log P_\theta(x_i \mid x_1, \dots, x_{i-1}) \right]
\end{equation}
This is typically implemented using a decoder-only Transformer architecture with causal attention masking and teacher forcing during training, ensuring that each token prediction is conditioned only on preceding tokens while enabling parallel computation of the loss.

The sequential generation formulation is both a strength and a limitation. 
On one hand, it aligns with text generation tasks and facilitates straightforward sampling, naturally suits various applications.
On the other hand, it imposes a fundamental bottleneck on inference speed, as token generation is inherently sequential and cannot be parallelized.
This trade-off between generation quality and latency has become a central challenge in advancing AR models.
Beyond the standard next-token prediction (NTP), recent research has explored multi-token prediction (MTP)~\cite{gloecklebetter, liu2024deepseek} to accelerate inference by generating multiple tokens per step.  
These efforts share conceptual similarities with parallel decoding strategies employed in DLMs, while some other works are directly inspired by diffusion process to align LLMs~\cite{chen2025diffpo}.

\subsubsection{Other Paradigms}
\noindent\textbf{Sequence-to-Sequence Models.}
Sequence-to-sequence (Seq2Seq) models~\cite{sutskever2014sequence}, an early yet powerful paradigm, are built on an encoder-decoder architecture and serve as a versatile framework for conditional text generation tasks such as machine translation and summarization.  
Modern models like T5~\cite{raffel2020exploring} and BART~\cite{lewis2020bart} are prominent examples.

In this architecture, the encoder processes the source sequence to produce an intermediate representation, which the decoder then uses to generate the target sequence, typically in an autoregressive manner.  
While standard Seq2Seq decoders are autoregressive, the framework itself is highly flexible.  
Many DLMs, such as DiffuSeq~\cite{gongdiffuseq} and SeqDiffuSeq~\cite{yuan2022seqdiffuseq}, adapt this architecture by replacing the autoregressive decoder with a non-autoregressive diffusion decoder, leveraging the encoder's strong conditioning ability to guide the denoising process in generation.

\noindent\textbf{Permutation Language Models.}
Permutation Language Models (PLM), exemplified by XLNet~\cite{yang2019xlnet}, offer an alternative approach to incorporating bidirectional context within a generative framework.  
PLMs are trained to predict tokens in a sequence, but in a random, permuted order rather than a fixed left-to-right order. 
The objective is to maximize the expected log-likelihood over all possible permutations of the factorization order:
\begin{equation}
\mathcal{L}_{\text{PLM}} = \mathbb{E}_{\mathbf{z} \sim \mathcal{Z}_T} \left[- \sum_{t=1}^{T} \log P_\theta(x_{z_t} | \mathbf{x}_{\mathbf{z}_{<t}}) \right]
\end{equation}
where $\mathcal{Z}_T$ denotes the set of all possible permutations of a sequence of length $T$, and $\mathbf{z}_t$, $\mathbf{z}_{<t}$ refer to the $t$-th and first $t-1$ elements of a given permutation $\mathbf{z} \in \mathcal{Z}_T$.
This formulation allows the model to capture bidirectional context for each token, combining the advantages of bidirectional context (like MLMs) with a coherent autoregressive generation process. 
This contrasts with DLMs, which achieve bidirectionality through a parallel iterative refinement process.

\subsection{Continuous Diffusion Language Models}
Continuous-space DLMs model language by first mapping discrete tokens into a continuous embedding space. 
A diffusion process then models the data distribution in this continuous space~\cite{li2022diffusion, strudel2022self}.
Typically, diffusion models define a generative process by learning to reverse a predefined corruption process that gradually transforms data into noise. 
This process consists of a \textbf{forward (noising) process} and a \textbf{reverse (denoising) process}.
The forward process gradually transforms a data sample \( \mathbf{x}_0 \) into noise over \( T \) timesteps via a fixed Markov chain:
\begin{equation}
q(\mathbf{x}_{1:T} \mid \mathbf{x}_0) = \prod_{t=1}^T q(\mathbf{x}_t \mid \mathbf{x}_{t-1})
\end{equation}
\begin{equation}
\quad q(\mathbf{x}_t \mid \mathbf{x}_{t-1}) = \mathcal{N}(\mathbf{x}_t; \mu_t(\mathbf{x}_{t-1}), \Sigma_t),
\end{equation}
where \( \mu_t \) and \( \Sigma_t \) define the noise schedule. 
In many implementations, such as DDPM~\cite{ho2020denoising} and Rectified Flow~\cite{liuflow}, the marginal distribution at each timestep has a closed-form expression:
\begin{equation}
\mathbf{x}_t = \alpha_t \mathbf{x}_0 + b_t \mathbf{\epsilon}, \quad \mathbf{\epsilon} \sim \mathcal{N}(0, \mathbf{I}),
\end{equation}
where \( \alpha_t \) and \( b_t \) are deterministic functions of time \( t \).

The reverse process learns to invert the corruption, starting from noise \( \mathbf{x}_T \sim \mathcal{N}(0, \mathbf{I}) \) and gradually denoising to recover a sample close to \( \mathbf{x}_0 \). 
This is parameterized by a neural network \( f_\theta(\mathbf{x}_t, t) \), typically implemented as a Transformer, which predicts a target quantity \( \mathbf{z} \) associated with the forward process (e.g., clean data, noise, or velocity). 
A common training objective takes the form:
\begin{equation}
\mathcal{L}_{\text{simple}} = \mathbb{E}_{t, \mathbf{x}_0, \mathbf{z}} \left[ \left\| f_\theta(\mathbf{x}_t, t) - \mathbf{z} \right\|^2 \right],
\end{equation}
where \( \mathbf{x}_t \) is sampled via the forward process given \( \mathbf{x}_0 \), 
and \( \mathbf{z} \) is the corresponding regression target derived from \( \mathbf{x}_0 \) and \( t \).

After training, generation proceeds by sampling from the learned reverse process, starting from noise \(\mathbf{x}_T \sim \mathcal{N}(\mathbf{0}, \mathbf{I})\). 
At each timestep \(t = T, T-1, \ldots, 1\), the model defines a conditional distribution $p_\theta(\mathbf{x}_{t-1} \mid \mathbf{x}_t)$ which aims to approximate the true reverse transition \(q(\mathbf{x}_{t-1} \mid \mathbf{x}_t)\). 
Sampling iteratively from these learned conditionals produces progressively less noisy latent states until an estimate of the original data \(\mathbf{x}_0\) is recovered. 
After generating a denoised embedding \( \hat{\mathbf{x}}_0 \), a \textbf{rounding step} maps it back to a discrete token. 
This is typically done by nearest-neighbor searching in the embedding space, using a decoder head, or thresholding techniques~\cite{chen2022analog}.

Diffusion-LM~\cite{li2022diffusion} first introduces a diffusion process in the embedding space to create a non-autoregressive language generation model.
By using a classifier-guidance mechanism similar to those in image diffusion models, it achieves highly controllable text generation and infilling.
LDEBM~\cite{yu2022latent} presents a novel symbiosis of latent space EBMs and diffusion models in a variational learning framework to address the learning issues of energy-based priors, with a focus on interpretable text modeling.
LATENTOPS~\cite{liu2022composable} proposes an efficient framework for composable text operations by working within a compact latent space. 
It introduces an efficient sampler based on ordinary differential equation (ODE) to generate latent vectors guided by arbitrary plug-in control operators, which are then decoded into the desired text.
Later, Diffuseq~\cite{gongdiffuseq}, a classifier-free DLM for sequence-to-sequence tasks is proposed, which corrupts only the target sequence embeddings in the forward process to achieve strong and diverse conditional text generation.
The Self-conditioned Embedding Diffusion (SED)~\cite{strudel2022self} framework conducts diffusion directly on a fixed, continuous token embedding space. By incorporating a self-conditioning mechanism, it achieves strong performance in both conditional and unconditional text generation, rivaling standard autoregressive models.
CDCD~\cite{dieleman2022continuous} applies continuous diffusion to categorical data by embedding tokens into a continuous space.
It proposes score interpolation, which uniquely allows the model to be trained with a cross-entropy loss, and time warping, an adaptive strategy to efficiently schedule noise levels during training. 
To address optimization challenges in the embedding space, Difformer~\cite{gao2022empowering} introduces an anchor loss to prevent embedding collapse and a noise rescaling framework to mitigate model degeneration. 
LD4LG~\cite{lovelace2023latent} leverages a pretrained language model as a powerful autoencoder to create a compact latent space, where a continuous diffusion model is then trained for high-quality text generation. 
GENIE~\cite{lin2023text} proposes a large-scale pre-training framework for diffusion language models, introducing a novel continuous paragraph denoise objective to effectively learn from large corpora by reconstructing corrupted text paragraphs.
InfoDiffusion~\cite{wang2023infodiffusion} introduces an information entropy-aware noise schedule to guide the model toward a more human-like ``key-info-first'' process that prioritizes generating core content.
EDDPMs~\cite{liu2024unified} unify generation, reconstruction, and representation by generalizing the diffusion process with a parameterized encoder-decoder, enabling stable, joint training of all components within a single framework. 
SMOOTHIE~\cite{shabalin2025smoothie} proposes a novel diffusion process that progressively smooths token embeddings based on semantic similarity, combining the advantages of continuous latent spaces and discrete token handling.

Continuous diffusion processes can also be formulated in the logit space rather than the embedding space. 
TESS~\cite{mahabadi2024tess} introduces a fully non-autoregressive framework that diffuses over a k-logit simplex representation of tokens and employs a novel self-conditioning mechanism tailored to this setting. 
Extending this, TESS 2~\cite{tae2025tess} scales the approach by adapting pretrained large autoregressive models into general-purpose diffusion language models through a diffusion-specific pretraining recipe and instruction tuning, enabling strong instruction-following capabilities. 

Recent work revisits continuous-space DLMs with more explicit choices of representation, transport path, and decoding interface. LangFlow~\cite{chen2026langflow} connects embedding-space DLMs with Flow Matching via Bregman divergence and studies continuous flow-based language modeling with an ODE-based likelihood bound, information-uniform noise scheduling, and self-conditioning. ELF~\cite{hu2026elf} formulates language modeling as continuous-time flow matching in embedding space, keeping the denoising trajectory continuous until a final mapping back to discrete tokens. Cola-DLM~\cite{guo2026continuous} instead uses a hierarchical latent formulation: a Text VAE maps text into continuous latents, a block-causal DiT models a global semantic prior, and a conditional decoder realizes text from that latent representation. BitLM~\cite{zhuang2026bitlm} replaces vocabulary-level token prediction with bitwise denoising over fixed-length binary token codes while preserving causal attention across blocks.

\begin{table*}[ht]
  % bowei
  \centering
  \caption{Summary of diffusion language models, configurations, and their design choices.}
  \begin{tabular}{l|cccccc}
    \toprule
    \textbf{Model} & \textbf{Parameters} & \textbf{Diffusion type}  &  \textbf{Task} & \textbf{Training data}\\
    \midrule
    D3PM~\cite{austin2021structured} & 70M & Discrete   & Language  & 65B tokens\\
    Diffusion-LM~\cite{li2022diffusion}  & 100M \& 300M & Continuous  & Language & -- \\
    Diffuseq~\cite{gongdiffuseq}  & 91M & Continuous  & Language & 565K sentence pairs \\
    SSD-LM~\cite{han2023ssd} & 400M & Continuous  & Language & 123B tokens \\
    DiffusionBERT~\cite{he2023diffusionbert}& 110M & Discrete    & Language & 16B tokens \\
    CDCD~\cite{dieleman2022continuous} & 1.3B & Continuous  & Language  & 315B tokens \\
    LD4LG~\cite{lovelace2023latent}               & 188M & Continuous             & Language & 5.2M sentence pairs \\
    SeqDiffuSeq~\cite{yuan2022seqdiffuseq}   & 65M \& 110M& Continuous           & Language          & 45B tokens \\ 
    TESS~\cite{mahabadi2024tess}   & 125M \& 355M & Continuous           & Language & --\\ 
    MDLM~\cite{sahoo2024simple} & 110M & Discrete  & Language & 622B tokens\\
    DFM~\cite{gat2024discrete} & 1.7B & Discrete & Language \& Code & 2.5T tokens\\ 
    TESS-2~\cite{tae2025tess} & 7B & Continuous  & Language & 360B tokens\\
    LLaDA~\cite{nie2025largelanguagediffusionmodels} & 1B \& 8B & Discrete & Language \& Code & 2.3T tokens\\
    Mercury~\cite{labs2025mercury} & -- & Discrete  & Code & Trillions of tokens\\
    LLaDA-1.5~\cite{zhu2025llada} & 8B & Discrete  & Language & 2.3T tokens\\
    MMaDA~\cite{yang2025mmada} & 8B & Discrete  & Multimodal Unified & 900B image-text tokens \\
    Dream~\cite{dream2025} & 7B & Discrete  & Language \& Code & 580B tokens \\
    LLaDA-V~\cite{you2025llada} & 8.4B & Discrete  & Multimodal  & 3M image-text samples \\
    LaViDa~\cite{li2025lavida} & 8.4B & Discrete  & Multimodal & 1.6M image-text samples \\
    Dimple~\cite{yu2025dimple} & 7B & Discrete  & Multimodal & 0.8B tokens\\
    LongLLaDA~\cite{liu2025longllada} & 8B & Discrete  & Language \& Code & 2.3T tokens \\
    DiffuCoder~\cite{gong2025diffucoder} & 7B & Discrete  & Code & 130B tokens \\
    LaViDa-O~\cite{li2025lavidao} & 10.4B & Discrete  & Multimodal Unified & \(>\) 200M image-text pairs \\
    Lumina-DiMOO~\cite{xin2025lumina} & 8B & Discrete & Multimodal Unified & \(>\) 110M image-text pairs \\
    LLaDA-MoE~\cite{zhu2025lladamoe} & 7B (1.4B activated) & Discrete & Language & 20T tokens \\
    SDAR~\cite{cheng2025sdar} & 1.7B-30B series & Discrete & Language & 54B tokens \\
    TiDAR~\cite{liu2025tidar} & 1.5B \& 8B & Discrete & Language & 150B tokens \\
    SDLM~\cite{liu2025sequential} & 3B \& 32B & Discrete & Language & 2.3B tokens \\
    LLaDA2.0~\cite{bie2025llada2} & 16B \& 100B & Discrete & Language & -- \\
    Stable-DiffCoder~\cite{fan2026stable} & 8B & Discrete & Code & 1.3T tokens \\
    Efficient-DLM~\cite{fu2025efficient} & 1.5B, 4B \& 8B & Discrete & Language & 300B-500B tokens \\
    DiffusionVL~\cite{zeng2025diffusionvl} & 3B \& 7B & Multimodal & Vision-language & 738K instruction samples \\
    VidLaDA~\cite{he2026vidlada} & 8B & Multimodal & Video understanding & 2.8M samples \\
    LangFlow~\cite{chen2026langflow} & 130M & Continuous & Language & -- \\
    ELF~\cite{hu2026elf} & 105M, 342M \& 652M & Continuous & Language & 45.2B tokens \\
    BitLM~\cite{zhuang2026bitlm} & 0.6B, 1.7B, 4B \& 8B & Continuous & Language & FineWeb-350B tokens \\

    \bottomrule
  \end{tabular}
  \label{tab:diffusion_models_summary}
\end{table*}

\subsection{Discrete Diffusion Language Models}
Discrete space DLMs define the diffusion process directly on the vocabulary of tokens, avoiding the need for a continuous embedding space during the diffusion itself.
% D3PM
D3PM~\cite{austin2021structured} first illustrates this by introducing a structured diffusion process over discrete tokens.
The \textbf{forward process} corrupts a sequence by applying a transition matrix $\mathbf{Q}_t$ at each step. 
This matrix defines the probability of a token transitioning to any other token in the vocabulary. 
The probability of a state $\mathbf{x}_t$ given an initial state $\mathbf{x}_0$ is given by a categorical distribution:
$$
q(\mathbf{x}_t|\mathbf{x}_0) = \text{Cat}(\mathbf{x}_t; \mathbf{p} = \mathbf{x}_0 \bar{\mathbf{Q}}_t), \quad \text{where} \quad \bar{\mathbf{Q}}_t = \prod_{i=1}^t \mathbf{Q}_i
$$
A common choice for $\mathbf{Q}_t$ is an absorbing state transition, where each token has a probability of either remaining unchanged or transitioning to a special `[MASK]` token.
The \textbf{reverse process} learns to reverse these transitions, predicting the probability distribution of the original tokens given the corrupted sequence. 

% LLaDA
Over time, masked DLMs have emerged as a modern and highly effective evolution of discrete diffusion language models, forming the foundation for several recent large-scale efforts~\cite{nie2025largelanguagediffusionmodels, gongscaling}. 
We take LLaDA~\cite{nie2025largelanguagediffusionmodels}, the most representative model of this kind as an example.
Inspired by earlier work on reparameterized and simplified training objectives~\cite{zhengreparameterized,shi2024simplified,ou2024your}, 
LLaDA is trained from scratch using a cross-entropy loss that is computed only over masked tokens:
\begin{equation}
\label{eq:objective}
   \mathcal{L}(\theta)  \triangleq   -  \mathbb{E}_{t, x_0,  x_t} \left[\frac{1}{t} \sum_{ i = 1 }^L \textbf{1}[x_t^i = \textrm{M}] \log p_{\theta}(x_0^i|x_t) \right] , 
\end{equation}
where \( x_0 \) is sampled from the training corpus, 
\( t \) is sampled uniformly from \( [0, 1] \), 
and \( x_t \) is obtained by corrupting \( x_0 \) through the forward process.
The indicator function \( \textbf{1}[\cdot] \) ensures that the loss is applied only to positions that have been masked.
% LLaDA inference
During inference, the generation process starts with a fully masked sequence of desired length. 
In each iterative step, the model takes the current sequence (containing a mix of generated tokens and `[MASK]` tokens) and predicts a complete sequence of tokens. 
Based on the model's prediction confidence and noise schedule, a certain number of the highest-confidence predictions are unmasked and fixed, while the remaining positions are re-masked. 
This refinement process continues iteratively until all `[MASK]` tokens are resolved.
This approach elegantly combines the bidirectional context of MLMs with a controllable, parallel generation process.
LLaDA-8B, in particular, exhibits strong scalability and instruction-following ability, achieving performance on par with powerful autoregressive models such as LLaMA3-8B. 
This challenges the long-standing dominance of autoregressive models in large-scale language generation.

% DiffusionBERT
DiffusionBERT~\cite{he2023diffusionbert} combines a pre-trained BERT with a discrete diffusion process, leveraging its powerful denoising capabilities to learn the reverse process from a masked state. 
The model is further enhanced by a novel spindle noise schedule that considers token informativeness, achieving significant improvements in generation quality compared with previous DLMs.
% RDM
A different approach, Reparameterized Discrete diffusion Models (RDMs)~\cite{zhengreparameterized}, establishes an alternative formulation for the reverse process, which simplifies the training objective to a weighted cross-entropy loss.
This enables more flexible and adaptive decoding strategies, leading to significant performance gains over previous discrete diffusion models.
% MD4
Similarly, MD4~\cite{shi2024simplified} derives a simple weighted integral of cross-entropy losses as the continuous-time variational objective of masked diffusion models, providing a simple and generalized framework for training DLMs.
% MDLM
Another analogous approach is MDLM~\cite{sahoo2024simple}, which introduces a simplified, Rao-Blackwellized objective that takes the form of a weighted average of masked language modeling losses.
% Diffusion-LLM
Diffusion-LLM~\cite{ye2023diffusion} demonstrates the scalability of DLMs by adapting pre-trained masked language models to diffusion paradigm and further task-specific finetuning and instruction finetuning, unlocking their versatility in solving general language tasks.
% DiffusionNAT
Diffusion-NAT~\cite{zhou2024diffusion} unifies a discrete diffusion model with a PLM by reformulating the denoising process as a non-autoregressive masked token recovery task, allowing BART to act as an effective denoiser.
% Plaid
Plaid~\cite{gulrajani2023likelihood} is the first diffusion language model trained to maximize data likelihood, demonstrating through scaling laws that it can outperform autoregressive models like GPT-2 on standard benchmarks.
% SEDD
To improve the training objective, SEDD~\cite{lou2024discrete} introduces a score entropy loss to directly learn the ratios of the data distribution, which serves as a discrete extension of score matching.
% RADD
Reparameterized Absorbing Discrete Diffusion (RADD)~\cite{ou2024your} reveals that the concrete score in absorbing diffusion can be expressed as a time-independent conditional probability of the clean data, multiplied by an analytic, time-dependent scalar. It also formally unifies the training objectives of absorbing discrete diffusion and any-order autoregressive models.
% DFM
Discrete Flow Matching (DFM)~\cite{gat2024discrete} introduces a novel generative paradigm for discrete data that is analogous to continuous Flow Matching.
The method learns a generating probability velocity to transform samples along a general family of probability paths from a source to a target distribution. By scaling the model architecture, 
DFM significantly closes the performance gap with autoregressive models on various benchmarks.
% DDPD
DDPD~\cite{liu2024think} presents a framework that decouples the generation process into two specialized models: a planner and a denoiser.
At each step, the planner identifies the most corrupted token positions needing refinement, after which the denoiser predicts their values.
% MGDM
To improve performance on complex reasoning tasks, MGDM~\cite{ye2024beyond} is introduced to address the problem of subgoal imbalance.
This approach enhances discrete diffusion by prioritizing more difficult subgoals during the learning process through a token-level reweighting mechanism.
% Diffu-LLaMA
To address the challenge of scaling, a continual pre-training approach~\cite{gongscaling} is proposed to adapt existing autoregressive models, such as LLaMA, into diffusion language models.
The resulting models, named DiffuGPT and DiffuLLaMA, are competitive with their AR counterparts while gaining diffusion-native capabilities like flexible infilling.
% Dream
Building on this observation, Dream-7B~\cite{dream2025} is initialized from Qwen2.5 7B~\cite{qwen2.5} and further trained with 580B tokens, largely outperforming existing DLMs and matching the performance of top-tier AR models.
% GIDD
GIDD~\cite{von2025generalized} is introduced to overcome the limitation that masked diffusion models cannot revise generated tokens.
This framework generalizes the noising process by combining masking with uniform noise, which unlocks the model's ability to self-correct mistakes and improves sample quality.
% LongLLaDA
Recently, to address long-context capabilities, LongLLaDA~\cite{liu2025longllada} provides the first systematic analysis of DLMs in this domain.
It reveals that DLMs can maintain stable perplexity during direct context extrapolation and have better retrieval capabilities. 
LongLLaDA also introduces a training-free NTK-based RoPE extrapolation method, which significantly improves the extrapolation performance of DLMs,
validating that established extrapolation scaling laws remain effective for DLMs.
UltraLLaDA~\cite{he2025ultrallada} extends this line of work by introducing a diffusion-aware NTK RoPE scaling and lightweight long-context post-training, enabling diffusion LLMs to reach 128K context windows and achieving substantially better retrieval and perplexity performance than training-free extrapolation methods.
% LLaDA-MoE
LLaDA-MoE~\cite{zhu2025lladamoe} is the first work to integrate a sparse Mixture-of-Experts (MoE) architecture into diffusion language models, training a new MoE-based DLM from scratch on 20T tokens. 
Despite activating only about 1.4B parameters during inference, it surpasses larger dense diffusion models and achieves performance comparable to Qwen2.5-3B-Instruct across knowledge, coding, and reasoning benchmarks.

\subsection{Hybrid AR-Diffusion Language Models}
Hybrid AR-Diffusion models aim to strike a balance between the full parallelism of non-autoregressive models and the strong causal dependency modeling of autoregressive models. 
A prominent strategy for hybrid AR-diffusion modeling adopts a block-wise semi-autoregressive generation process.
In this setting, the model generates blocks of tokens autoregressively, while the tokens within each block are generated in parallel using a diffusion-like iterative process.
Early efforts such as SSD-LM~\cite{han2023ssd} pioneered hybrid approaches by a block-wise continuous diffusion process on simplex representations, 
AR-DIFFUSION~\cite{wu2023ar} illustrates a multi-level diffusion process and achieves semi-autoregressive generation by adjusting timestep according token position.
Recent representative model BD3-LM~\cite{arriolablock} further advances this direction on discrete models, demonstrating strong performance compared to pure AR and diffusion models.
CtrlDiff~\cite{huang2025ctrldiff} improves this paradigm by introducing dynamic block prediction techniques to enhance block-level efficiency and control.
SDAR~\cite{cheng2025sdar} further strengthens this hybrid paradigm by converting a pretrained autoregressive model into a blockwise diffusion model through a lightweight adaptation stage. 
It preserves AR-level performance while enabling efficient parallel intra-block generation, achieving scalable speedups without sacrificing quality.

The generation process in these models usually consists of two nested loops.  
In the outer loop, blocks of tokens are generated autoregressively, with each block conditioned on previously generated blocks.  
Within each block, the inner loop performs parallel token-wise generation through a diffusion-style iterative denoising process.  
In BD3-LM, the training objective is formalized as:
\begin{equation}
    \mathcal{L}_\text{BD}(\mathbf{x}, \theta) :=  - \sum_{b=1}^{B} \mathbb{E}_{t \sim [0, 1]} \mathbb{E}_{q} \frac{1}{t} \log p_\theta(\mathbf{x}^b | \mathbf{x}_{t}^b, \mathbf{x}^{<b})
\end{equation}

This hybrid strategy enables the model to capture long-range dependencies across blocks via autoregression, while simultaneously accelerating generation within each block through parallel diffusion.
The design also supports flexible output lengths and KV-Cache which is widely used in AR models~\cite{arriolablock}.

Notably, recent masked diffusion language models~\cite{nie2025largelanguagediffusionmodels, yang2025mmada} also adopt similar semi-autoregressive block-based decoding strategies, which can be seen as instances of hybrid AR-diffusion modeling.

Beyond block-based approaches that combine AR and diffusion at sequence level, hybridization can also occur at the architectural level, where
some part of the neural network, typically the encoder, diffuses the entire sequence altogether to an intermediate representation, then an autoregressive decoder generates the final sequence~\cite{zhu2024segment}. 
LADIDA~\cite{ladida} is a slightly different approach that diffuses at document level but decodes sentences by an AR decoder.
SpecDiff~\cite{christopher2025speculative} proposes a collaborative speculative decoding framework, where a lightweight diffusion model drafts candidate outputs, which are then validated and finalized by a large AR model.
TiDAR~\cite{liu2025tidar} proposes a sequence-level hybrid architecture that integrates diffusion-based parallel drafting and autoregressive sampling within a single forward pass through structured causal-bidirectional attention. 
It effectively unifies the efficiency of diffusion models with the quality of AR decoding, achieving up to $5\times$ throughput improvements while maintaining AR-level performance.
SDLM~\cite{liu2025sequential} introduces the Next Sequence Prediction (NSP) paradigm, which unifies next-token and next-block prediction to enable adaptive-length generation. 
By retrofitting pretrained autoregressive models with parallel block training and confidence-based dynamic decoding, SDLM achieves efficient diffusion-style intra-block generation while remaining KV-cache compatible.

\section{DLMs: Pre-training and Post-training}
% Tianyi
\label{sec:training}
\subsection{Pre-training and Supervised Fine-tuning}
The pretraining process of DLMs largely follows procedures similar to those used in autoregressive language models (for discrete DLMs) or image diffusion models (for continuous DLMs), with relatively fewer design spaces.
This section briefly summarizes existing approaches for DLM pretraining, aiming to bridge the methodological gap between DLMs and AR models.

To accelerate training, particularly for large-scale models, it is common practice to initialize DLMs from pretrained AR language models or image diffusion models~\cite{cetin2025large,gongscaling}.
DiffuGPT and DiffuLLaMA~\cite{gongscaling} try to initialize masked DLMs with open-source LLMs from 127M to 7B parameters, found that DLMs can be efficiently adapted from AR models, significantly reducing training time and cost while achieving comparable or even superior performance to their AR counterparts.
Building on this insight, Dream-7B is initialized from Qwen 2.5 7B~\cite{qwen2.5}, and is reported to outperform both LLaDA-8B and LLaMA3-8B on various benchmarks.
Some multimodal DLMs, on the other hand, are initialized from pretrained image diffusion models.
D-DiT~\cite{li2025dual} and Muddit~\cite{shi2025muddit} are initialized from pretrained MM-DiT backbones from SD3~\cite{esser2024scaling} and Meissonic~\cite{bai2024meissonic} respectively. 
Although these models are not originally designed for text generation, their latent representations contain intrinsic language-aligned knowledge, which can effectively facilitate the training of language modeling while retaining strong visual generation capabilities.

In terms of scaling properties, recent scaling-law analyses~\cite{ni2025training, ni2025diffusion} reveal that DLMs exhibit distinct compute-data tradeoffs from AR models: they are substantially more data-hungry under compute constraints, yet possess far greater data reuse potential under multi-epoch training, offering a principled foundation for designing optimal DLM training regimes.

Supervised fine-tuning in DLMs generally mirrors that of AR models. 
For masked DLMs like LLaDA~\cite{nie2025largelanguagediffusionmodels}, prompt tokens are left unmasked while response tokens are selectively masked, enabling the model to learn conditional response generation in a manner compatible with pre-training.
In continuous DLMs, SFT can also be performed by corrupting only the response segment, as demonstrated in TESS2~\cite{tae2025tess}.

Despite the overall similarity with AR training paradigms, DLMs face several unique challenges due to their diffusion-based formulation.
A major issue lies in the loss computation efficiency of masked DLMs.
In typical masked DLM training, only $\sim$50\% of tokens (on average) are involved in the loss computation, if timesteps are sampled uniformly.
This reduces data utilization and may lead to suboptimal gradients, particularly if critical answer tokens are excluded from the loss.
To address this, LaViDa~\cite{li2025lavida} proposes a complementary masking strategy: each training sample is duplicated with two disjoint masking patterns, ensuring that all tokens are included in the loss computation at least once.
Furthermore, due to the training-inference discrepancy, as illustrated in ~\cite{asada2025addressing}, the model performs significantly better during training than at inference time.
The authors propose a two-step diffusion process and an improved scheduling technique to mitigate this issue.

\subsection{Post-training for Reasoning Capabilities}
% llada-1.5, diffucoder, mmada, d1, DCOLT, SEPO, DOT ok
Exploration of reasoning capabilities is becoming increasingly popular in DLMs as their performance on language tasks improves. Typically, reasoning capabilities are gained through fine-tuning on reasoning datasets. 
For DLMs, this presents a unique and formidable challenge. 
Traditional Chain-of-Thought (CoT) methods are based on the sequential nature of AR models to reason step-by-step, but DLMs generate tokens in parallel.
The most successful post-training techniques in the AR domain, particularly those based on reinforcement learning (RL) and policy gradient methods, are built upon the ability to efficiently compute the log-probability of a generated sequence. 
This is straightforward in AR models due to their factorizable, sequential nature. 
In DLMs, where generation is an iterative, non-sequential process, the log-likelihood is intractable, creating a significant technical barrier to applying the mature suite of RL algorithms developed for AR models to DLMs.
Intuitively, we categorize these works into three main streams, which form the structure of this subsection:
(1) Parallelizing the reasoning chain, where CoT in AR models is adapted to DLMs in parallel generation.
(2) Adapting policy gradient methods, where variants of popular algorithms like GRPO are introduced to DLMs.
(3) Adapting preference optimization methods such as DPO to DLMs.

\subsubsection{DoT and DCoLT: Parallelizing the Reasoning Chain}

% DoT
One of the pioneering works to elicit complex reasoning in DLMs is Diffusion-of-Thought (DoT)~\cite{ye2024diffusion}, which adapts the popular Chain-of-Thought paradigm to the diffusion framework. 
Instead of generating reasoning steps sequentially like autoregressive models, DoT formulates them as intermediate thoughts that are refined in parallel throughout the diffusion denoising process.
The approach is implemented by fine-tuning pre-trained DLMs such as Plaid~\cite{gulrajani2023likelihood} and SEDD~\cite{lou2024discrete} on 
datasets containing problems and their corresponding step-by-step 
rationales. 
To enhance the model's ability to recover from its own mistakes, 
DoT introduces specialized training techniques like scheduled sampling and coupled sampling, which exposes the model to its own generated errors during training to improve its self-correction capabilities. 
This post-training methodology enables smaller DLMs to achieve impressive reasoning performance, even outperforming significantly larger autoregressive models on certain mathematical and logical reasoning benchmarks.

% DCoLT
A more recent approach, Diffusion Chain of Lateral Thought (DCoLT)~\cite{huang2025reinforcing}, introduces a distinct RL-based reasoning framework inspired by the cognitive concept of lateral thinking, which contrasts with the step-by-step vertical thinking of traditional CoT methods.
Instead of supervising intermediate steps, DCoLT treats each step of reverse diffusion process as a latent thinking action, but optimizes the entire multi-step denoising trajectory with outcome-based RL to maximize a reward on the final answer. When applied to masked DLMs like LLaDA, 
DCoLT innovatively introduces an Unmasking Policy Module (UPM), which learns the optimal order for revealing tokens as part of the RL action space.
This approach significantly boosts the reasoning capabilities of DLMs, with the DCoLT-reinforced LLaDA model achieving gains of +9.8\% on GSM8K and +19.5\% on HumanEval.

\begin{table*}[t!]
\centering
\caption{A brief summary of current post-training methods for DLMs' reasoning capabilities, focusing on their algorithm type, major goal, key technical innovations, and applicable model types. Notably, most of these methods are based on policy gradient, and are designed for discrete DLMs.}
\label{tab:summary_reasoning}
\resizebox{\textwidth}{!}{%
\begin{tabular}{@{}lllll@{}}
\toprule
\textbf{Method} & \textbf{Algorithm Type} & \textbf{Core Goal} & \textbf{Key Technical Innovation} & \textbf{Model Type} \\ \midrule
\textbf{DoT}~\cite{ye2024diffusion} & Non-RL Fine-tuning & Enable parallel Chain-of- & Converts serial CoT into a parallel diffusion & Continuous/  \\
&  & Thought reasoning & process; training-time self-correction & Discrete \\ \addlinespace

\textbf{DCoLT}~\cite{huang2025reinforcing} & Outcome-based RL & Enable non-linear latent  & Lateral thought; outcome-based RL; & Continuous/ \\
&  & reasoning & Unmask Policy Module & Discrete \\ \addlinespace

\textbf{SEPO}~\cite{zekri2025fine} & Policy Gradient & Finetune discrete DLMs with & Low-variance gradient estimator via score & Discrete \\
&  Framework(PPO/GRPO) & non-differentiable rewards & entropy \& importance sampling & \\ \addlinespace

\textbf{diffu-GRPO}~\cite{zhao2025d1} & Policy Gradient (GRPO) & Introduce policy gradient & Efficient one-step log-probability estimator & Discrete \\
& & method to DLMs & for applying GRPO to masked DLMs &  \\ \addlinespace

\textbf{coupled-GRPO}~\cite{gong2025diffucoder} & Policy Gradient (GRPO) & Reduce variance and maintain  & Coupled-sampling with complementary & Discrete \\
& & training efficiency & masks &  \\ \addlinespace

\textbf{UniGRPO}~\cite{yang2025mmada} & Policy Gradient (GRPO) & Unified reinforcement & Structured noising strategy; diversified  & Multimodal \\
& & learning & reward modeling &  Discrete \\ \addlinespace

\textbf{VRPO}~\cite{zhu2025llada} & Preference Optimization & Align with human preferences & Sample budget allocation; antithetic & Discrete \\ 
& (DPO) &  & sampling & \\ \addlinespace

\textbf{IGPO}~\cite{zhao2025inpainting} & Policy Gradient (GRPO) & Use inpaint ability in DLMs & Inpainting-guided sampling;  insert partial & Discrete \\
& & to guide exploration & gt reasoning traces when sampling  & \\ \addlinespace

\textbf{wd1}~\cite{tang2025wd1} & Policy Gradient & Mitigate computational overhead & Weighted likelihood requiring only & Discrete \\
& & and bias & one approximation & \\ \addlinespace

\textbf{SAPO}~\cite{xie2025step} & Policy Gradient (GRPO) & Learn structured coherent reasoning & Process-based reward function & Discrete \\
& & & & \\ \addlinespace

\textbf{SPG}~\cite{wang2025spg} & Policy Gradient & Reduce bias in single bound & Sandwiched by upper and lower bound; & Discrete \\
& & methods & block-wise masking  & \\ \addlinespace

\textbf{BGPO}~\cite{lin2025boundary} & Policy Gradient & Reduce memory overhead, scale to & Boundary-guided lower bound; & Discrete \\
& & larger MC sample size & constant memory via gradient accumulation & \\ \addlinespace

\textbf{JustGRPO}~\cite{ni2026flexibility} & Policy Gradient (GRPO) & Expose arbitrary-order limits  & Applies standard GRPO while avoiding & Discrete \\
& & in reasoning & harmful arbitrary-order exploration &  \\ \bottomrule
\end{tabular}%
}
\end{table*}

\subsubsection{Adapting Policy Gradient Methods to DLMs}
% SEPO, d1, diffucoder, mmada

% SEPO
Score Entropy Policy Optimization (SEPO)~\cite{zekri2025fine} introduces RLHF to discrete DLMs, 
proposing a theoretically grounded framework to fine-tune discrete diffusion models 
using policy gradient methods and non-differentiable rewards. 
Operating within the score entropy framework, SEPO adapts modern policy gradient methods 
like PPO and GRPO by using importance sampling to derive a stable and low-variance gradient estimate. 
This allows the model's policy to be iteratively updated to maximize a reward function,
making it a general framework for both 
conditional and unconditional generation. The objective function of SEPO is defined as follows:
\begin{equation}
l^{A}(\theta)=\mathbb{E}_{x\sim\pi_{\theta_{old}}}\left[\sum_{\substack{y\in\mathcal{X} \\ y\neq x}}w_{x,y}\log s_{\theta}(x,T-T_{0})_{y}\right]
\end{equation}

where the model parameters $\theta$ are optimized to maximize the expected log-likelihood of the score entropy $s_\theta$ weighted by $w_{x,y} = \pi_\theta(y)f(r^{T - T_0}_{x,y})$. 
The expectation is taken over samples $x$ from the previous policy $\pi_{\theta_{\text{old}}}$. 
The function $f$ can be selected to recover different policy gradient variants; for example, a clipped function yields PPO, while group-standardized rewards yield GRPO. 
This formulation enables stable and low-variance gradient estimation, even with non-differentiable rewards, and provides a flexible objective for fine-tuning discrete diffusion models.
Numerical experiments across several discrete generative tasks showcase scalability and efficiency of SEPO, demonstrating that policy gradient RL can be soundly applied to discrete diffusion models.

% d1
d1~\cite{zhao2025d1} provides a two-stage post-training framework for masked DLMs that combines supervised finetuning (SFT) with a novel policy gradient algorithm, diffu-GRPO. 
To adapt GRPO for DLMs, which lack a factorized likelihood, it introduces novel methods for both sequence log-probability and per-token log-probability estimation. d1 uses a simple mean-field decomposition to approximate sequence log-probability by a product of independent per-token probabilities, while per-token log-probability is computed by performing a single forward pass on a fully masked completion, conditioned on a randomly masked prompt during each policy gradient update. 
Using different random masks for the prompt in each inner gradient update step serves as a form of regularization, improving training efficiency and stability. 
The full d1 pipeline, leveraging SFT followed by diffu-GRPO, demonstrates significant performance improvements on mathematical and planning reasoning tasks for the LLaDA model.

% MMaDA
MMaDA~\cite{yang2025mmada}, a unified multimodal diffusion model, presents a three-stage training pipeline.
After first-stage pre-training, MMaDA employs a mixed Long chain-of-thought fine-tuning strategy, where reasoning trajectories from diverse tasks are curated into a unified format to align reasoning processes across modalities.
This facilitates code start training of the third stage, which introduces UniGRPO, a policy-gradient reinforcement learning algorithm tailored for diffusion language models.
UniGRPO overcomes the limitations of baseline methods like d1, by leveraging a structured noising strategy which samples a masking ratio $p_i \in [0, 1]$ uniformly rather than masking all response tokens.
This ensures the model is exposed to various stages of multi-step diffusion denoising process, from almost fully masked to nearly unmasked, which is consistent with conventional diffusion training and boosts the utilization of the model's multi-step denoising capabilities. Additionally, the sequence-level log-likelihood is approximated by averaging over masked tokens.

% DiffuCoder
DiffuCoder~\cite{gong2025diffucoder} is a 7B-parameter DLM specifically developed and analyzed for code generation. 
This work introduces an RL algorithm named coupled-GRPO, which is designed to be diffusion-native by leveraging the unique properties of the DLM generation process.
The central innovation of coupled-GRPO is its coupled-sampling scheme for log-likelihood estimation. 
To obtain a more robust and lower-variance estimate, it constructs paired, complementary masks for each completion sequence in a training batch. 
For a given sequence, two masks are generated such that every token position is masked in exactly one of the two masks. 
The log-probability estimate is then derived by averaging the losses from these two complementary forward passes. 
This ensures that every token is evaluated in a partial-masking context during training, providing full token coverage and a more stable gradient signal compared to methods that use a single random mask or a full mask. 
Coupled-GRPO is shown to substantially improve DiffuCoder’s performance on code generation tasks, while also encouraging more parallel, less autoregressive generation patterns.

% SPG
Sandwiched Policy Gradient (SPG)~\cite{wang2025spg} leverages both an upper and a lower bound of the true log-likelihood to reduce bias in single-sided approximation policy gradient methods for DLMs.
The two bounds of likelihood are estimated via Monte Carlo using a block-wise masking strategy to improve training stability.
SPG reports to achieve state-of-the-art performance compared with baseline methods on various reasoning benchmarks when applied to LLaDA.

% wd1
wd1~\cite{tang2025wd1} introduces a novel policy optimization approach that reformulates the objective as a weighted likelihood, requiring only a single approximation for the current parametrized policy likelihood.
This formulation reduces bias and improves both stability and training efficiency, outperforming prior diffusion-based RL methods by up to 16\% accuracy on reasoning tasks.

% IGPO
IGPO~\cite{zhao2025inpainting} leverages the unique inpainting ability of masked diffusion models to guide exploration during reinforcement learning. By partially injecting ground-truth reasoning traces,
IGPO alleviates the zero-advantage problem in group-based RL.

% SAPO
SAPO~\cite{xie2025step} proposes a step-aware policy optimization scheme that introduces fine-grained process rewards aligned with the latent reasoning hierarchy, mitigating "unstructured refinement" and yielding more interpretable multi-step reasoning traces.

% BGPO
To improve memory efficiency in ELBO-based RL for DLMs, BGPO~\cite{lin2025boundary} introduces a boundary-guided lower bound that allows large Monte Carlo sample sizes without increased memory usage, achieving stronger reasoning performance under the same hardware limits.

% JustGRPO
A recent work, JustGRPO~\cite{ni2026flexibility}, shows that effective reasoning in dLLMs can be better elicited by intentionally forgoing arbitrary-order generation during RL training and applying standard GRPO. This finding challenges the common view that flexible token orders are inherently advantageous: for reasoning tasks, generation order can become a critical training design choice rather than a free benefit.

\subsubsection{Adapting Preference Optimization to DLMs}
% llada 1.5
LLaDA 1.5~\cite{zhu2025llada} proposes a novel framework called Variance-Reduced Preference Optimization (VRPO) to adapt preference optimization methods to discrete DLMs.
The work identifies that applying Direct Preference Optimization (DPO) to discrete DLMs is challenging due to the high variance of the Evidence Lower Bound (ELBO) used to approximate log-likelihoods. 
VRPO addresses this by introducing two key unbiased variance reduction techniques: 
(1) Optimal allocation of the Monte Carlo sampling budget by sampling more diffusion timesteps rather than multiple masked versions per timestep, i.e. $n_t=n$ and $n_{y_t}=1$
(2) Antithetic sampling, where the same timesteps and masked data are shared between the ELBO estimates of the current policy $\pi_\theta$ and the reference policy $\pi_{\text{ref}}$ for the same input $y_w$ or $y_l$.
By applying VRPO to LLaDA, the resulting LLaDA 1.5 model shows significant and consistent improvements across mathematics, code, and alignment benchmarks.

\section{Inference Strategies}
\begin{figure*}[!t]
    \centering
    \includegraphics[width=\textwidth]{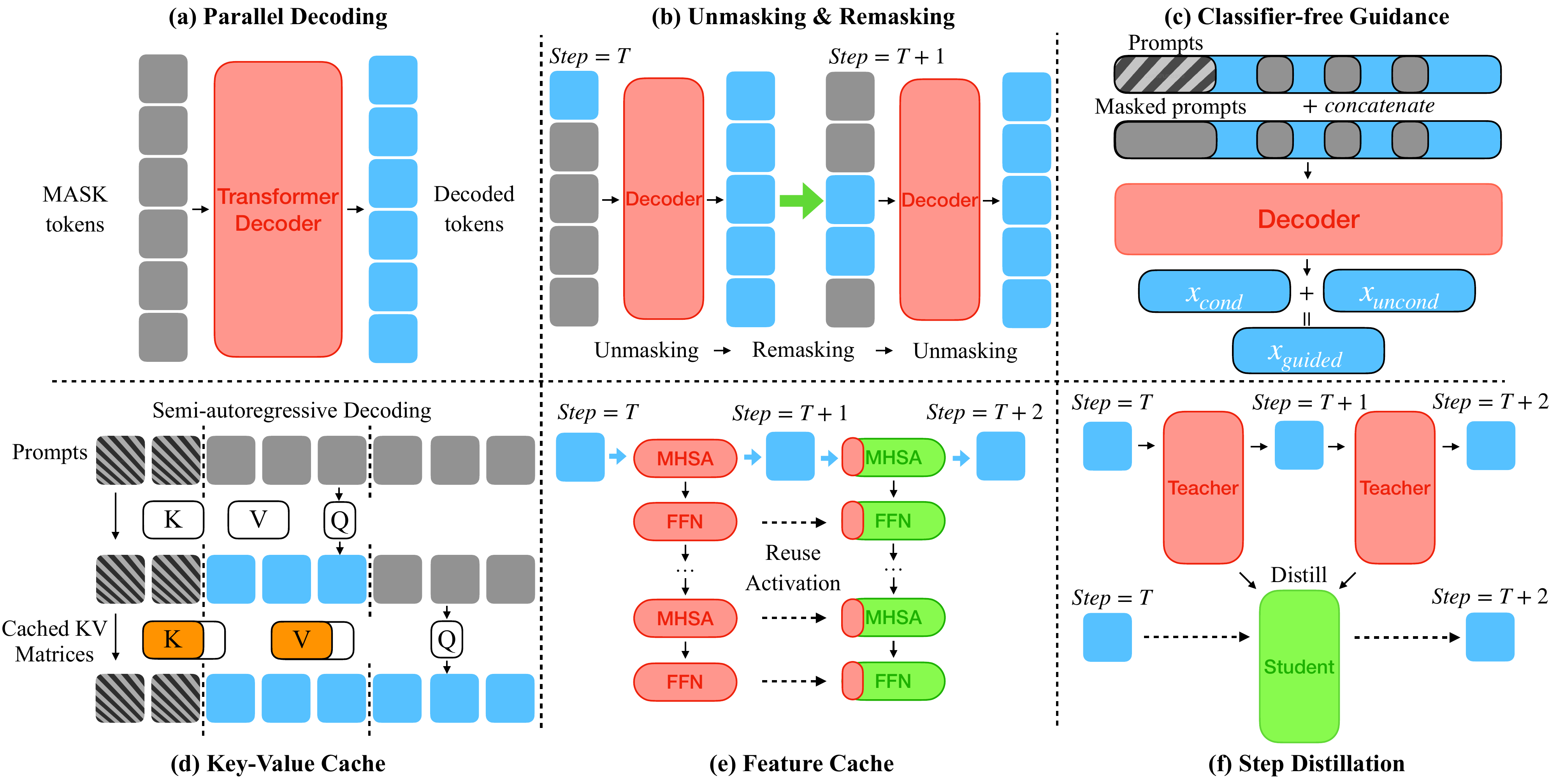}
    \caption{Inference Techniques of Diffusion Language Models. We illustrate six different strategies here, including: (a) Parallel Decoding; (b) Unmasking \& Remasking; (c) Classifier-free Guidance; (d) Key-Value Cache; (e) Feature Cache; and (f) Step Distillation.}
    \label{fig:inference}
\end{figure*}

% bowei
\label{sec:inference}
Inference strategies for DLMs serve three key goals: (i) boosting generation quality like unmasking and remasking schedules, (ii) enabling finer content control, and (iii) improving efficiency via techniques such as KV/feature cache and step distillation. A brief overview is presented in Fig.~\ref{fig:inference}.

\subsection{Parallel Decoding}
Parallel decoding naturally aligns with DLMs, leveraging their inherent mask-predict capability to generate multiple tokens simultaneously rather than sequentially. However, naïve parallelization can degrade coherence, motivating a series of adaptive strategies that balance efficiency and quality. Fast-dLLM \cite{wu2025fast} employs confidence-aware decoding, selectively unmasking tokens whose predicted probabilities exceed a threshold, and realizes up to 27.6× speedups without compromising quality. Adaptive Parallel Decoding (APD) \cite{israel2025accelerating} modulates the degree of parallelism on-the-fly by consulting a lightweight autoregressive auxiliary model, thus trading throughput for fidelity when necessary. SlowFast Sampling \cite{wei2025accelerating} introduces a two-stage schedule. Firstly, a cautious ``slow'' phase to locate stable tokens, then an aggressive "fast" phase to finalize them in bulk, achieving up to 34× acceleration when combined with caching. SpecDiff \cite{christopher2025speculative} further pushes throughput by using a discrete diffusion model as a fully parallel `drafter' whose output is quickly verified (and corrected if needed) by a larger autoregressive model, yielding up to 7.2× speedups over vanilla AR generation. 
Dimple \cite{yu2025dimple} employs confident parallel decoding, dynamically adapting the number of tokens revealed per step and cutting generation iterations by 1.5 - 7$\times$. 
Recent research has further advanced the efficiency of parallel decoding in diffusion language models
by introducing learnable and training-level optimization strategies.
Learn2PD~\cite{bao2025learning} introduces a learned adaptive parallel decoding policy, 
where a lightweight filter model predicts whether each token should be unmasked, 
replacing fixed confidence thresholds. 
dParallel~\cite{chen2025dparallel} enhances decoding efficiency through 
certainty-forcing distillation, training diffusion models to reach high confidence 
for multiple tokens in parallel. 
Collectively, these parallel decoding approaches substantially narrow the latency gap between diffusion and autoregressive models while preserving, and in some cases improving, generation quality.

\subsection{Unmasking/Remasking}
State-of-the-art open-source discrete DLMs such as LLaDA \cite{nie2025largelanguagediffusionmodels} and Dream \cite{dream2025} adopt a mask-predict paradigm: at each diffusion step they unmask high-confidence tokens and remask uncertain positions, iteratively refining the sequence. The choice of unmasking/remasking policy, i.e., low-confidence sampling, random selection, or adaptive temperature, therefore dominates both generation quality and convergence speed, making it one of the most critical inference levers. Early work Masked DLM \cite{sahoo2024simple} formalized two baselines: random remasking and confidence-ranked remasking, showing that prioritizing low-confidence positions yields better quality at no extra cost. Building on this insight, Fast-dLLM \cite{wu2025fast} introduces confidence-aware parallel decoding: every step it unmasks all positions whose predicted probabilities exceed a global threshold, realizing up to 13× speedups while maintaining accuracy. Most recently, ReMDM \cite{wang2025remasking} proposes a principled inference-time remasking sampler that can remask already decoded tokens for further refinement; by scaling the remasking budget, it offers a smooth compute–quality trade-off and closes the quality gap with autoregressive models under fixed compute. Collectively, these adaptive unmasking/remasking strategies substantially boost the efficiency and quality of diffusion language models, and they integrate cleanly with orthogonal accelerators that will be discussed later such as caching and step distillation.

\subsection{Guidance}
Guidance is a pivotal inference technique in diffusion models, steering the generative trajectory toward desired attributes and thereby enhancing output quality. In diffusion models, guidance refers to any technique that modifies the model’s denoising trajectory so that samples conform to a desired condition, such as a text prompt, a class label, or a stylistic attribute. The idea was popularized by classifier guidance \cite{dhariwal2021diffusion}, where gradients from an external classifier are added to the score estimate to nudge the sample toward a target class. Soon after, classifier-free guidance \cite{li2025adaptive} removed the need for an extra classifier: the model is trained once with and without conditioning, and at inference the two score estimates are combined:
\begin{equation}
s_{guided}=s_{uncond} + \lambda (s_{cond} - s_{uncond}),
\end{equation}
where $\lambda$ is the guidance scale that balances fidelity to the condition against sample diversity. This simple formulation now underpins most text-to-image systems (e.g., Stable Diffusion \cite{rombach2022high}) and has been adopted by DLMs for prompt-controlled generation. Subsequent work refines CFG along several axes: dropout-augmented CFG smooths the quality–diversity curve; particle-based guidance blends multiple conditions; and $p2$-weighting rescales the noise term to stabilize high-$\lambda$ sampling.  In the text domain, newer schemes extend guidance to structural and semantic constraints.  FreeCache~\cite{hu2025accelerating} couples a lightweight autoregressive verifier with a discrete DLM: the verifier approves (or vetoes) draft tokens before they are committed, simultaneously enforcing coherence and enabling aggressive feature caching.  DINGO~\cite{suresh2025dingo} formulates regular-expression control as a dynamic-programming search over a DFA, guaranteeing constraint satisfaction without altering the model distribution. In other discrete DLMs, guidance can also be applied at each diffusion step, optionally combined with masking/remasking or caching, to steer content (e.g., topic, sentiment) while preserving efficiency. Overall, guidance has become a cornerstone of diffusion inference, offering a lightweight, tunable handle for aligning model outputs with user intent.

\subsection{Efficient Inference}
Recent state-of-the-art diffusion language models \cite{zhu2025llada,labs2025mercury,dream2025} integrate the canonical Transformer architectures \cite{vaswani2017attention} with the step-wise stochastic inference procedures of diffusion processes. Consequently, efforts to accelerate inference in DLMs have converged on two complementary strategies: (1) lowering the per-step computational overhead of the Transformer backbone, e.g., through Key–Value (KV) Cache or Feature Cache.  (2) reducing the total number of diffusion sampling steps, e.g., via Step Distillation.

\noindent{\bf Key-Value Cache.} The conventional KV cache leverages the strictly autoregressive decoding pattern of LLMs and is therefore ill-suited to the bidirectional, multi-step generation paradigm of DLMs~\cite{ma2025dkv}. Recent work, however, shows that carefully redesigning the decoding schedule can recover much of its benefit. Block Diffusion \cite{arriolablock} introduces Block Discrete Denoising Diffusion Language Models (BD3-LMs), which decode text autoregressively across coarse blocks while running diffusion within each block; once a block is finished, its keys and values are frozen and reused, enabling variable-length generation and measurable speedups. Fast-dLLM \cite{wu2025fast} keeps the blockwise view but adds a training-free, approximate DualCache that exploits the near-identity of KV activations across successive diffusion steps for both prefix and suffix tokens, delivering up to $27\times$ end-to-end throughput gains on LLaDA and Dream with $<1\%$ accuracy loss. Complementing these block-based schemes, dKV-Cache \cite{ma2025dkv} observes that token representations stabilize only after a position is decoded and therefore deploys a delayed, conditional cache that stores KVs one step later; 
this design achieves $2$-$10\times$ speedups on the same models with negligible quality drop. 
d$^{2}$Cache~\cite{jiang2025d} introduces a fine-grained dual adaptive caching scheme that adaptively refreshes only rapidly changing KV states while reusing stable ones.
Elastic-Cache~\cite{nguyen2025attention} proposes an attention/depth-aware
adaptive refresh mechanism that selectively updates deeper layers and reuses stable 
shallow-layer caches.  
It performs attention-based drift detection to trigger cache refreshes only when
the most-attended tokens exhibit significant changes, 
yielding up to $45\times$ speedup with minimal quality loss.
Together, these results show that semi-autoregressive scheduling and delayed caching provide practical bridges between diffusion's bidirectional conditioning and Transformer tricks originally devised for autoregression.

\noindent{\bf Feature Cache.} Feature caching was first introduced by DeepCache \cite{ma2024deepcache}, which leverages the strong similarity of intermediate U-Net activations across consecutive diffusion steps to avoid redundant computation. Follow-up work $\Delta$-DiT \cite{chen2024delta}, Learning-to-Cache \cite{ma2024learning}, and FasterCache \cite{lvfastercache} demonstrate that the same principle transfers cleanly to Transformer-based diffusion models, yielding comparable speedups without retraining. With the rise of diffusion language models, dLLM-Cache \cite{liu2025dllm} extends feature caching to text by distinguishing two redundancies: prompt tokens remain almost static throughout denoising, whereas response tokens evolve only sparsely. It therefore pairs a long-interval prompt cache with an adaptive short-interval response cache refreshed only when a lightweight value-similarity test (“V-verify”) detects substantial change, achieving up to $9\times$ end-to-end speedups on LLaDA-8B and Dream-7B. Most recently, FreeCache \cite{hu2025accelerating} caches the KV/feature projections of already “clean’’ tokens and refreshes only dynamic positions, pushing acceleration further to $34\times$ while preserving fidelity. Collectively, these advances illustrate that feature caching can bring diffusion language models within striking distance of autoregressive LLMs in inference latency without sacrificing output quality.

\noindent{\bf Step Distillation.} Step distillation is a widely adopted acceleration technique for diffusion models, collapsing the typical thousand-step denoising process into only a few and sometimes even a single sampling step, thereby drastically reducing inference time. Unlike the training-free methods discussed earlier, it imposes an offline cost: a compact student network must first be trained to mimic the teacher. Early work such as Progressive Distillation \cite{salimansprogressive}, followed by ADD \cite{sauer2024adversarial} and LADD \cite{sauer2024fast}, progressively halves the step count or aligns intermediate distributions to preserve fidelity. Di4C \cite{hayakawadistillation} extends the framework to discrete diffusion by explicitly distilling inter-token correlations, enabling four to ten steps students that match teacher quality while providing $\sim$2× speedups. Most recently, DLM-One \cite{chen2025dlm} employs score-based distillation with adversarial regularisation to train a continuous diffusion language model that generates an entire sequence in a single forward pass, achieving up to 500× acceleration with near-teacher quality. Collectively, these works establish step distillation as the principal route toward closing the latency gap between diffusion and autoregressive language models.
\section{Multimodal and Unified Approaches}
% Tianyi
\label{sec:multimodal}
This section explores recent developments in extending DLMs to multimodal and unified architectures. 
Similar to autoregressive LLMs, DLMs can be naturally adapted to handle multimodal inputs and outputs. 
A straightforward approach is to accept vision inputs through a pretrained vision encoder.
Following the success of LLaVA~\cite{liu2023visual} in the AR domain, models such as LLaDA-V~\cite{you2025llada}, LaViDa~\cite{li2025lavida}, and Dimple~\cite{yu2025dimple} employ vision encoders to extract image features, which are then projected into the same embedding space as text tokens.
Beyond simple visual understanding, DLMs offer a promising pathway toward unified multimodal generation and understanding. 
Thanks to their shared denoising diffusion framework, DLMs naturally support joint modeling of different modalities. 
Visual inputs can be discretized using VQ-VAE, enabling training on multimodal inputs and outputs in a unified token space. 
Representative models such as MMaDA~\cite{yang2025mmada}, Fudoki~\cite{wang2025fudoki}, and Muddit~\cite{shi2025muddit} exemplify this direction.

% \paragraph*
\noindent{\textbf{LLaDA and LLaDA's Derivatives.}}
% llada-v, mmada, lavida, 
We begin by introducing the LLaDA~\cite{nie2025largelanguagediffusionmodels} family and its derivatives, which are built on the architecture and pretrained weights of the base LLaDA model.
LLaDA-V~\cite{you2025llada} integrates a vision encoder with an MLP-based projector that maps visual features into the language token embedding space, enabling effective visual instruction tuning. 
Following LLaVA-NeXT~\cite{lillava}, LLaDA-V adopts three-stage tuning strategies.
In the first stage, they only train the MLP projector to align visual representations with text embeddings using LLaVA's training data. In the second stage, the model is further tuned by large-scale visual instruction data~\cite{guo2024mammoth} using DLM objective. The third stage is to enhance multimodal reasoning capabilities by training on QA pairs with reasoning chains. 
Although the LLaDA backbone is slightly weaker than LLaMA3-8B~\cite{grattafiori2024llama} on pure text tasks, LLaDA-V achieves strong performance and better scalability across various benchmarks compared with LLaMA3-V trained on the same data.
It narrows the performance gap with Qwen2-VL~\cite{wang2024qwen2} and outperforms both hybrid and pure DLM-based models~\cite{li2025dual, xieshow, kou2024orthus}, demonstrating the effectiveness of diffusion architectures in multimodal understanding.

LaViDa~\cite{li2025lavida} introduces a family of VLM based on LLaDA and Dream-7B~\cite{dream2025}.
Also utilizing a pretrained vision encoder, LaViDa uses a two-stage training strategy to train the projector and finetune the model respectively.
LaViDa makes notable contributions to address training and inference challenges of multimodal DLMs.
Typically, in masked DLMs, only about 50\% of the tokens are masked for loss computation on average, which reduces efficiency and may omit critical answer tokens during VLM training, thereby causing gradient misalignment.
LaViDa introduces complementary masking for effective training:
For each sample, two masked versions with disjoint corrupted spans are generated, ensuring all tokens are eventually used in training and improving sample efficiency and gradient flow.
During inference, LaViDa employs Prefix KV-Cache to cache the keys and values of visual and prompt tokens, significantly reducing latency and achieving a maximum speedup of 3.9$\times$ with a marginal performance drop. 
Additionally, timestep shifting is used to unmask tokens earlier, further boosting generation quality. 
Empirical results show that LaViDa achieves competitive or superior performance to AR-based VLMs, while enjoying significant inference speedup.

Lavida-O~\cite{li2025lavidao} further extends LaViDa into a full-spectrum unified multimodal model capable of both high-quality image generation and fine-grained understanding tasks. 
It introduces a novel Elastic Mixture-of-Transformers (Elastic-MoT) architecture that decouples the model into a lightweight generation branch and a more powerful understanding branch, enabling scalable training and inference. 
Lavida-O uniquely supports localized object-level understanding, instruction-based image editing, high-resolution text-to-image synthesis (1024px), and interleaved reasoning and planning within a single unified diffusion framework.

Building upon LLaDA, MMaDA~\cite{yang2025mmada} further generalizes the architecture to support both multimodal understanding and generation. 
Unlike prior models, MMaDA eliminates the need for an explicit vision encoder by tokenizing images into discrete codes using VQ-VAE, and modeling all modalities jointly with a modality-agnostic diffusion transformer. 
This design allows seamless integration across text and image modalities without modality-specific components.
MMaDA also implements a mixed long CoT fine-tuning strategy that aligns CoT reasoning format across modalities.
Moreover, UniGRPO, a unified policy-gradient based RL algorithm, is tailored specially for diffusion language models, making it possible to reason across modalities.
Not only surpass similar-sized models like LLaMA3 for textual reasoning and Show-o~\cite{xieshow} for multimodal understanding, MMaDA even excels professional image generation models like SDXL~\cite{podellsdxl} in image generation.

MMaDA-Parallel~\cite{tian2025mmada} replaces the sequential reasoning-then-generation pipeline in MMaDA with a fully parallel multimodal diffusion framework, enabling text and images to interact bidirectionally at every denoising step. 
By jointly generating reasoning traces and visual outputs and further optimizing cross-modal consistency via a trajectory-level Parallel RL (ParaRL) algorithm, MMaDA-Parallel substantially improves semantic alignment and thinking-aware image synthesis performance.

\noindent{\textbf{Dimple}.}
Dimple \cite{yu2025dimple} introduces a large multimodal DLM, combining a vision encoder with a discrete DLM backbone. 
The authors identify that a pure discrete diffusion training approach suffers from significant instability, poor performance, and severe length bias. 
To overcome these challenges, Dimple proposes a novel two-phase training paradigm called Autoregressive-then-Diffusion. 
In the first phase, the model undergoes standard autoregressive training to effectively align the vision and language modalities. 
In the second phase, it switches to diffusion-based training to restore its parallel decoding capabilities. 
This hybrid strategy ensures stable and efficient training while achieving performance comparable to or even better than contemporary autoregressive models like LLaVA-NEXT.

For inference, 
Dimple introduces several techniques to improve efficiency and controllability. 
Confident Decoding dynamically adjusts the number of tokens generated in each step based on a confidence threshold, which reduces the total number of generation iterations. 
The model also successfully re-implements the prefilling technique, common in autoregressive models, to cache prompt tokens and achieve a speedup of up to 7$\times$ with minimal performance loss. 
Furthermore, Dimple explores the use of Structure Priors, allowing for precise, fine-grained control over the response format and length, a feature that is difficult to achieve in autoregressive models.

\noindent{\textbf{D-DiT}.}
Dual Diffusion Transformer (D-DiT) \cite{li2025dual} is a large-scale fully end-to-end unified multimodal diffusion model that supports both text-to-image (T2I) and image-to-text (I2T) tasks.
It directly addresses the challenges previous diffusion models faced in visual understanding tasks, which have been largely dominated by autoregressive models.
The architecture is inspired by the Multimodal Diffusion Transformer (MM-DiT), featuring a dual-branch transformer that processes image and text tokens, with attention mechanisms allowing interaction between modalities in every layer.
The model uses a frozen VAE for image processing and a frozen T5 encoder for text, and the major backbone MM-DiT is initialized from pretrained SD3~\cite{esser2024scaling} weight.

One core innovation of D-DiT is its joint training objective, which combines continuous latent-space diffusion for images and discrete masked-token diffusion for text by jointly optimizing the sum of both modalities' losses.
Unlike prior multimodal diffusion models that required an autoregressive component to decode text latents, D-DiT is fully diffusion-based and demonstrates competitive performance against other unified models.

\noindent{\textbf{UniDisc}.}
Unified Multimodal Discrete Diffusion (UniDisc)~\cite{swerdlow2025unified} is proposed as a unified generative model for the joint text and image modeling, building upon discrete diffusion as an alternative to dominant AR approaches.
Different from previously discussed D-DiT, UniDisc employs an entire masked diffusion process jointly on text and image tokens with full attention, learning to map a sequence of masked tokens back to a clean sequence from a shared vocabulary.
Training is performed using a unified discrete diffusion objective from scratch, where tokens from both modalities are randomly masked and the model is supervised with a re-weighted cross-entropy loss.

A key advantage of UniDisc is its superior performance in conditional generation tasks, which is largely attributed to the effective use of classifier-free guidance.
One of the most notable capabilities of UniDisc is its ability to perform joint image and text inpainting in a zero-shot manner, a feature not possible with previous AR or unified generative models.
The author performs scaling analysis by scaling up the model up to 1.4B, demonstrating UniDisc outperforms AR models in terms of both performance and inference-time compute, with enhanced controllability and editability.
However, UniDisc is found to be less training-efficient than a comparable AR model in terms of achieving the same validation loss.

\noindent{\textbf{Fudoki}.}
Fudoki~\cite{wang2025fudoki} is introduced as the first general-purpose unified multimodal model built entirely on the discrete flow matching framework, challenging the dominance of autoregressive (AR) and masking-based diffusion models.
Instead of relying on a simple masking corruption process, Fudoki leverages a more general metric-induced probability path with kinetic optimal velocities, which allows for a more semantically meaningful corruption process and enables the model to continuously self-correct its predictions during iterative refinement. This self-correction capability is a key distinction from masked DLMs, where unmasked tokens are typically fixed and cannot be revised.

To reduce the high cost of training from scratch, Fudoki is initialized from a pre-trained AR-based MLLM, Janus-1.5B~\cite{wu2025janus}, and is then adapted to the discrete flow matching paradigm in a two-stage process.
Its architecture is based on Janus-1.5B but uses a full attention mask to better capture global context and removes time embedding layers, as the model can implicitly infer the timestep from the corrupted input.
Fudoki achieves performance comparable to state-of-the-art AR models in both visual understanding and image generation tasks, demonstrating a flexible trade-off between inference speed and quality.
The model shows significant performance gains when test-time inference scaling techniques are applied, suggesting the potential of this architecture to be further explored for next-generation unified models.

\noindent{\textbf{Muddit}.}
Muddit~\cite{shi2025muddit} is a pure unified discrete diffusion transformer that integrates a strong text-to-image backbone with a lightweight text decoder, enabling flexible and high-quality multimodal generation under a truly unified architecture. 
Initialized from pretrained MM-DiT from Meissonic~\cite{bai2024meissonic}, the model is trained using a unified discrete diffusion objective, where text and image tokens are stochastically masked according to a cosine schedule and the model learns to predict the original tokens via a re-weighted cross-entropy loss.
By a combination of the strength from a semantically rich visual prior and parallel discrete diffusion, Muddit achieves competitive or superior performance compared to significantly larger AR models across generation and understanding benchmarks.
It also demonstrates several times speedup over AR baseline, highlighting the efficiency and scalability of a discrete diffusion approach when properly initialized.

\noindent{\textbf{Lumina-DiMOO.}} 
Lumina-DiMOO~\cite{xin2025lumina} is a state-of-the-art open-source unified multimodal diffusion model that achieves fast and high-quality multi-modal generation and understanding through a fully discrete diffusion framework. 
Built on LLaDA, it expands the vocabulary to include 8,192 visual tokens from aMUSEd-VQ~\cite{patil2024amused} and employs a unified training objective over mixed text-image sequences. 
Lumina-DiMOO supports a wide range of tasks, including text-to-image generation, image editing, subject-driven and controllable generation, and advanced image understanding. 
It introduces innovations such as Max Logit-based Cache (ML-Cache) for sampling acceleration, parallel and block-wise sampling for efficient decoding, and an end-of-line special token to support arbitrary image resolutions. 
The training of Lumina-DiMOO is performed in four stages, culminating with Self-GRPO, a self-improving reinforcement learning algorithm that enhances generation and understanding alignment. 
Lumina-DiMOO ranks first among open-source models on the UniGenBench~\cite{Pref-GRPO&UniGenBench} leaderboard, offering 32$\times$ speedup over AR baselines while delivering superior generation quality.

\section{Performance Study} 
\label{sec:performance}

\begin{figure*}[h!]
  \label{fig:performance}
  \centering
    \includegraphics[width=\textwidth, trim=70 00 100 0, clip, keepaspectratio=false]{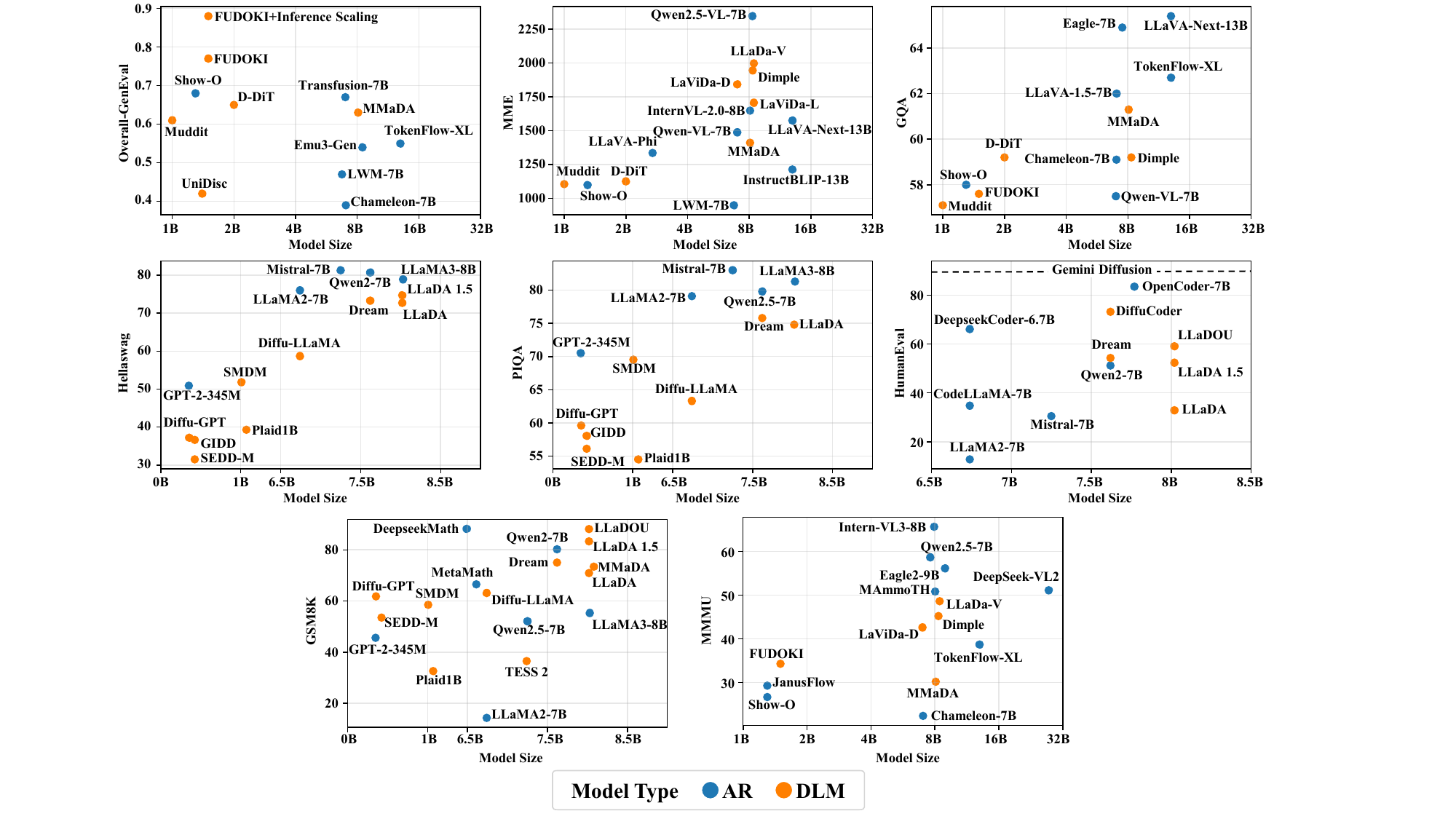}
  \caption{Performance comparison on eight benchmarks:
  Overall-GenEval, MME, CQA, Hellaswag, PIQA, HumanEval, GSM8K, and MMMU. The horizontal axis in each subplot represents the model size, measured in the number of parameters. The vertical axis indicates the score under the corresponding benchmark, with higher scores reflecting better performance. Model types are distinguished by color: blue represents AR language models, while orange represents DLMs.
}
\end{figure*}

% mingda
In this section, we briefly compare the performance of various DLMs with AR models. 
We present visualizations based on several widely used benchmarks for evaluating DLMs, including PIQA~\cite{bisk2020piqa} and HellaSwag~\cite{zellers2019hellaswag} for general language understanding, 
HumanEval~\cite{chen2021evaluating} for code generation, and GenEval~\cite{ghosh2023geneval}, MME~\cite{fu2023mme}, MMMU~\cite{yue2024mmmu} and GQA~\cite{hudson2019gqa} for multimodal generation and comprehension. 
We also include GSM8K~\cite{cobbe2021training}, a popular benchmark in DLM literature for assessing mathematical reasoning capabilities.
The corresponding performance visualizations are shown in Fig.~\ref{fig:performance}.

The DLMs surveyed range in size from under 1B to 8B parameters. 
For comparison, we also report the performance of representative AR models of similar scale. 
Performance data are primarily taken from original publications. 
If results were not available in the source papers, we consulted subsequent works that reported comparable evaluations.

Our findings suggest that DLMs generally perform competitively with AR models of comparable size. 
On general language understanding benchmarks such as PIQA and HellaSwag, models like LLaDA achieve performance that is slightly below or on par with AR models such as LLaMA2~\cite{touvron2023llama} and Qwen2.5~\cite{team2024qwen2}.
However, DLMs exhibit stronger performance in math and science-related benchmarks, including GSM8K, GPQA~\cite{rein2024gpqa}, and MATH~\cite{hendrycks2measuring}, where models such as LLaDA and Dream consistently outperform similarly sized AR counterparts. 
In multimodal tasks, models like MMaDA~\cite{yang2025mmada} and LLaDA-V~\cite{you2025llada} often surpass AR-based multimodal models, highlighting the potential of DLMs in unified and cross-modal reasoning.
On code generation tasks, DLMs also demonstrate competitive capabilities. 
Notably, DiffuCoder~\cite{gong2025diffucoder} achieves competitive HumanEval performance among open-source models, illustrating the potential of DLMs in structured, logic-heavy domains.
Furthermore, closed-source DLMs such as Gemini Diffusion~\cite{deepmind2024geminidiffusion} and Mercury~\cite{labs2025mercury} achieve state-of-the-art results among all DLMs, rivaling top-tier AR models like GPT-4o. 

Given the relatively limited training data and computational resources used to train most current DLMs, these results suggest that DLMs hold strong potential as viable alternatives to AR models in many real-world applications.

Recent scaling studies further show that DLMs tend to outperform AR models in data-constrained, multi-epoch regimes, likely because their any-order denoising objective enables 
more effective reuse of limited data~\cite{ni2025diffusion}.

\section{Applications on Downstream Tasks}
% mingda

\label{sec:application}
\subsection{Conventional NLP Tasks}
Before the emergence of large-scale DLMs for general-purpose language generation, 
DLMs have already been applied to various conventional NLP tasks, such as text classification~\cite{yuan2024roic}, named entity/scene recognition~\cite{shen2023diffusionner,yang2025ipad}, sentiment analysis~\cite{liu2024let}, document summarization~\cite{zhang2023diffusum,dong2025termdiffusum}, style transfer~\cite{horvitz2024paraguide,lyu2023fine}, constrained generation~\cite{zhang2023planner,liu2023diffucom,xiang2024diffusiondialog,zou2024improved,hu2024poetrydiffusion}, and machine translation~\cite{chen2023xdlm, demirag2024benchmarking}, etc.

% classification
ROIC-DM~\cite{yuan2024roic} is the first work to adapt diffusion models for robust text classification and inference. 
It applies the diffusion process directly to the class labels and conditions the denoising process on the input text, which can be further enhanced by incorporating traditional language models as advisors. 
% named entity recognition
DiffusionNER~\cite{shen2023diffusionner} formulates Named Entity Recognition as a boundary-denoising task. 
It applies a diffusion process to the start and end boundaries of entities, generating entity spans from random noise through an iterative refinement process.
For scene text recognition, IPAD~\cite{yang2025ipad} introduces a parallel, iterative network that frames the task as conditional text generation, employing discrete diffusion and an easy-first decoding method to effectively balance recognition accuracy and inference speed.
% sentiment analysis
For aspect-based sentiment analysis, DiffusionABSA~\cite{liu2024let} employs a diffusion model to progressively extract the aspects step-by-step.
% summarization
DiffuSum~\cite{zhang2023diffusum} proposes a novel paradigm for extractive summarization by using a diffusion model to directly generate desired summary sentence representations. 
The final summary is then formed by extracting document sentences that best match these generated representations.
For legal document summarization, TermDiffuSum~\cite{dong2025termdiffusum} proposes a term-guided diffusion model that prioritizes sentences with legal terminology via a multifactor fusion noise weighting schedule.
% extraction
For keyphrase extraction, Diff-KPE~\cite{luo2024enhancing} enhances phrase representations by guiding a text diffusion process with a Variational Information Bottleneck to generate and inject keyphrase information.
IPED~\cite{zhao2024iped} treats relational triple extraction as an implicit block diffusion task.
% editing
EdiText~\cite{lee2025editext} introduces a controllable coarse-to-fine text editing framework by integrating an SDEdit-based technique with a novel self-conditioning method for precise editing control.
% empathetic and detoxic
To generate more specific empathetic responses, DIFFUSEMP~\cite{bi2023diffusemp} utilizes a conditional diffusion model guided by multi-grained control signals (e.g., intent and semantic frames) that are integrated via a special masking strategy.
DiffuDetox~\cite{floto2023diffudetox} utilizes a mixed diffusion approach for text detoxification, combining a conditional model to reduce toxicity with an unconditional model to ensure the fluency of the output text.
% style transfer
A finetuned DiffuSeq model is shown to achieve state-of-the-art performance on fine-grained text style transfer tasks~\cite{lyu2023fine}, while ParaGuide~\cite{horvitz2024paraguide} introduces a more flexible plug-and-play framework that guides a paraphrase-conditioned diffusion model with off-the-shelf classifiers and style embedders at inference time.
% paragraph generation
To generate fluent and diverse paragraphs while avoiding repetition, PLANNER~\cite{zhang2023planner} combines a latent diffusion planning module to generate semantic paragraph embeddings with an autoregressive decoding module to render the final text.
DiffuCom~\cite{liu2023diffucom} presents an efficient diffusion model for comment generation that uses context-aware attention mechanism and self-conditioning technology.
DiffusionDialog~\cite{xiang2024diffusiondialog} tackles the one-to-many problem in dialogue generation by performing a diffusion process with continuous latent variables, improving response diversity and inference speed. 
For paraphrase generation, LDP~\cite{zou2024improved} models diffusion in a pretrained model's latent space, avoiding the typical rounding step to achieve greater efficiency.
% pottery generation
For the highly constrained task of poetry generation, PoetryDiffusion~\cite{hu2024poetrydiffusion} uniquely separates the task by using the diffusion model to generate semantics while a novel, independently trained metrical controller enforces structural rules like format and rhyme.
% translation
In machine translation, XDLM~\cite{chen2023xdlm} pioneers a cross-lingual pre-training objective for diffusion models, enabling them to effectively learn the mapping between languages in the pretraining stage.
% retrieval
DiffusionRet~\cite{qiao2023diffusionret} proposes a two-stage generative retrieval method that first utilizes a diffusion model to generate a pseudo-document from a query, which then serves as input for an n-gram-based model to retrieve the final document.
% fake news detection
DIFND~\cite{yan2025debunk} employs a diffusion model to generate debunking evidence and a multi-agent MLLM system for chain-of-debunk reasoning to improve accuracy and interpretability for multimodal fake news detection.

\subsection{Code Generation}
% DCOLT, DiffuCoder, Mercury Coder, DUS
Although DLMs are rarely explicitly designed for code generation, the global planning and iterative refinement capabilities of them are particularly well-suited for the non-sequential nature of code generation.
Foundational models like DiffuCoder~\cite{gong2025diffucoder}, a 7B open-source model, have been developed specifically for this domain.
DiffuCoder's analysis reveals unique decoding behaviors, such as generation order becoming more flexible at higher temperatures. 
It also proposes coupled-GRPO, a novel sampling scheme that constructs complementary mask noise for completions used in training, which significantly improves the model's performance on code generation tasks.
Building on the reasoning aspect, DCoLT~\cite{huang2025reinforcing} treats the entire reverse diffusion process as a form of non-linear, "lateral" thinking.
With outcome-based RL and unmasking policy module, it achieves strong results on complex coding tasks.
Dilated Unmasking Scheduler (DUS)~\cite{luxembourg2025plan} offers an inference-only, planner-free method that unmasks tokens in a non-adjacent pattern to minimize an upper bound on joint entropy gain at each denoising step, achieving promising results on code generation while improving speed-quality trade-off.
Demonstrating the real-world potential of DLMs' speed, Mercury Coder~\cite{labs2025mercury} is a commercial-scale diffusion model that achieves state-of-the-art throughput, outperforming speed-optimized autoregressive models by up to 10$\times$ while maintaining comparable quality on major code benchmarks. 
Stable-DiffCoder~\cite{fan2026stable} builds a block diffusion code model on the Seed-Coder training pipeline with block diffusion continual pretraining. DICE~\cite{bai2026dice} targets CUDA kernel generation with the CuKe dataset and a bi-phase curated RL framework.

\subsection{Biological and Scientific Applications}
TransDLM~\cite{xiong2024text} performs molecular optimization guided by a textual description of target properties to avoid the error propagation.
Another text-guided approach, TGM-DLM~\cite{gong2024text}, focuses on molecular generation by collectively and iteratively updating token embeddings of SMILES strings. 
Without relying on additional data resources, TGM-DLM surpasses MolT5-Base in generation performance. 
DRAKES~\cite{wangfine} introduces an RL-based fine-tuning method for discrete diffusion models that backpropagate rewards using the Gumbel-Softmax trick for DNA and protein design.
For protein modeling, ForceGen~\cite{ni2024forcegen} enables de novo protein design by using a protein language diffusion model to generate sequences that meet complex, nonlinear mechanical property-design objectives.
MeMDLM~\cite{goel2024memdlm} introduces a masked diffusion language model for de novo membrane protein design by fine-tuning the ESM-2 protein language model to generate novel and realistic transmembrane sequences.
Inspired by LLaDA, DSM~\cite{hallee2025diffusion} introduces a enabling both high-quality representation learning and effective generative protein design. 
DPLM~\cite{wangdiffusion} offers a versatile protein language model that exhibits strong generative and predictive capabilities for protein sequences, and demonstrates superior performance in representation learning. 
DPLM2~\cite{wang2024dplm} further extends the model into a multimodal protein foundation model that can simultaneously process both sequences and structures. 
By converting 3D structural coordinates into discrete tokens, DPLM-2 learns the joint distribution of these two modalities. 
This enables the simultaneous co-generation of compatible protein sequences and their 3D structures, in addition to supporting conditional tasks such as protein folding and inverse folding.
CFP-GEN~\cite{yincfp} is a novel diffusion language model designed for Combinatorial Functional Protein Generation. 
It facilitates de novo protein design by integrating multimodal constraints, including functional, sequence, and structural information. 
CFP-GEN supports high-throughput generation of novel proteins with functionality comparable to that of natural proteins and achieves a high success rate in the design of multifunctional proteins.

\subsection{Robotics}
Recently, DLM-based vision-language-action (VLA) models have demonstrated strong potential in unifying perception, reasoning, and control. 
Built upon LLaDA, LLaDA-VLA~\cite{wen2025dvla} incorporates localized special-token classification alongside hierarchical action-structured decoding, leading to significant improvements over autoregressive VLA baselines in both simulation and real-world evaluations.
dVLA~\cite{wen2025dvla} leverages pretrained MMaDA as a diffusion backbone to jointly 
generate visual subgoal images, textual chain-of-thought, and discretized actions, further introducing prefix-attention masking and KV caching for efficient long-horizon manipulation. 
Unified Diffusion VLA (UD-VLA)~\cite{chen2025unified} proposes a Joint Discrete Denoising Diffusion Process that synchronously denoises future image and action tokens within a shared token space, achieving state-of-the-art performance on benchmarks with substantially improved inference speed.

\section{Challenges and Future Directions}
\label{sec:challenges}
While diffusion language models have shown considerable promise across a wide range of tasks, several key challenges still remain and limit their practical deployment and broader application.
In this section, we outline and discuss critical areas that require further research and innovation.

\subsection{Major Challenges}

\begin{figure*}[!t]
    \centering
    \includegraphics[width=1\textwidth]{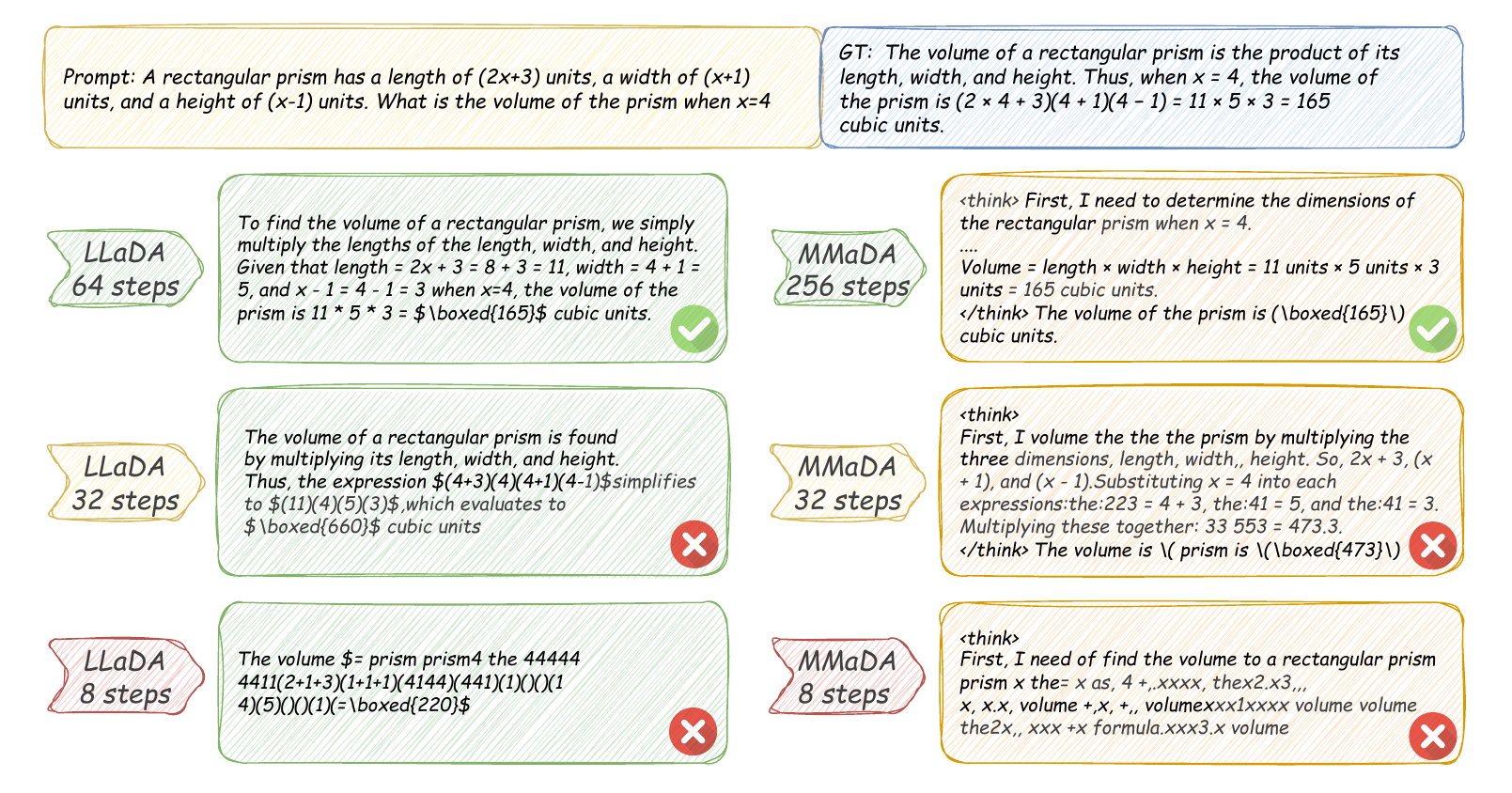}
    \caption{Generation results of LLaDA~\cite{nie2025largelanguagediffusionmodels} and MMaDA~\cite{yang2025mmada} under different denoising step settings. Note that the generation length is set to 128 tokens and 256 tokens for LLaDA and MMaDA respectively. Both models generate a correct and coherent response only when 1 or 2 tokens are unmasked at each step. With fewer steps and more parallelism, the responses are either incorrect or lack fluency and consistency. This illustrates the trade-off between parallelism and output quality in DLMs. We omit part of the thinking process of MMaDA with 256 steps for simplicity.
    }
    \label{fig:decoding_tradeoff}
\end{figure*}

\noindent{\textbf{1) Parallelism–Performance Trade-off.}}  
Diffusion language models are designed to generate multiple tokens in parallel.  
However, this parallelism often comes at the expense of generation quality and consistency.  
In discrete DLMs, unmasking multiple tokens simultaneously in a single step increases the denoising burden, which can lead to error accumulation.  
A central issue is the interdependence between tokens, known as the \textbf{Parallel Decoding Curse}~\cite{wu2025fast}.  
When predicting multiple tokens at once, the model produces a distribution for each position and samples from them independently, failing to account for dependencies among positions.  
Consider a simple example where the training data consists only of two sequences: ``ABABAB'' and ``BABABA''.  
Statistically, ``A'' and ``B'' appear with equal frequency at each position in the training data, leading DLMs to assign them similar probabilities during prediction.
In autoregressive models, once the first ``A'' is generated, the model is likely to predict ``B'' next, preserving consistency.  
In contrast, a DLM generating tokens in parallel may independently sample ``A'' for both the first and second positions, producing a sequence like ``AAABBA'', which deviates from valid training patterns.
Empirical studies show that this issue significantly affects DLM performance, particularly when the number of denoising steps is reduced~\cite{gongscaling}. 
This phenomenon is illustrated in Fig.~\ref{fig:decoding_tradeoff}.
Future work may focus on mitigating this trade-off. Potential directions include introducing structured constraints, modeling inter-token dependencies more explicitly, or refining sampling strategies to improve coherence during parallel generation.

\noindent{\textbf{2) Infrastructure.}}
While the training, fine-tuning, and inference of AR models have been significantly simplified and accelerated by open-source, highly optimized libraries and frameworks (e.g., Hugging Face Transformers~\cite{wolf2020transformers}), DLMs still lag behind in this regard.
Currently, major machine learning ecosystems offer little to no native support for DLMs, posing practical challenges for researchers and developers.
Furthermore, during inference, DLMs lack mature, open-source deployment infrastructure akin to vLLM~\cite{kwon2023efficient}, making efficient serving of DLMs difficult.

\noindent{\textbf{3) Long Sequence and Dynamic-Length Generation.}}
DLMs are typically trained to denoise fixed-length sequences under a diffusion-based objective, which makes it challenging to generalize to longer or dynamically sized sequences at inference time.
Most existing DLMs are limited to a maximum context length of 4,096 tokens, and widely used extrapolation techniques in AR models for longer sequences remain underexplored in the DLM setting.
This limitation hinders the applicability of DLMs in tasks requiring long-context understanding or complex reasoning.
In addition, DLMs generally require the generation length to be predetermined during inference, making them ill-suited for dynamic-length generation.
Although DLMs can predict an [EOS] token and omit displaying tokens generated afterward, the entire sequence is still fully updated throughout the denoising process, regardless of whether the generation has logically ended, which leads to unnecessary computational overhead.
Recent work has begun to address this limitation through both training-based~\cite{Dreamon2025,yang2025diffusion} and training-free~\cite{li2025beyond,chen2025dpad} approaches.
In addition, masked DLMs utilize full bidirectional attention at every denoising step, which incurs a computational cost of $\mathcal{O}(N^2)$ per step, where $N$ is the sequence length. 
Assuming a fixed number of tokens are unmasked at each step, the total number of denoising steps scales linearly with $N$, leading to an overall inference complexity of $\mathcal{O}(N^3)$. 
Without architectural optimizations such as KV-Cache, this cubic time complexity severely limits the scalability of DLMs for long-sequence generation in real-world applications.

\noindent{\textbf{4) Scalability.}}
Scalability remains an underexplored challenge for diffusion language models, particularly in comparison to autoregressive models.
Although DLMs have shown promising results on certain metrics and benchmarks, they have yet to be scaled to the same extent as AR counterparts.
% The largest publicly available DLM currently contains only around 8B parameters, 
% significantly smaller than leading AR models that have been scaled to hundreds of billions or even trillions, such as Llama-3.1-405B~\cite{grattafiori2024llama}, DeepSeek-V3-671B-A37B MoE~\cite{liu2024deepseek}, Qwen3-235B-A22B MoE~\cite{yang2025qwen3}, Kimi-K2-1T-A32B MoE~\cite{team2025kimi}, etc.
Earlier public DLMs were mostly concentrated around the 7B--8B scale, while recent models such as LLaDA2.0~\cite{bie2025llada2} have begun to extend public DLMs to larger model sizes.
Nevertheless, compared with leading AR models that have been scaled to hundreds of billions or even trillions of parameters, 
such as Llama-3.1-405B~\cite{grattafiori2024llama}, DeepSeek-V3-671B-A37B MoE~\cite{liu2024deepseek}, 
Qwen3-235B-A22B MoE~\cite{yang2025qwen3}, and Kimi-K2-1T-A32B MoE~\cite{team2025kimi}, large-scale DLMs remain relatively less explored.
Closed-source DLMs, such as Mercury and Gemini Diffusion, also fall short of state-of-the-art AR models across a wide range of benchmarks.
Furthermore, many existing DLMs are trained either from previously pretrained AR models or built upon baseline DLMs (e.g., LLaDA) using limited datasets, which further constrains their scalability and performance.
Therefore, the ability to further scale up DLMs still needs to be validated or explored.

\subsection{Future Directions}
Despite the challenges discussed above, DLMs present many promising directions for future exploration. 
Below, we briefly outline several under-explored directions and opportunities that could significantly advance the field:
\begin{itemize}
    \item \textbf{Training Efficiency:} Current DLMs generally exhibit lower training efficiency compared to AR models, due to factors such as limited token usage during loss computation. Future research could explore hybrid DLM architectures or improved training schemes that match or exceed AR models in efficiency.
    \item \textbf{Quantization and Binarization (Low-bit DLMs):} While extensively studied in AR models, low-bit quantization and binarization remain largely unexplored in DLMs. Adapting these techniques to the diffusion paradigm could yield faster inference and reduced memory consumption, 
    benefiting deployment in real-world systems.
    \item \textbf{Pruning and Distillation:} Model compression techniques such as pruning and knowledge distillation have been successful in reducing model size and inference cost for AR models. Applying these techniques to DLMs could enhance their deployability, especially in resource-constrained or latency-critical environments.

    \item \textbf{Multimodal Unified Reasoning:} Although recent multimodal DLMs demonstrate impressive capabilities in cross-modal understanding and generation, most models are still limited to reasoning within a single modality at a time. Future efforts can focus on building unified DLMs capable of performing complex reasoning across multiple modalities in a truly integrated manner.
    \item \textbf{DLM-based Agents:} The potential of DLMs in powering intelligent agents remains largely underexplored. Leveraging their bidirectional context modeling, parallel decoding, and iterative refinement capabilities, 
    DLM-based agents could offer greater flexibility and adaptability in dynamic environments, 
    making them a promising alternative to traditional AR-based agent approaches.
\end{itemize}
\section{Conclusion}
In this survey, we present an in-depth overview of the entire landscape of diffusion language models.
We outline the fundamental principles, taxonomy, and modeling paradigms of DLMs, and compare them with mainstream autoregressive models, highlighting their unique characteristics and advantages.
We further explore the design space of training and inference, covering various training strategies and inference techniques for both quality and efficiency.
Moreover, we highlight recent advances in multimodal diffusion language models, demonstrating their capabilities in handling diverse data modalities.
Finally, we discuss the limitations and challenges in this field, and outline promising directions for future research.
We hope this survey serves as a comprehensive reference for researchers interested in diffusion-based language modeling, offering valuable insights about the current state of the field and its future prospects.
We also encourage further exploration and innovation in this exciting area of research, as diffusion language models continue to evolve and push the boundaries of language understanding and generation.

\bibliography{main}

% Generated by IEEEtran.bst, version: 1.14 (2015/08/26)
\begin{thebibliography}{100}
\providecommand{\url}[1]{#1}
\csname url@samestyle\endcsname
\providecommand{\newblock}{\relax}
\providecommand{\bibinfo}[2]{#2}
\providecommand{\BIBentrySTDinterwordspacing}{\spaceskip=0pt\relax}
\providecommand{\BIBentryALTinterwordstretchfactor}{4}
\providecommand{\BIBentryALTinterwordspacing}{\spaceskip=\fontdimen2\font plus
\BIBentryALTinterwordstretchfactor\fontdimen3\font minus \fontdimen4\font\relax}
\providecommand{\BIBforeignlanguage}[2]{{%
\expandafter\ifx\csname l@#1\endcsname\relax
\typeout{** WARNING: IEEEtran.bst: No hyphenation pattern has been}%
\typeout{** loaded for the language `#1'. Using the pattern for}%
\typeout{** the default language instead.}%
\else
\language=\csname l@#1\endcsname
\fi
#2}}
\providecommand{\BIBdecl}{\relax}
\BIBdecl

\bibitem{brown2020language}
T.~Brown, B.~Mann, N.~Ryder, M.~Subbiah, J.~D. Kaplan, P.~Dhariwal, A.~Neelakantan, P.~Shyam, G.~Sastry, A.~Askell \emph{et~al.}, ``Language models are few-shot learners,'' \emph{Advances in neural information processing systems}, vol.~33, pp. 1877--1901, 2020.

\bibitem{achiam2023gpt}
J.~Achiam, S.~Adler, S.~Agarwal, L.~Ahmad, I.~Akkaya, F.~L. Aleman, D.~Almeida, J.~Altenschmidt, S.~Altman, S.~Anadkat \emph{et~al.}, ``Gpt-4 technical report,'' \emph{arXiv preprint arXiv:2303.08774}, 2023.

\bibitem{chowdhery2023palm}
A.~Chowdhery, S.~Narang, J.~Devlin, M.~Bosma, G.~Mishra, A.~Roberts, P.~Barham, H.~W. Chung, C.~Sutton, S.~Gehrmann \emph{et~al.}, ``Palm: Scaling language modeling with pathways,'' \emph{Journal of Machine Learning Research}, vol.~24, no. 240, pp. 1--113, 2023.

\bibitem{touvron2023llama}
H.~Touvron, T.~Lavril, G.~Izacard, X.~Martinet, M.-A. Lachaux, T.~Lacroix, B.~Rozi{\`e}re, N.~Goyal, E.~Hambro, F.~Azhar \emph{et~al.}, ``Llama: Open and efficient foundation language models,'' \emph{arXiv preprint arXiv:2302.13971}, 2023.

\bibitem{bai2023qwen}
J.~Bai, S.~Bai, Y.~Chu, Z.~Cui, K.~Dang, X.~Deng, Y.~Fan, W.~Ge, Y.~Han, F.~Huang \emph{et~al.}, ``Qwen technical report,'' \emph{arXiv preprint arXiv:2309.16609}, 2023.

\bibitem{zhao2023survey}
W.~X. Zhao, K.~Zhou, J.~Li, T.~Tang, X.~Wang, Y.~Hou, Y.~Min, B.~Zhang, J.~Zhang, Z.~Dong \emph{et~al.}, ``A survey of large language models,'' \emph{arXiv preprint arXiv:2303.18223}, vol.~1, no.~2, 2023.

\bibitem{guo2025deepseek}
D.~Guo, D.~Yang, H.~Zhang, J.~Song, R.~Zhang, R.~Xu, Q.~Zhu, S.~Ma, P.~Wang, X.~Bi \emph{et~al.}, ``Deepseek-r1: Incentivizing reasoning capability in llms via reinforcement learning,'' \emph{arXiv preprint arXiv:2501.12948}, 2025.

\bibitem{rombach2022high}
R.~Rombach, A.~Blattmann, D.~Lorenz, P.~Esser, and B.~Ommer, ``High-resolution image synthesis with latent diffusion models,'' in \emph{Proceedings of the IEEE/CVF conference on computer vision and pattern recognition}, 2022, pp. 10\,684--10\,695.

\bibitem{saharia2022photorealistic}
C.~Saharia, W.~Chan, S.~Saxena, L.~Li, J.~Whang, E.~L. Denton, K.~Ghasemipour, R.~Gontijo~Lopes, B.~Karagol~Ayan, T.~Salimans \emph{et~al.}, ``Photorealistic text-to-image diffusion models with deep language understanding,'' \emph{Advances in neural information processing systems}, vol.~35, pp. 36\,479--36\,494, 2022.

\bibitem{podellsdxl}
D.~Podell, Z.~English, K.~Lacey, A.~Blattmann, T.~Dockhorn, J.~M{\"u}ller, J.~Penna, and R.~Rombach, ``Sdxl: Improving latent diffusion models for high-resolution image synthesis,'' in \emph{The Twelfth International Conference on Learning Representations}.

\bibitem{esser2024scaling}
P.~Esser, S.~Kulal, A.~Blattmann, R.~Entezari, J.~M{\"u}ller, H.~Saini, Y.~Levi, D.~Lorenz, A.~Sauer, F.~Boesel \emph{et~al.}, ``Scaling rectified flow transformers for high-resolution image synthesis,'' in \emph{Forty-first international conference on machine learning}, 2024.

\bibitem{brooks2024video}
T.~Brooks, B.~Peebles, C.~Holmes, W.~DePue, Y.~Guo, L.~Jing, D.~Schnurr, J.~Taylor, T.~Luhman, E.~Luhman \emph{et~al.}, ``Video generation models as world simulators,'' \emph{OpenAI Blog}, vol.~1, p.~8, 2024.

\bibitem{radford2018improving}
A.~Radford, K.~Narasimhan, T.~Salimans, I.~Sutskever \emph{et~al.}, ``Improving language understanding by generative pre-training,'' 2018.

\bibitem{radford2019language}
A.~Radford, J.~Wu, R.~Child, D.~Luan, D.~Amodei, I.~Sutskever \emph{et~al.}, ``Language models are unsupervised multitask learners,'' \emph{OpenAI blog}, vol.~1, no.~8, p.~9, 2019.

\bibitem{team2023gemini}
G.~Team, R.~Anil, S.~Borgeaud, J.-B. Alayrac, J.~Yu, R.~Soricut, J.~Schalkwyk, A.~M. Dai, A.~Hauth, K.~Millican \emph{et~al.}, ``Gemini: a family of highly capable multimodal models,'' \emph{arXiv preprint arXiv:2312.11805}, 2023.

\bibitem{liu2024deepseek}
A.~Liu, B.~Feng, B.~Xue, B.~Wang, B.~Wu, C.~Lu, C.~Zhao, C.~Deng, C.~Zhang, C.~Ruan \emph{et~al.}, ``Deepseek-v3 technical report,'' \emph{arXiv preprint arXiv:2412.19437}, 2024.

\bibitem{dhariwal2021diffusion}
P.~Dhariwal and A.~Nichol, ``Diffusion models beat gans on image synthesis,'' \emph{Advances in neural information processing systems}, vol.~34, pp. 8780--8794, 2021.

\bibitem{ho2020denoising}
J.~Ho, A.~Jain, and P.~Abbeel, ``Denoising diffusion probabilistic models,'' \emph{Advances in neural information processing systems}, vol.~33, pp. 6840--6851, 2020.

\bibitem{song2020denoising}
J.~Song, C.~Meng, and S.~Ermon, ``Denoising diffusion implicit models,'' \emph{arXiv preprint arXiv:2010.02502}, 2020.

\bibitem{songscore}
Y.~Song, J.~Sohl-Dickstein, D.~P. Kingma, A.~Kumar, S.~Ermon, and B.~Poole, ``Score-based generative modeling through stochastic differential equations,'' in \emph{International Conference on Learning Representations}.

\bibitem{liuflow}
X.~Liu, C.~Gong \emph{et~al.}, ``Flow straight and fast: Learning to generate and transfer data with rectified flow,'' in \emph{The Eleventh International Conference on Learning Representations}.

\bibitem{li2022diffusion}
X.~Li, J.~Thickstun, I.~Gulrajani, P.~S. Liang, and T.~B. Hashimoto, ``Diffusion-lm improves controllable text generation,'' \emph{Advances in neural information processing systems}, vol.~35, pp. 4328--4343, 2022.

\bibitem{strudel2022self}
R.~Strudel, C.~Tallec, F.~Altch{\'e}, Y.~Du, Y.~Ganin, A.~Mensch, W.~Grathwohl, N.~Savinov, S.~Dieleman, L.~Sifre \emph{et~al.}, ``Self-conditioned embedding diffusion for text generation,'' \emph{arXiv preprint arXiv:2211.04236}, 2022.

\bibitem{austin2021structured}
J.~Austin, D.~D. Johnson, J.~Ho, D.~Tarlow, and R.~Van Den~Berg, ``Structured denoising diffusion models in discrete state-spaces,'' \emph{Advances in neural information processing systems}, vol.~34, pp. 17\,981--17\,993, 2021.

\bibitem{he2023diffusionbert}
Z.~He, T.~Sun, Q.~Tang, K.~Wang, X.~Huang, and X.~Qiu, ``Diffusionbert: Improving generative masked language models with diffusion models,'' in \emph{The 61st Annual Meeting Of The Association For Computational Linguistics}, 2023.

\bibitem{dream2025}
\BIBentryALTinterwordspacing
J.~Ye, Z.~Xie, L.~Zheng, J.~Gao, Z.~Wu, X.~Jiang, Z.~Li, and L.~Kong, ``Dream 7b,'' 2025. [Online]. Available: \url{https://hkunlp.github.io/blog/2025/dream}
\BIBentrySTDinterwordspacing

\bibitem{gongscaling}
S.~Gong, S.~Agarwal, Y.~Zhang, J.~Ye, L.~Zheng, M.~Li, C.~An, P.~Zhao, W.~Bi, J.~Han \emph{et~al.}, ``Scaling diffusion language models via adaptation from autoregressive models,'' in \emph{The Thirteenth International Conference on Learning Representations}.

\bibitem{nie2025largelanguagediffusionmodels}
S.~Nie, F.~Zhu, Z.~You, X.~Zhang, J.~Ou, J.~Hu, J.~Zhou, Y.~Lin, J.-R. Wen, and C.~Li, ``Large language diffusion models,'' \emph{arXiv preprint arXiv:2502.09992}, 2025.

\bibitem{you2025llada}
Z.~You, S.~Nie, X.~Zhang, J.~Hu, J.~Zhou, Z.~Lu, J.-R. Wen, and C.~Li, ``Llada-v: Large language diffusion models with visual instruction tuning,'' \emph{arXiv preprint arXiv:2505.16933}, 2025.

\bibitem{yu2025dimple}
R.~Yu, X.~Ma, and X.~Wang, ``Dimple: Discrete diffusion multimodal large language model with parallel decoding,'' \emph{arXiv preprint arXiv:2505.16990}, 2025.

\bibitem{yang2025mmada}
L.~Yang, Y.~Tian, B.~Li, X.~Zhang, K.~Shen, Y.~Tong, and M.~Wang, ``Mmada: Multimodal large diffusion language models,'' \emph{arXiv preprint arXiv:2505.15809}, 2025.

\bibitem{labs2025mercury}
I.~Labs, S.~Khanna, S.~Kharbanda, S.~Li, H.~Varma, E.~Wang, S.~Birnbaum, Z.~Luo, Y.~Miraoui, A.~Palrecha \emph{et~al.}, ``Mercury: Ultra-fast language models based on diffusion,'' \emph{arXiv preprint arXiv:2506.17298}, 2025.

\bibitem{deepmind2024geminidiffusion}
DeepMind, ``Gemini diffusion,'' \url{https://deepmind.google/technologies/gemini}, 2024, accessed: 2025-07-09.

\bibitem{song2025seed}
Y.~Song, Z.~Zhang, C.~Luo, P.~Gao, F.~Xia, H.~Luo, Z.~Li, Y.~Yang, H.~Yu, X.~Qu \emph{et~al.}, ``Seed diffusion: A large-scale diffusion language model with high-speed inference,'' \emph{arXiv preprint arXiv:2508.02193}, 2025.

\bibitem{xu2024energy}
M.~Xu, T.~Geffner, K.~Kreis, W.~Nie, Y.~Xu, J.~Leskovec, S.~Ermon, and A.~Vahdat, ``Energy-based diffusion language models for text generation,'' \emph{arXiv preprint arXiv:2410.21357}, 2024.

\bibitem{deschenauxbeyond}
J.~Deschenaux and C.~Gulcehre, ``Beyond autoregression: Fast llms via self-distillation through time,'' in \emph{The Thirteenth International Conference on Learning Representations}.

\bibitem{han2024transfer}
K.~Han, K.~Kenealy, A.~Barua, N.~Fiedel, and N.~Constant, ``Transfer learning for text diffusion models,'' \emph{arXiv preprint arXiv:2401.17181}, 2024.

\bibitem{sahoo2025diffusion}
S.~S. Sahoo, J.~Deschenaux, A.~Gokaslan, G.~Wang, J.~Chiu, and V.~Kuleshov, ``The diffusion duality,'' \emph{arXiv preprint arXiv:2506.10892}, 2025.

\bibitem{zhang2025non}
Y.~Zhang, S.~He, D.~Levine, L.~Zhao, D.~Zhang, S.~A. Rizvi, E.~Zappala, R.~Ying, and D.~van Dijk, ``Non-markovian discrete diffusion with causal language models,'' \emph{arXiv preprint arXiv:2502.09767}, 2025.

\bibitem{dang2025inference}
M.~Dang, J.~Han, M.~Xu, K.~Xu, A.~Srivastava, and S.~Ermon, ``Inference-time scaling of diffusion language models with particle gibbs sampling,'' \emph{arXiv preprint arXiv:2507.08390}, 2025.

\bibitem{rout2025anchored}
L.~Rout, C.~Caramanis, and S.~Shakkottai, ``Anchored diffusion language model,'' \emph{arXiv preprint arXiv:2505.18456}, 2025.

\bibitem{shao2024deepseekmath}
Z.~Shao, P.~Wang, Q.~Zhu, R.~Xu, J.~Song, X.~Bi, H.~Zhang, M.~Zhang, Y.~Li, Y.~Wu \emph{et~al.}, ``Deepseekmath: Pushing the limits of mathematical reasoning in open language models,'' \emph{arXiv preprint arXiv:2402.03300}, 2024.

\bibitem{zhao2025d1}
S.~Zhao, D.~Gupta, Q.~Zheng, and A.~Grover, ``d1: Scaling reasoning in diffusion large language models via reinforcement learning,'' \emph{arXiv preprint arXiv:2504.12216}, 2025.

\bibitem{chen2025dlm}
T.~Chen, S.~Zhang, and M.~Zhou, ``Dlm-one: Diffusion language models for one-step sequence generation,'' \emph{arXiv e-prints}, pp. arXiv--2506, 2025.

\bibitem{wu2025fast}
C.~Wu, H.~Zhang, S.~Xue, Z.~Liu, S.~Diao, L.~Zhu, P.~Luo, S.~Han, and E.~Xie, ``Fast-dllm: Training-free acceleration of diffusion llm by enabling kv cache and parallel decoding,'' \emph{arXiv preprint arXiv:2505.22618}, 2025.

\bibitem{israel2025accelerating}
D.~Israel, G.~V.~d. Broeck, and A.~Grover, ``Accelerating diffusion llms via adaptive parallel decoding,'' \emph{arXiv preprint arXiv:2506.00413}, 2025.

\bibitem{wang2025remasking}
G.~Wang, Y.~Schiff, S.~S. Sahoo, and V.~Kuleshov, ``Remasking discrete diffusion models with inference-time scaling,'' in \emph{ICLR 2025 Workshop on Deep Generative Model in Machine Learning: Theory, Principle and Efficacy}.

\bibitem{liu2025dllm}
Z.~Liu, Y.~Yang, Y.~Zhang, J.~Chen, C.~Zou, Q.~Wei, S.~Wang, and L.~Zhang, ``dllm-cache: Accelerating diffusion large language models with adaptive caching,'' \emph{arXiv preprint arXiv:2506.06295}, 2025.

\bibitem{ma2025dkv}
X.~Ma, R.~Yu, G.~Fang, and X.~Wang, ``dkv-cache: The cache for diffusion language models,'' \emph{arXiv preprint arXiv:2505.15781}, 2025.

\bibitem{liu2022composable}
G.~Liu, Z.~Feng, Y.~Gao, Z.~Yang, X.~Liang, J.~Bao, X.~He, S.~Cui, Z.~Li, and Z.~Hu, ``Composable text controls in latent space with odes,'' \emph{arXiv preprint arXiv:2208.00638}, 2022.

\bibitem{gongdiffuseq}
S.~Gong, M.~Li, J.~Feng, Z.~Wu, and L.~Kong, ``Diffuseq: Sequence to sequence text generation with diffusion models,'' in \emph{The Eleventh International Conference on Learning Representations}.

\bibitem{dieleman2022continuous}
S.~Dieleman, L.~Sartran, A.~Roshannai, N.~Savinov, Y.~Ganin, P.~H. Richemond, A.~Doucet, R.~Strudel, C.~Dyer, C.~Durkan \emph{et~al.}, ``Continuous diffusion for categorical data,'' \emph{arXiv preprint arXiv:2211.15089}, 2022.

\bibitem{gao2022empowering}
Z.~Gao, J.~Guo, X.~Tan, Y.~Zhu, F.~Zhang, J.~Bian, and L.~Xu, ``Empowering diffusion models on the embedding space for text generation,'' \emph{arXiv preprint arXiv:2212.09412}, 2022.

\bibitem{lovelace2023latent}
J.~Lovelace, V.~Kishore, C.~Wan, E.~Shekhtman, and K.~Q. Weinberger, ``Latent diffusion for language generation,'' \emph{Advances in Neural Information Processing Systems}, vol.~36, pp. 56\,998--57\,025, 2023.

\bibitem{lin2023text}
Z.~Lin, Y.~Gong, Y.~Shen, T.~Wu, Z.~Fan, C.~Lin, N.~Duan, and W.~Chen, ``Text generation with diffusion language models: A pre-training approach with continuous paragraph denoise,'' in \emph{International Conference on Machine Learning}.\hskip 1em plus 0.5em minus 0.4em\relax PMLR, 2023, pp. 21\,051--21\,064.

\bibitem{wang2023infodiffusion}
R.~Wang, J.~Li, and P.~Li, ``Infodiffusion: Information entropy aware diffusion process for non-autoregressive text generation,'' in \emph{Findings of the Association for Computational Linguistics: EMNLP 2023}, 2023, pp. 13\,757--13\,770.

\bibitem{liu2024unified}
G.~Liu, Y.~Wang, Z.~Feng, Q.~Wu, L.~Tang, Y.~Gao, Z.~Li, S.~Cui, J.~Mcauley, Z.~Yang \emph{et~al.}, ``Unified generation, reconstruction, and representation: Generalized diffusion with adaptive latent encoding-decoding,'' in \emph{International Conference on Machine Learning}.\hskip 1em plus 0.5em minus 0.4em\relax PMLR, 2024, pp. 31\,964--31\,993.

\bibitem{shabalin2025smoothie}
A.~Shabalin, V.~Meshchaninov, and D.~Vetrov, ``Smoothie: Smoothing diffusion on token embeddings for text generation,'' \emph{arXiv preprint arXiv:2505.18853}, 2025.

\bibitem{mahabadi2024tess}
R.~K. Mahabadi, H.~Ivison, J.~Tae, J.~Henderson, I.~Beltagy, M.~E. Peters, and A.~Cohan, ``Tess: Text-to-text self-conditioned simplex diffusion,'' in \emph{Proceedings of the 18th Conference of the European Chapter of the Association for Computational Linguistics (Volume 1: Long Papers)}, 2024, pp. 2347--2361.

\bibitem{tae2025tess}
J.~Tae, H.~Ivison, S.~Kumar, and A.~Cohan, ``Tess 2: A large-scale generalist diffusion language model,'' \emph{arXiv preprint arXiv:2502.13917}, 2025.

\bibitem{yu2022latent}
P.~Yu, S.~Xie, X.~Ma, B.~Jia, B.~Pang, R.~Gao, Y.~Zhu, S.-C. Zhu, and Y.~N. Wu, ``Latent diffusion energy-based model for interpretable text modelling,'' in \emph{International Conference on Machine Learning}.\hskip 1em plus 0.5em minus 0.4em\relax PMLR, 2022, pp. 25\,702--25\,720.

\bibitem{chen2026langflow}
Y.~Chen, C.~Liang, H.~Sui, R.~Guo, C.~Cheng, J.~You, and G.~Liu, ``Langflow: Continuous diffusion rivals discrete in language modeling,'' \emph{arXiv preprint arXiv:2604.11748}, 2026.

\bibitem{hu2026elf}
K.~Hu, L.~Qiu, Y.~Lu, H.~Zhao, T.~Li, Y.~Kim, J.~Andreas, and K.~He, ``Elf: Embedded language flows,'' \emph{arXiv preprint arXiv:2605.10938}, 2026.

\bibitem{zhuang2026bitlm}
S.~Zhuang, Y.~Ai, J.~Han, X.~Li, H.~Huang, X.~Yue, X.~Hu, K.~Xu, Y.~Wang, and H.~Chen, ``Bitlm: Unlocking multi-token language generation with bitwise continuous diffusion,'' \emph{arXiv preprint arXiv:2605.11577}, 2026.

\bibitem{lee2026flow}
C.~Lee, J.~Yoo, M.~Agarwal, S.~Shah, J.~Huang, A.~Raghunathan, S.~Hong, N.~M. Boffi, and J.~Kim, ``Flow map language models: One-step language modeling via continuous denoising,'' \emph{arXiv preprint arXiv:2602.16813}, 2026.

\bibitem{guo2026continuous}
H.~Guo, Q.~Zhao, Y.~Zhao, S.~Nie, R.~Zhu, Q.~Guo, F.~Wang, T.~Yang, H.~Zhao, G.~Wei \emph{et~al.}, ``Continuous latent diffusion language model,'' \emph{arXiv preprint arXiv:2605.06548}, 2026.

\bibitem{zhengreparameterized}
L.~Zheng, J.~Yuan, L.~Yu, and L.~Kong, ``A reparameterized discrete diffusion model for text generation,'' in \emph{First Conference on Language Modeling}.

\bibitem{shi2024simplified}
J.~Shi, K.~Han, Z.~Wang, A.~Doucet, and M.~Titsias, ``Simplified and generalized masked diffusion for discrete data,'' \emph{Advances in neural information processing systems}, vol.~37, pp. 103\,131--103\,167, 2024.

\bibitem{sahoo2024simple}
S.~Sahoo, M.~Arriola, Y.~Schiff, A.~Gokaslan, E.~Marroquin, J.~Chiu, A.~Rush, and V.~Kuleshov, ``Simple and effective masked diffusion language models,'' \emph{Advances in Neural Information Processing Systems}, vol.~37, pp. 130\,136--130\,184, 2024.

\bibitem{ye2023diffusion}
J.~Ye, Z.~Zheng, Y.~Bao, L.~Qian, and Q.~Gu, ``Diffusion language models can perform many tasks with scaling and instruction-finetuning,'' \emph{arXiv preprint arXiv:2308.12219}, 2023.

\bibitem{zhou2024diffusion}
K.~Zhou, Y.~Li, W.~X. Zhao, and J.-R. Wen, ``Diffusion-nat: Self-prompting discrete diffusion for non-autoregressive text generation,'' in \emph{Proceedings of the 18th Conference of the European Chapter of the Association for Computational Linguistics (Volume 1: Long Papers)}, 2024, pp. 1438--1451.

\bibitem{gulrajani2023likelihood}
I.~Gulrajani and T.~B. Hashimoto, ``Likelihood-based diffusion language models,'' \emph{Advances in Neural Information Processing Systems}, vol.~36, pp. 16\,693--16\,715, 2023.

\bibitem{lou2024discrete}
A.~Lou, C.~Meng, and S.~Ermon, ``Discrete diffusion modeling by estimating the ratios of the data distribution,'' in \emph{International Conference on Machine Learning}.\hskip 1em plus 0.5em minus 0.4em\relax PMLR, 2024, pp. 32\,819--32\,848.

\bibitem{ou2024your}
J.~Ou, S.~Nie, K.~Xue, F.~Zhu, J.~Sun, Z.~Li, and C.~Li, ``Your absorbing discrete diffusion secretly models the conditional distributions of clean data,'' in \emph{The Thirteenth International Conference on Learning Representations}, 2024.

\bibitem{gat2024discrete}
I.~Gat, T.~Remez, N.~Shaul, F.~Kreuk, R.~T. Chen, G.~Synnaeve, Y.~Adi, and Y.~Lipman, ``Discrete flow matching,'' \emph{Advances in Neural Information Processing Systems}, vol.~37, pp. 133\,345--133\,385, 2024.

\bibitem{liu2024think}
S.~Liu, J.~Nam, A.~Campbell, H.~Stark, Y.~Xu, T.~Jaakkola, and R.~Gomez-Bombarelli, ``Think while you generate: Discrete diffusion with planned denoising,'' in \emph{The Thirteenth International Conference on Learning Representations}, 2024.

\bibitem{ye2024beyond}
J.~Ye, J.~Gao, S.~Gong, L.~Zheng, X.~Jiang, Z.~Li, and L.~Kong, ``Beyond autoregression: Discrete diffusion for complex reasoning and planning,'' \emph{arXiv preprint arXiv:2410.14157}, 2024.

\bibitem{von2025generalized}
D.~von R{\"u}tte, J.~Fluri, Y.~Ding, A.~Orvieto, B.~Sch{\"o}lkopf, and T.~Hofmann, ``Generalized interpolating discrete diffusion,'' in \emph{Forty-second International Conference on Machine Learning}, 2025.

\bibitem{liu2025longllada}
X.~Liu, Z.~Liu, Z.~Huang, Q.~Guo, Z.~He, and X.~Qiu, ``Longllada: Unlocking long context capabilities in diffusion llms,'' \emph{arXiv preprint arXiv:2506.14429}, 2025.

\bibitem{zhu2025lladamoe}
F.~Zhu, Z.~You, Y.~Xing, Z.~Huang, L.~Liu, Y.~Zhuang, G.~Lu, K.~Wang, X.~Wang, L.~Wei \emph{et~al.}, ``Llada-moe: A sparse moe diffusion language model,'' \emph{arXiv preprint arXiv:2509.24389}, 2025.

\bibitem{bie2025llada2}
T.~Bie, M.~Cao, K.~Chen, L.~Du, M.~Gong, Z.~Gong, Y.~Gu, J.~Hu, Z.~Huang, Z.~Lan \emph{et~al.}, ``Llada2. 0: Scaling up diffusion language models to 100b,'' \emph{arXiv preprint arXiv:2512.15745}, 2025.

\bibitem{han2023ssd}
X.~Han, S.~Kumar, and Y.~Tsvetkov, ``Ssd-lm: Semi-autoregressive simplex-based diffusion language model for text generation and modular control,'' in \emph{Proceedings of the 61st Annual Meeting of the Association for Computational Linguistics (Volume 1: Long Papers)}, 2023, pp. 11\,575--11\,596.

\bibitem{wu2023ar}
T.~Wu, Z.~Fan, X.~Liu, H.-T. Zheng, Y.~Gong, J.~Jiao, J.~Li, J.~Guo, N.~Duan, W.~Chen \emph{et~al.}, ``Ar-diffusion: Auto-regressive diffusion model for text generation,'' \emph{Advances in Neural Information Processing Systems}, vol.~36, pp. 39\,957--39\,974, 2023.

\bibitem{arriolablock}
M.~Arriola, A.~Gokaslan, J.~T. Chiu, Z.~Yang, Z.~Qi, J.~Han, S.~S. Sahoo, and V.~Kuleshov, ``Block diffusion: Interpolating between autoregressive and diffusion language models,'' in \emph{The Thirteenth International Conference on Learning Representations}.

\bibitem{huang2025ctrldiff}
C.~Huang and H.~Tang, ``Ctrldiff: Boosting large diffusion language models with dynamic block prediction and controllable generation,'' \emph{arXiv preprint arXiv:2505.14455}, 2025.

\bibitem{christopher2025speculative}
J.~K. Christopher, B.~R. Bartoldson, T.~Ben-Nun, M.~Cardei, B.~Kailkhura, and F.~Fioretto, ``Speculative diffusion decoding: Accelerating language generation through diffusion,'' in \emph{Proceedings of the 2025 Conference of the Nations of the Americas Chapter of the Association for Computational Linguistics: Human Language Technologies (Volume 1: Long Papers)}, 2025, pp. 12\,042--12\,059.

\bibitem{cheng2025sdar}
S.~Cheng, Y.~Bian, D.~Liu, L.~Zhang, Q.~Yao, Z.~Tian, W.~Wang, Q.~Guo, K.~Chen, B.~Qi \emph{et~al.}, ``Sdar: A synergistic diffusion-autoregression paradigm for scalable sequence generation,'' \emph{arXiv preprint arXiv:2510.06303}, 2025.

\bibitem{liu2025tidar}
J.~Liu, X.~Dong, Z.~Ye, R.~Mehta, Y.~Fu, V.~Singh, J.~Kautz, C.~Zhang, and P.~Molchanov, ``Tidar: Think in diffusion, talk in autoregression,'' \emph{arXiv preprint arXiv:2511.08923}, 2025.

\bibitem{liu2025sequential}
Y.~Liu, Y.~Cao, H.~Li, G.~Luo, Z.~Chen, W.~Wang, X.~Liang, B.~Qi, L.~Wu, C.~Tian \emph{et~al.}, ``Sequential diffusion language models,'' \emph{arXiv preprint arXiv:2509.24007}, 2025.

\bibitem{tian2025next}
Y.~Tian, Y.~Liang, S.~Zhang, Y.~Shu, G.~Yang, W.~He, S.~Fang, T.~Guo, K.~Han, C.~Xu \emph{et~al.}, ``From next-token to next-block: A principled adaptation path for diffusion llms,'' \emph{arXiv preprint arXiv:2512.06776}, 2025.

\bibitem{fu2025efficient}
Y.~Fu, L.~Whalen, Z.~Ye, X.~Dong, S.~Diao, J.~Liu, C.~Wu, H.~Zhang, E.~Xie, S.~Han \emph{et~al.}, ``Efficient-dlm: From autoregressive to diffusion language models, and beyond in speed,'' \emph{arXiv preprint arXiv:2512.14067}, 2025.

\bibitem{yu2026introspective}
Y.~Yu, Y.~Jian, J.~Wang, Z.~Zhou, D.~Zhuang, X.~Fang, S.~Yanamandra, X.~Wu, Q.~Wu, S.~L. Song \emph{et~al.}, ``Introspective diffusion language models,'' \emph{arXiv preprint arXiv:2604.11035}, 2026.

\bibitem{li2025refusion}
J.-N. Li, J.~Guan, W.~Wu, and C.~Li, ``Refusion: A diffusion large language model with parallel autoregressive decoding,'' \emph{arXiv preprint arXiv:2512.13586}, 2025.

\bibitem{ruan2026causal}
J.~Ruan, B.~Li, Y.~Yin, P.~Huang, X.~Chen, J.~Wang, X.~Cai, T.~Xiao, and J.~Zhu, ``Causal autoregressive diffusion language model,'' \emph{arXiv preprint arXiv:2601.22031}, 2026.

\bibitem{li2025dual}
Z.~Li, H.~Li, Y.~Shi, A.~B. Farimani, Y.~Kluger, L.~Yang, and P.~Wang, ``Dual diffusion for unified image generation and understanding,'' in \emph{Proceedings of the Computer Vision and Pattern Recognition Conference}, 2025, pp. 2779--2790.

\bibitem{shi2025muddit}
Q.~Shi, J.~Bai, Z.~Zhao, W.~Chai, K.~Yu, J.~Wu, S.~Song, Y.~Tong, X.~Li, X.~Li \emph{et~al.}, ``Muddit: Liberating generation beyond text-to-image with a unified discrete diffusion model,'' \emph{arXiv preprint arXiv:2505.23606}, 2025.

\bibitem{ye2024diffusion}
J.~Ye, S.~Gong, L.~Chen, L.~Zheng, J.~Gao, H.~Shi, C.~Wu, X.~Jiang, Z.~Li, W.~Bi \emph{et~al.}, ``Diffusion of thought: Chain-of-thought reasoning in diffusion language models,'' \emph{Advances in Neural Information Processing Systems}, vol.~37, pp. 105\,345--105\,374, 2024.

\bibitem{huang2025reinforcing}
Z.~Huang, Z.~Chen, Z.~Wang, T.~Li, and G.-J. Qi, ``Reinforcing the diffusion chain of lateral thought with diffusion language models,'' \emph{arXiv preprint arXiv:2505.10446}, 2025.

\bibitem{zekri2025fine}
O.~Zekri and N.~Boull{\'e}, ``Fine-tuning discrete diffusion models with policy gradient methods,'' \emph{arXiv preprint arXiv:2502.01384}, 2025.

\bibitem{gong2025diffucoder}
S.~Gong, R.~Zhang, H.~Zheng, J.~Gu, N.~Jaitly, L.~Kong, and Y.~Zhang, ``Diffucoder: Understanding and improving masked diffusion models for code generation,'' \emph{arXiv preprint arXiv:2506.20639}, 2025.

\bibitem{tang2025wd1}
X.~Tang, R.~Dolga, S.~Yoon, and I.~Bogunovic, ``wd1: Weighted policy optimization for reasoning in diffusion language models,'' \emph{arXiv preprint arXiv:2507.08838}, 2025.

\bibitem{zhao2025inpainting}
S.~Zhao, M.~Liu, J.~Huang, M.~Liu, C.~Wang, B.~Liu, Y.~Tian, G.~Pang, S.~Bell, A.~Grover \emph{et~al.}, ``Inpainting-guided policy optimization for diffusion large language models,'' \emph{arXiv preprint arXiv:2509.10396}, 2025.

\bibitem{wang2025spg}
C.~Wang, P.~Rashidinejad, D.~Su, S.~Jiang, S.~Wang, S.~Zhao, C.~Zhou, S.~Z. Shen, F.~Chen, T.~Jaakkola \emph{et~al.}, ``Spg: Sandwiched policy gradient for masked diffusion language models,'' \emph{arXiv preprint arXiv:2510.09541}, 2025.

\bibitem{xie2025step}
S.~Xie, L.~Kong, X.~Song, X.~Dong, G.~Chen, E.~P. Xing, and K.~Zhang, ``Step-aware policy optimization for reasoning in diffusion large language models,'' \emph{arXiv preprint arXiv:2510.01544}, 2025.

\bibitem{lin2025boundary}
N.~Lin, J.~Zhang, L.~Hou, and J.~Li, ``Boundary-guided policy optimization for memory-efficient rl of diffusion large language models,'' \emph{arXiv preprint arXiv:2510.11683}, 2025.

\bibitem{ni2026flexibility}
Z.~Ni, S.~Wang, Y.~Yue, T.~Yu, W.~Zhao, Y.~Hua, T.~Chen, J.~Song, C.~Yu, B.~Zheng \emph{et~al.}, ``The flexibility trap: Why arbitrary order limits reasoning potential in diffusion language models,'' \emph{arXiv preprint arXiv:2601.15165}, 2026.

\bibitem{zhu2025llada}
F.~Zhu, R.~Wang, S.~Nie, X.~Zhang, C.~Wu, J.~Hu, J.~Zhou, J.~Chen, Y.~Lin, J.-R. Wen \emph{et~al.}, ``Llada 1.5: Variance-reduced preference optimization for large language diffusion models,'' \emph{arXiv preprint arXiv:2505.19223}, 2025.

\bibitem{wei2025accelerating}
Q.~Wei, Y.~Zhang, Z.~Liu, D.~Liu, and L.~Zhang, ``Accelerating diffusion large language models with slowfast: The three golden principles,'' \emph{arXiv preprint arXiv:2506.10848}, 2025.

\bibitem{bao2025learning}
W.~Bao, Z.~Chen, D.~Xu, and Y.~Shang, ``Learning to parallel: Accelerating diffusion large language models via learnable parallel decoding,'' \emph{arXiv preprint arXiv:2509.25188}, 2025.

\bibitem{chen2025dparallel}
Z.~Chen, G.~Fang, X.~Ma, R.~Yu, and X.~Wang, ``dparallel: Learnable parallel decoding for dllms,'' \emph{arXiv preprint arXiv:2509.26488}, 2025.

\bibitem{chen2026dflash}
J.~Chen, Y.~Liang, and Z.~Liu, ``Dflash: Block diffusion for flash speculative decoding,'' \emph{arXiv preprint arXiv:2602.06036}, 2026.

\bibitem{chen2026dmax}
Z.~Chen, G.~Fang, X.~Ma, R.~Yu, and X.~Wang, ``Dmax: Aggressive parallel decoding for dllms,'' \emph{arXiv preprint arXiv:2604.08302}, 2026.

\bibitem{li2026diffusion}
P.~Li, D.~Muhtar, T.~Chen, L.~Yin, and S.~Liu, ``Why diffusion language models struggle with truly parallel (non-autoregressive) decoding?'' \emph{arXiv preprint arXiv:2602.23225}, 2026.

\bibitem{li2025adaptive}
P.~Li, S.~Yan, J.~Tsai, R.~Zhang, R.~An, Z.~Guo, and X.~Gao, ``Adaptive classifier-free guidance via dynamic low-confidence masking,'' \emph{arXiv preprint arXiv:2505.20199}, 2025.

\bibitem{hu2025accelerating}
Z.~Hu, J.~Meng, Y.~Akhauri, M.~S. Abdelfattah, J.-s. Seo, Z.~Zhang, and U.~Gupta, ``Accelerating diffusion language model inference via efficient kv caching and guided diffusion,'' \emph{arXiv preprint arXiv:2505.21467}, 2025.

\bibitem{suresh2025dingo}
T.~Suresh, D.~Banerjee, S.~Ugare, S.~Misailovic, and G.~Singh, ``Dingo: Constrained inference for diffusion llms,'' in \emph{ICML 2025 Workshop on Reliable and Responsible Foundation Models}.

\bibitem{nguyen2025attention}
Q.~Nguyen-Tri, M.~Ranjan, and Z.~Shen, ``Attention is all you need for kv cache in diffusion llms,'' \emph{arXiv preprint arXiv:2510.14973}, 2025.

\bibitem{jiang2025d}
Y.~Jiang, Y.~Cai, X.~Luo, J.~Fu, J.~Wang, C.~Liu, and X.~Yang, ``d $^2$ cache: Accelerating diffusion-based llms via dual adaptive caching,'' \emph{arXiv preprint arXiv:2509.23094}, 2025.

\bibitem{ma2024deepcache}
X.~Ma, G.~Fang, and X.~Wang, ``Deepcache: Accelerating diffusion models for free,'' in \emph{Proceedings of the IEEE/CVF conference on computer vision and pattern recognition}, 2024, pp. 15\,762--15\,772.

\bibitem{chen2024delta}
P.~Chen, M.~Shen, P.~Ye, J.~Cao, C.~Tu, C.-S. Bouganis, Y.~Zhao, and T.~Chen, ``{$\Delta$}-dit: A training-free acceleration method tailored for diffusion transformers,'' \emph{arXiv preprint arXiv:2406.01125}, 2024.

\bibitem{ma2024learning}
X.~Ma, G.~Fang, M.~Bi~Mi, and X.~Wang, ``Learning-to-cache: Accelerating diffusion transformer via layer caching,'' \emph{Advances in Neural Information Processing Systems}, vol.~37, pp. 133\,282--133\,304, 2024.

\bibitem{lvfastercache}
Z.~Lv, C.~Si, J.~Song, Z.~Yang, Y.~Qiao, Z.~Liu, and K.-Y.~K. Wong, ``Fastercache: Training-free video diffusion model acceleration with high quality,'' in \emph{The Thirteenth International Conference on Learning Representations}.

\bibitem{hayakawadistillation}
S.~Hayakawa, Y.~Takida, M.~Imaizumi, H.~Wakaki, and Y.~Mitsufuji, ``Distillation of discrete diffusion through dimensional correlations,'' in \emph{Forty-second International Conference on Machine Learning}.

\bibitem{salimansprogressive}
T.~Salimans and J.~Ho, ``Progressive distillation for fast sampling of diffusion models,'' in \emph{International Conference on Learning Representations}.

\bibitem{qian2026d3llm}
Y.-Y. Qian, J.~Su, L.~Hu, P.~Zhang, Z.~Deng, P.~Zhao, and H.~Zhang, ``d3llm: Ultra-fast diffusion llm using pseudo-trajectory distillation,'' \emph{arXiv preprint arXiv:2601.07568}, 2026.

\bibitem{myrzakhan2026sink}
A.~Myrzakhan, T.~Li, B.~Guo, S.~Tang, and Z.~Shen, ``Sink-aware pruning for diffusion language models,'' \emph{arXiv preprint arXiv:2602.17664}, 2026.

\bibitem{li2025lavida}
S.~Li, K.~Kallidromitis, H.~Bansal, A.~Gokul, Y.~Kato, K.~Kozuka, J.~Kuen, Z.~Lin, K.-W. Chang, and A.~Grover, ``Lavida: A large diffusion language model for multimodal understanding,'' \emph{arXiv preprint arXiv:2505.16839}, 2025.

\bibitem{wang2025fudoki}
J.~Wang, Y.~Lai, A.~Li, S.~Zhang, J.~Sun, N.~Kang, C.~Wu, Z.~Li, and P.~Luo, ``Fudoki: Discrete flow-based unified understanding and generation via kinetic-optimal velocities,'' \emph{arXiv preprint arXiv:2505.20147}, 2025.

\bibitem{swerdlow2025unified}
A.~Swerdlow, M.~Prabhudesai, S.~Gandhi, D.~Pathak, and K.~Fragkiadaki, ``Unified multimodal discrete diffusion,'' \emph{arXiv preprint arXiv:2503.20853}, 2025.

\bibitem{xin2025lumina}
Y.~Xin, Q.~Qin, S.~Luo, K.~Zhu, J.~Yan, Y.~Tai, J.~Lei, Y.~Cao, K.~Wang, Y.~Wang \emph{et~al.}, ``Lumina-dimoo: An omni diffusion large language model for multi-modal generation and understanding,'' \emph{arXiv preprint arXiv:2510.06308}, 2025.

\bibitem{li2025lavidao}
S.~Li, J.~Gu, K.~Liu, Z.~Lin, Z.~Wei, A.~Grover, and J.~Kuen, ``Lavida-o: Elastic large masked diffusion models for unified multimodal understanding and generation,'' \emph{arXiv preprint arXiv:2509.19244}, 2025.

\bibitem{tian2025mmada}
Y.~Tian, L.~Yang, J.~Yang, A.~Wang, Y.~Tian, J.~Zheng, H.~Wang, Z.~Teng, Z.~Wang, Y.~Wang \emph{et~al.}, ``Mmada-parallel: Multimodal large diffusion language models for thinking-aware editing and generation,'' \emph{arXiv preprint arXiv:2511.09611}, 2025.

\bibitem{zeng2025diffusionvl}
L.~Zeng, J.~Yao, B.~Liao, H.~Tao, W.~Liu, and X.~Wang, ``Diffusionvl: Translating any autoregressive models into diffusion vision language models,'' \emph{arXiv preprint arXiv:2512.15713}, 2025.

\bibitem{wu2026fast}
C.~Wu, S.~Lan, Y.~Fu, S.~Gao, J.~Wang, J.~Yu, J.~M. Alvarez, P.~Molchanov, P.~Luo, S.~Han \emph{et~al.}, ``Fast-dvlm: Efficient block-diffusion vlm via direct conversion from autoregressive vlm,'' \emph{arXiv preprint arXiv:2604.06832}, 2026.

\bibitem{he2026vidlada}
Z.~He, T.~Chen, K.~Wang, Z.~Qin, Y.~Shao, C.~Gan, S.~Li, Z.~Wu, and W.~Lin, ``Vidlada: Bidirectional diffusion large language models for efficient video understanding,'' \emph{arXiv preprint arXiv:2601.17868}, 2026.

\bibitem{yuan2024roic}
S.~Yuan, W.~Yuan, H.~Yin, and T.~He, ``Roic-dm: Robust text inference and classification via diffusion model,'' \emph{arXiv preprint arXiv:2401.03514}, 2024.

\bibitem{shen2023diffusionner}
Y.~Shen, K.~Song, X.~Tan, D.~Li, W.~Lu, and Y.~Zhuang, ``Diffusionner: Boundary diffusion for named entity recognition,'' in \emph{Proceedings of the 61st Annual Meeting of the Association for Computational Linguistics (Volume 1: Long Papers)}, 2023, pp. 3875--3890.

\bibitem{yang2025ipad}
X.~Yang, Z.~Qiao, and Y.~Zhou, ``Ipad: Iterative, parallel, and diffusion-based network for scene text recognition,'' \emph{International Journal of Computer Vision}, pp. 1--21, 2025.

\bibitem{liu2024let}
S.~Liu, J.~Zhou, Q.~Zhu, Q.~Chen, Q.~Bai, J.~Xiao, and L.~He, ``Let’s rectify step by step: Improving aspect-based sentiment analysis with diffusion models,'' in \emph{Proceedings of the 2024 Joint International Conference on Computational Linguistics, Language Resources and Evaluation (LREC-COLING 2024)}, 2024, pp. 10\,324--10\,335.

\bibitem{zhang2023diffusum}
H.~Zhang, X.~Liu, and J.~Zhang, ``Diffusum: Generation enhanced extractive summarization with diffusion,'' in \emph{Findings of the Association for Computational Linguistics: ACL 2023}, 2023, pp. 13\,089--13\,100.

\bibitem{dong2025termdiffusum}
X.~Dong, W.~Li, Y.~Le, Z.~Jiang, J.~Zhong, and Z.~Wang, ``Termdiffusum: a term-guided diffusion model for extractive summarization of legal documents,'' in \emph{Proceedings of the 31st international conference on computational linguistics}, 2025, pp. 3222--3235.

\bibitem{luo2024enhancing}
Y.~Luo, Q.~Zhou, and F.~Zhou, ``Enhancing phrase representation by information bottleneck guided text diffusion process for keyphrase extraction,'' in \emph{Proceedings of the 2024 Joint International Conference on Computational Linguistics, Language Resources and Evaluation (LREC-COLING 2024)}, 2024, pp. 6036--6047.

\bibitem{zhao2024iped}
J.~Zhao, C.~Xu, and B.~Jiang, ``Iped: An implicit perspective for relational triple extraction based on diffusion model,'' in \emph{Proceedings of the 2024 Conference of the North American Chapter of the Association for Computational Linguistics: Human Language Technologies (Volume 1: Long Papers)}, 2024, pp. 2080--2092.

\bibitem{lee2025editext}
C.~H. Lee, H.~Kim, J.~Yeom, and S.~Yoon, ``Editext: Controllable coarse-to-fine text editing with diffusion language models,'' \emph{arXiv preprint arXiv:2502.19765}, 2025.

\bibitem{bi2023diffusemp}
G.~Bi, L.~Shen, Y.~Cao, M.~Chen, Y.~Xie, Z.~Lin, and X.~He, ``Diffusemp: A diffusion model-based framework with multi-grained control for empathetic response generation,'' in \emph{Proceedings of the 61st Annual Meeting of the Association for Computational Linguistics (Volume 1: Long Papers)}, 2023, pp. 2812--2831.

\bibitem{floto2023diffudetox}
G.~Floto, M.~M.~A. Pour, P.~Farinneya, Z.~Tang, A.~Pesaranghader, M.~Bharadwaj, and S.~Sanner, ``Diffudetox: A mixed diffusion model for text detoxification,'' in \emph{Findings of the Association for Computational Linguistics: ACL 2023}, 2023, pp. 7566--7574.

\bibitem{horvitz2024paraguide}
Z.~Horvitz, A.~Patel, C.~Callison-Burch, Z.~Yu, and K.~McKeown, ``Paraguide: Guided diffusion paraphrasers for plug-and-play textual style transfer,'' in \emph{Proceedings of the AAAI conference on artificial intelligence}, vol.~38, no.~16, 2024, pp. 18\,216--18\,224.

\bibitem{zhang2023planner}
Y.~Zhang, J.~Gu, Z.~Wu, S.~Zhai, J.~Susskind, and N.~Jaitly, ``Planner: Generating diversified paragraph via latent language diffusion model,'' \emph{Advances in Neural Information Processing Systems}, vol.~36, pp. 80\,178--80\,190, 2023.

\bibitem{liu2023diffucom}
J.~Liu, P.~Cheng, J.~Dai, and J.~Liu, ``Diffucom: A novel diffusion model for comment generation,'' \emph{Knowledge-Based Systems}, vol. 281, p. 111069, 2023.

\bibitem{xiang2024diffusiondialog}
J.~Xiang, Z.~Liu, H.~Liu, Y.~Bai, J.~Cheng, and W.~Chen, ``Diffusiondialog: A diffusion model for diverse dialog generation with latent space,'' in \emph{Proceedings of the 2024 Joint International Conference on Computational Linguistics, Language Resources and Evaluation (LREC-COLING 2024)}, 2024, pp. 4912--4921.

\bibitem{zou2024improved}
W.~Zou, Z.~Zhuang, X.~Geng, S.~Huang, J.~Liu, and J.~Chen, ``Improved paraphrase generation via controllable latent diffusion,'' \emph{arXiv preprint arXiv:2404.08938}, 2024.

\bibitem{hu2024poetrydiffusion}
Z.~Hu, C.~Liu, Y.~Feng, A.~T. Luu, and B.~Hooi, ``Poetrydiffusion: Towards joint semantic and metrical manipulation in poetry generation,'' in \emph{Proceedings of the AAAI Conference on Artificial Intelligence}, vol.~38, no.~16, 2024, pp. 18\,279--18\,288.

\bibitem{chen2023xdlm}
L.~Chen, A.~Feng, B.~Yang, and Z.~Li, ``Xdlm: Cross-lingual diffusion language model for machine translation,'' \emph{arXiv preprint arXiv:2307.13560}, 2023.

\bibitem{qiao2023diffusionret}
S.~Qiao, X.~Liu, and S.-H. Na, ``Diffusionret: Diffusion-enhanced generative retriever using constrained decoding,'' in \emph{Findings of the Association for Computational Linguistics: EMNLP 2023}, 2023, pp. 9515--9529.

\bibitem{yan2025debunk}
K.~Yan, M.~Liu, Y.~Liu, R.~Fu, Z.~Wen, J.~Tao, and X.~Liu, ``Debunk and infer: Multimodal fake news detection via diffusion-generated evidence and llm reasoning,'' \emph{arXiv preprint arXiv:2506.21557}, 2025.

\bibitem{luxembourg2025plan}
O.~Luxembourg, H.~Permuter, and E.~Nachmani, ``Plan for speed--dilated scheduling for masked diffusion language models,'' \emph{arXiv preprint arXiv:2506.19037}, 2025.

\bibitem{fan2026stable}
C.~Fan, W.~Heng, B.~Li, S.~Liu, Y.~Song, J.~Su, X.~Qu, K.~Shen, and W.~Wei, ``Stable-diffcoder: Pushing the frontier of code diffusion large language model,'' \emph{arXiv preprint arXiv:2601.15892}, 2026.

\bibitem{bai2026dice}
H.~Bai, L.~Kong, X.~Chen, J.~Wang, Z.~Tao, and H.~Wang, ``Dice: Diffusion large language models excel at generating cuda kernels,'' \emph{arXiv preprint arXiv:2602.11715}, 2026.

\bibitem{xiong2024text}
Y.~Xiong, K.~Li, J.~Chen, H.~Zhang, D.~Lin, Y.~Che, and W.~Hu, ``Text-guided multi-property molecular optimization with a diffusion language model,'' \emph{arXiv preprint arXiv:2410.13597}, 2024.

\bibitem{gong2024text}
H.~Gong, Q.~Liu, S.~Wu, and L.~Wang, ``Text-guided molecule generation with diffusion language model,'' in \emph{Proceedings of the AAAI Conference on Artificial Intelligence}, vol.~38, no.~1, 2024, pp. 109--117.

\bibitem{goel2024memdlm}
S.~Goel, V.~Thoutam, E.~M. Marroquin, A.~Gokaslan, A.~Firouzbakht, S.~Vincoff, V.~Kuleshov, H.~T. Kratochvil, and P.~Chatterjee, ``Memdlm: De novo membrane protein design with masked discrete diffusion protein language models,'' in \emph{NeurIPS 2024 Workshop on AI for New Drug Modalities}.

\bibitem{wangdiffusion}
X.~Wang, Z.~Zheng, D.~Xue, S.~Huang, Q.~Gu \emph{et~al.}, ``Diffusion language models are versatile protein learners,'' in \emph{Forty-first International Conference on Machine Learning}.

\bibitem{yincfp}
J.~Yin, C.~Zha, W.~He, C.~Xu, and X.~Gao, ``Cfp-gen: Combinatorial functional protein generation via diffusion language models,'' in \emph{Forty-second International Conference on Machine Learning}.

\bibitem{wangfine}
C.~Wang, M.~Uehara, Y.~He, A.~Wang, A.~Lal, T.~Jaakkola, S.~Levine, A.~Regev, T.~Biancalani \emph{et~al.}, ``Fine-tuning discrete diffusion models via reward optimization with applications to dna and protein design,'' in \emph{The Thirteenth International Conference on Learning Representations}.

\bibitem{ni2024forcegen}
B.~Ni, D.~L. Kaplan, and M.~J. Buehler, ``Forcegen: End-to-end de novo protein generation based on nonlinear mechanical unfolding responses using a language diffusion model,'' \emph{Science Advances}, vol.~10, no.~6, p. eadl4000, 2024.

\bibitem{hallee2025diffusion}
L.~Hallee, N.~Rafailidis, D.~B. Bichara, and J.~P. Gleghorn, ``Diffusion sequence models for enhanced protein representation and generation,'' \emph{arXiv preprint arXiv:2506.08293}, 2025.

\bibitem{wang2024dplm}
X.~Wang, Z.~Zheng, F.~Ye, D.~Xue, S.~Huang, and Q.~Gu, ``Dplm-2: A multimodal diffusion protein language model,'' \emph{arXiv preprint arXiv:2410.13782}, 2024.

\bibitem{wen2025llada}
Y.~Wen, H.~Li, K.~Gu, Y.~Zhao, T.~Wang, and X.~Sun, ``Llada-vla: Vision language diffusion action models,'' \emph{arXiv preprint arXiv:2509.06932}, 2025.

\bibitem{wen2025dvla}
J.~Wen, M.~Zhu, J.~Liu, Z.~Liu, Y.~Yang, L.~Zhang, S.~Zhang, Y.~Zhu, and Y.~Xu, ``dvla: Diffusion vision-language-action model with multimodal chain-of-thought,'' \emph{arXiv preprint arXiv:2509.25681}, 2025.

\bibitem{chen2025unified}
J.~Chen, W.~Song, P.~Ding, Z.~Zhou, H.~Zhao, F.~Tang, D.~Wang, and H.~Li, ``Unified diffusion vla: Vision-language-action model via joint discrete denoising diffusion process,'' \emph{arXiv preprint arXiv:2511.01718}, 2025.

\bibitem{sohl2015deep}
J.~Sohl-Dickstein, E.~Weiss, N.~Maheswaranathan, and S.~Ganguli, ``Deep unsupervised learning using nonequilibrium thermodynamics,'' in \emph{International conference on machine learning}.\hskip 1em plus 0.5em minus 0.4em\relax pmlr, 2015, pp. 2256--2265.

\bibitem{devlin2019bert}
J.~Devlin, M.-W. Chang, K.~Lee, and K.~Toutanova, ``Bert: Pre-training of deep bidirectional transformers for language understanding,'' in \emph{Proceedings of the 2019 conference of the North American chapter of the association for computational linguistics: human language technologies, volume 1 (long and short papers)}, 2019, pp. 4171--4186.

\bibitem{liu2019roberta}
Y.~Liu, M.~Ott, N.~Goyal, J.~Du, M.~Joshi, D.~Chen, O.~Levy, M.~Lewis, L.~Zettlemoyer, and V.~Stoyanov, ``Roberta: A robustly optimized bert pretraining approach,'' \emph{arXiv preprint arXiv:1907.11692}, 2019.

\bibitem{lanalbert}
Z.~Lan, M.~Chen, S.~Goodman, K.~Gimpel, P.~Sharma, and R.~Soricut, ``Albert: A lite bert for self-supervised learning of language representations,'' in \emph{International Conference on Learning Representations}.

\bibitem{hedeberta}
P.~He, X.~Liu, J.~Gao, and W.~Chen, ``Deberta: Decoding-enhanced bert with disentangled attention,'' in \emph{International Conference on Learning Representations}.

\bibitem{dai2019transformer}
Z.~Dai, Z.~Yang, Y.~Yang, J.~Carbonell, Q.~V. Le, and R.~Salakhutdinov, ``Transformer-xl: Attentive language models beyond a fixed-length context,'' \emph{arXiv preprint arXiv:1901.02860}, 2019.

\bibitem{zhang2022opt}
S.~Zhang, S.~Roller, N.~Goyal, M.~Artetxe, M.~Chen, S.~Chen, C.~Dewan, M.~Diab, X.~Li, X.~V. Lin \emph{et~al.}, ``Opt: Open pre-trained transformer language models,'' \emph{arXiv preprint arXiv:2205.01068}, 2022.

\bibitem{gloecklebetter}
F.~Gloeckle, B.~Y. Idrissi, B.~Roziere, D.~Lopez-Paz, and G.~Synnaeve, ``Better \& faster large language models via multi-token prediction,'' in \emph{Forty-first International Conference on Machine Learning}.

\bibitem{chen2025diffpo}
R.~Chen, W.~Chai, Z.~Yang, X.~Zhang, J.~T. Zhou, T.~Quek, S.~Poria, and Z.~Liu, ``Diffpo: Diffusion-styled preference optimization for efficient inference-time alignment of large language models,'' \emph{arXiv preprint arXiv:2503.04240}, 2025.

\bibitem{sutskever2014sequence}
I.~Sutskever, O.~Vinyals, and Q.~V. Le, ``Sequence to sequence learning with neural networks,'' \emph{Advances in neural information processing systems}, vol.~27, 2014.

\bibitem{raffel2020exploring}
C.~Raffel, N.~Shazeer, A.~Roberts, K.~Lee, S.~Narang, M.~Matena, Y.~Zhou, W.~Li, and P.~J. Liu, ``Exploring the limits of transfer learning with a unified text-to-text transformer,'' \emph{Journal of machine learning research}, vol.~21, no. 140, pp. 1--67, 2020.

\bibitem{lewis2020bart}
M.~Lewis, Y.~Liu, N.~Goyal, M.~Ghazvininejad, A.~Mohamed, O.~Levy, V.~Stoyanov, and L.~Zettlemoyer, ``Bart: Denoising sequence-to-sequence pre-training for natural language generation, translation, and comprehension,'' in \emph{Proceedings of the 58th Annual Meeting of the Association for Computational Linguistics}, 2020, pp. 7871--7880.

\bibitem{yuan2022seqdiffuseq}
H.~Yuan, Z.~Yuan, C.~Tan, F.~Huang, and S.~Huang, ``Seqdiffuseq: Text diffusion with encoder-decoder transformers,'' \emph{arXiv preprint arXiv:2212.10325}, 2022.

\bibitem{yang2019xlnet}
Z.~Yang, Z.~Dai, Y.~Yang, J.~Carbonell, R.~R. Salakhutdinov, and Q.~V. Le, ``Xlnet: Generalized autoregressive pretraining for language understanding,'' \emph{Advances in neural information processing systems}, vol.~32, 2019.

\bibitem{chen2022analog}
T.~Chen, R.~Zhang, and G.~Hinton, ``Analog bits: Generating discrete data using diffusion models with self-conditioning,'' \emph{arXiv preprint arXiv:2208.04202}, 2022.

\bibitem{qwen2.5}
\BIBentryALTinterwordspacing
Q.~Team, ``Qwen2.5: A party of foundation models,'' September 2024. [Online]. Available: \url{https://qwenlm.github.io/blog/qwen2.5/}
\BIBentrySTDinterwordspacing

\bibitem{he2025ultrallada}
G.~He, S.~Nie, F.~Zhu, Y.~Zhao, T.~Bai, R.~Yan, J.~Fu, C.~Li, and B.~Yuan, ``Ultrallada: Scaling the context length to 128k for diffusion large language models,'' \emph{arXiv preprint arXiv:2510.10481}, 2025.

\bibitem{zhu2024segment}
X.~Zhu, G.~Karadzhov, C.~Whitehouse, and A.~Vlachos, ``Segment-level diffusion: A framework for controllable long-form generation with diffusion language models,'' \emph{arXiv preprint arXiv:2412.11333}, 2024.

\bibitem{ladida}
\BIBentryALTinterwordspacing
Y.~Zihuiwen, Y.~Elle~Michelle, and B.~Phil, ``Latent diffusion for document generation with sequential decoding,'' in \emph{NeurIPS 2023 Workshop on Diffusion Models}, 2023. [Online]. Available: \url{https://neurips.cc/virtual/2023/74876}
\BIBentrySTDinterwordspacing

\bibitem{cetin2025large}
E.~Cetin, T.~Zhao, and Y.~Tang, ``Large language models to diffusion finetuning,'' \emph{arXiv preprint arXiv:2501.15781}, 2025.

\bibitem{bai2024meissonic}
J.~Bai, T.~Ye, W.~Chow, E.~Song, Q.-G. Chen, X.~Li, Z.~Dong, L.~Zhu, and S.~Yan, ``Meissonic: Revitalizing masked generative transformers for efficient high-resolution text-to-image synthesis,'' in \emph{The Thirteenth International Conference on Learning Representations}, 2024.

\bibitem{ni2025training}
J.~Ni, Q.~Liu, C.~Du, L.~Dou, H.~Yan, Z.~Wang, T.~Pang, and M.~Q. Shieh, ``Training optimal large diffusion language models,'' \emph{arXiv preprint arXiv:2510.03280}, 2025.

\bibitem{ni2025diffusion}
J.~Ni, Q.~Liu, L.~Dou, C.~Du, Z.~Wang, H.~Yan, T.~Pang, and M.~Q. Shieh, ``Diffusion language models are super data learners,'' \emph{arXiv preprint arXiv:2511.03276}, 2025.

\bibitem{asada2025addressing}
M.~Asada and M.~Miwa, ``Addressing the training-inference discrepancy in discrete diffusion for text generation,'' in \emph{Proceedings of the 31st International Conference on Computational Linguistics}, 2025, pp. 7156--7164.

\bibitem{vaswani2017attention}
A.~Vaswani, N.~Shazeer, N.~Parmar, J.~Uszkoreit, L.~Jones, A.~N. Gomez, {\L}.~Kaiser, and I.~Polosukhin, ``Attention is all you need,'' \emph{Advances in neural information processing systems}, vol.~30, 2017.

\bibitem{sauer2024adversarial}
A.~Sauer, D.~Lorenz, A.~Blattmann, and R.~Rombach, ``Adversarial diffusion distillation,'' in \emph{European Conference on Computer Vision}.\hskip 1em plus 0.5em minus 0.4em\relax Springer, 2024, pp. 87--103.

\bibitem{sauer2024fast}
A.~Sauer, F.~Boesel, T.~Dockhorn, A.~Blattmann, P.~Esser, and R.~Rombach, ``Fast high-resolution image synthesis with latent adversarial diffusion distillation,'' in \emph{SIGGRAPH Asia 2024 Conference Papers}, 2024, pp. 1--11.

\bibitem{liu2023visual}
H.~Liu, C.~Li, Q.~Wu, and Y.~J. Lee, ``Visual instruction tuning,'' \emph{Advances in neural information processing systems}, vol.~36, pp. 34\,892--34\,916, 2023.

\bibitem{lillava}
F.~Li, R.~Zhang, H.~Zhang, Y.~Zhang, B.~Li, W.~Li, Z.~Ma, and C.~Li, ``Llava-interleave: Tackling multi-image, video, and 3d in large multimodal models,'' in \emph{The Thirteenth International Conference on Learning Representations}.

\bibitem{guo2024mammoth}
J.~Guo, T.~Zheng, Y.~Bai, B.~Li, Y.~Wang, K.~Zhu, Y.~Li, G.~Neubig, W.~Chen, and X.~Yue, ``Mammoth-vl: Eliciting multimodal reasoning with instruction tuning at scale,'' \emph{arXiv preprint arXiv:2412.05237}, 2024.

\bibitem{grattafiori2024llama}
A.~Grattafiori, A.~Dubey, A.~Jauhri, A.~Pandey, A.~Kadian, A.~Al-Dahle, A.~Letman, A.~Mathur, A.~Schelten, A.~Vaughan \emph{et~al.}, ``The llama 3 herd of models,'' \emph{arXiv preprint arXiv:2407.21783}, 2024.

\bibitem{wang2024qwen2}
P.~Wang, S.~Bai, S.~Tan, S.~Wang, Z.~Fan, J.~Bai, K.~Chen, X.~Liu, J.~Wang, W.~Ge \emph{et~al.}, ``Qwen2-vl: Enhancing vision-language model's perception of the world at any resolution,'' \emph{arXiv preprint arXiv:2409.12191}, 2024.

\bibitem{xieshow}
J.~Xie, W.~Mao, Z.~Bai, D.~J. Zhang, W.~Wang, K.~Q. Lin, Y.~Gu, Z.~Chen, Z.~Yang, and M.~Z. Shou, ``Show-o: One single transformer to unify multimodal understanding and generation,'' in \emph{The Thirteenth International Conference on Learning Representations}.

\bibitem{kou2024orthus}
S.~Kou, J.~Jin, Z.~Liu, C.~Liu, Y.~Ma, J.~Jia, Q.~Chen, P.~Jiang, and Z.~Deng, ``Orthus: Autoregressive interleaved image-text generation with modality-specific heads,'' \emph{arXiv preprint arXiv:2412.00127}, 2024.

\bibitem{wu2025janus}
C.~Wu, X.~Chen, Z.~Wu, Y.~Ma, X.~Liu, Z.~Pan, W.~Liu, Z.~Xie, X.~Yu, C.~Ruan \emph{et~al.}, ``Janus: Decoupling visual encoding for unified multimodal understanding and generation,'' in \emph{Proceedings of the Computer Vision and Pattern Recognition Conference}, 2025, pp. 12\,966--12\,977.

\bibitem{patil2024amused}
S.~Patil, W.~Berman, R.~Rombach, and P.~von Platen, ``amused: An open muse reproduction,'' \emph{arXiv preprint arXiv:2401.01808}, 2024.

\bibitem{Pref-GRPO&UniGenBench}
Y.~Wang, Z.~Li, Y.~Zang, Y.~Zhou, J.~Bu, C.~Wang, Q.~Lu, C.~Jin, and J.~Wang, ``Pref-grpo: Pairwise preference reward-based grpo for stable text-to-image reinforcement learning,'' \emph{arXiv preprint arXiv:2508.20751}, 2025.

\bibitem{bisk2020piqa}
Y.~Bisk, R.~Zellers, J.~Gao, Y.~Choi \emph{et~al.}, ``Piqa: Reasoning about physical commonsense in natural language,'' in \emph{Proceedings of the AAAI conference on artificial intelligence}, vol.~34, no.~05, 2020, pp. 7432--7439.

\bibitem{zellers2019hellaswag}
R.~Zellers, A.~Holtzman, Y.~Bisk, A.~Farhadi, and Y.~Choi, ``Hellaswag: Can a machine really finish your sentence?'' in \emph{Proceedings of the 57th Annual Meeting of the Association for Computational Linguistics}, 2019, pp. 4791--4800.

\bibitem{chen2021evaluating}
M.~Chen, J.~Tworek, H.~Jun, Q.~Yuan, H.~P. D.~O. Pinto, J.~Kaplan, H.~Edwards, Y.~Burda, N.~Joseph, G.~Brockman \emph{et~al.}, ``Evaluating large language models trained on code,'' \emph{arXiv preprint arXiv:2107.03374}, 2021.

\bibitem{ghosh2023geneval}
D.~Ghosh, H.~Hajishirzi, and L.~Schmidt, ``Geneval: An object-focused framework for evaluating text-to-image alignment,'' \emph{Advances in Neural Information Processing Systems}, vol.~36, pp. 52\,132--52\,152, 2023.

\bibitem{fu2023mme}
C.~Fu, P.~Chen, Y.~Shen, Y.~Qin, M.~Zhang, X.~Lin, J.~Yang, X.~Zheng, K.~Li, X.~Sun \emph{et~al.}, ``Mme: A comprehensive evaluation benchmark for multimodal large language models,'' \emph{arXiv preprint arXiv:2306.13394}, 2023.

\bibitem{yue2024mmmu}
X.~Yue, Y.~Ni, K.~Zhang, T.~Zheng, R.~Liu, G.~Zhang, S.~Stevens, D.~Jiang, W.~Ren, Y.~Sun \emph{et~al.}, ``Mmmu: A massive multi-discipline multimodal understanding and reasoning benchmark for expert agi,'' in \emph{Proceedings of the IEEE/CVF Conference on Computer Vision and Pattern Recognition}, 2024, pp. 9556--9567.

\bibitem{hudson2019gqa}
D.~A. Hudson and C.~D. Manning, ``Gqa: A new dataset for real-world visual reasoning and compositional question answering,'' in \emph{Proceedings of the IEEE/CVF conference on computer vision and pattern recognition}, 2019, pp. 6700--6709.

\bibitem{cobbe2021training}
K.~Cobbe, V.~Kosaraju, M.~Bavarian, M.~Chen, H.~Jun, L.~Kaiser, M.~Plappert, J.~Tworek, J.~Hilton, R.~Nakano \emph{et~al.}, ``Training verifiers to solve math word problems,'' \emph{arXiv preprint arXiv:2110.14168}, 2021.

\bibitem{team2024qwen2}
Q.~Team, ``Qwen2 technical report,'' \emph{arXiv preprint arXiv:2407.10671}, 2024.

\bibitem{rein2024gpqa}
D.~Rein, B.~L. Hou, A.~C. Stickland, J.~Petty, R.~Y. Pang, J.~Dirani, J.~Michael, and S.~R. Bowman, ``Gpqa: A graduate-level google-proof q\&a benchmark,'' in \emph{First Conference on Language Modeling}, 2024.

\bibitem{hendrycks2measuring}
D.~Hendrycks, C.~Burns, S.~Kadavath, A.~Arora, S.~Basart, E.~Tang, D.~Song, and J.~Steinhardt, ``Measuring mathematical problem solving with the math dataset,'' in \emph{Thirty-fifth Conference on Neural Information Processing Systems Datasets and Benchmarks Track (Round 2)}.

\bibitem{lyu2023fine}
Y.~Lyu, T.~Luo, J.~Shi, T.~C. Hollon, and H.~Lee, ``Fine-grained text style transfer with diffusion-based language models,'' \emph{arXiv preprint arXiv:2305.19512}, 2023.

\bibitem{demirag2024benchmarking}
Y.~Demirag, D.~Liu, and J.~Niehues, ``Benchmarking diffusion models for machine translation,'' in \emph{Proceedings of the 18th Conference of the European Chapter of the Association for Computational Linguistics: Student Research Workshop}, 2024, pp. 313--324.

\bibitem{wolf2020transformers}
T.~Wolf, L.~Debut, V.~Sanh, J.~Chaumond, C.~Delangue, A.~Moi, P.~Cistac, T.~Rault, R.~Louf, M.~Funtowicz \emph{et~al.}, ``Transformers: State-of-the-art natural language processing,'' in \emph{Proceedings of the 2020 conference on empirical methods in natural language processing: system demonstrations}, 2020, pp. 38--45.

\bibitem{kwon2023efficient}
W.~Kwon, Z.~Li, S.~Zhuang, Y.~Sheng, L.~Zheng, C.~H. Yu, J.~E. Gonzalez, H.~Zhang, and I.~Stoica, ``Efficient memory management for large language model serving with pagedattention,'' in \emph{Proceedings of the ACM SIGOPS 29th Symposium on Operating Systems Principles}, 2023.

\bibitem{Dreamon2025}
\BIBentryALTinterwordspacing
Z.~Wu, L.~Zheng, Z.~Xie, J.~Ye, J.~Gao, Y.~Feng, Z.~Li, V.~W., G.~Zhou, and L.~Kong, ``Dreamon: Diffusion language models for code infilling beyond fixed-size canvas,'' 2025. [Online]. Available: \url{https://hkunlp.github.io/blog/2025/dreamon}
\BIBentrySTDinterwordspacing

\bibitem{yang2025diffusion}
Y.~Yang, C.~Wang, S.~Wang, Z.~Wen, B.~Qi, H.~Xu, and L.~Zhang, ``Diffusion llm with native variable generation lengths: Let [eos] lead the way,'' \emph{arXiv preprint arXiv:2510.24605}, 2025.

\bibitem{li2025beyond}
J.~Li, X.~Dong, Y.~Zang, Y.~Cao, J.~Wang, and D.~Lin, ``Beyond fixed: Training-free variable-length denoising for diffusion large language models,'' \emph{arXiv preprint arXiv:2508.00819}, 2025.

\bibitem{chen2025dpad}
X.~Chen, S.~Huang, C.~Guo, C.~Wei, Y.~He, J.~Zhang, H.~Li, Y.~Chen \emph{et~al.}, ``Dpad: Efficient diffusion language models with suffix dropout,'' \emph{arXiv preprint arXiv:2508.14148}, 2025.

\bibitem{yang2025qwen3}
A.~Yang, A.~Li, B.~Yang, B.~Zhang, B.~Hui, B.~Zheng, B.~Yu, C.~Gao, C.~Huang, C.~Lv \emph{et~al.}, ``Qwen3 technical report,'' \emph{arXiv preprint arXiv:2505.09388}, 2025.

\bibitem{team2025kimi}
K.~Team, Y.~Bai, Y.~Bao, G.~Chen, J.~Chen, N.~Chen, R.~Chen, Y.~Chen, Y.~Chen, Y.~Chen \emph{et~al.}, ``Kimi k2: Open agentic intelligence,'' \emph{arXiv preprint arXiv:2507.20534}, 2025.

\end{thebibliography}
\bibliographystyle{IEEEtran}

%{\appendices
%\section*{Proof of the First Zonklar Equation}
%Appendix one text goes here.
% You can choose not to have a title for an appendix if you want by leaving the argument blank
%\section*{Proof of the Second Zonklar Equation}
%Appendix two text goes here.}

\vfill

\end{document}